\documentclass[5p,times]{elsarticle}
\usepackage{amsmath}
\usepackage{mathalfa}
 \usepackage{soul,xcolor}
\sethlcolor{white}

\usepackage{wasysym}%
\usepackage{graphicx}
\usepackage{cite}
\usepackage{tikz}
\usetikzlibrary{arrows.meta,positioning,calc}

 \usepackage{amsfonts}
\usepackage{amsmath}
\usepackage{algpseudocode}
\usepackage{algorithm}
 \usepackage{multirow}
\usepackage{algpseudocode}
\usepackage{amsmath}
\newcommand{\emp}[1]{\emph{#1}}
\usepackage{subcaption} 
\usepackage{amsmath, amssymb, graphicx}
\usepackage{booktabs}
\usepackage{hyperref}
\usepackage[table]{xcolor}
\usepackage[graphicx]{realboxes}
\usepackage{titlesec}
\titleformat{\paragraph}[runin]{\normalfont\bfseries}{\theparagraph}{1em}{}
\usepackage{pifont}
\usepackage{amssymb}
\usepackage{bm} 
\newcommand{\vct}[1]{\bm{#1}}       
\newcommand{\mat}[1]{\mathbf{#1}}   
\newcommand{\set}[1]{\mathcal{#1}}  
\newcommand{\R}{\mathbb{R}}

\newcommand{\mX}{\mat{X}}   \newcommand{\mZ}{\mat{Z}}   \newcommand{\mW}{\mat{W}}
\newcommand{\vx}{\vct{x}}   \newcommand{\vz}{\vct{z}}   
   \newcommand{\vtheta}{\bm{\theta}}
\newcommand{\Ds}{\set{D}_s} \newcommand{\Dt}{\set{D}_t}

\begin{document}

\begin{frontmatter}



\title{From One Attack Domain to Another: Contrastive Transfer Learning with Siamese Networks for APT Detection.}

\author[nyuad]{Sidahmed Benabderrahmane\corref{cor1}}
\ead{sidahmed.benabderrahmane@nyu.edu}

\author[nyuad]{Talal Rahwan}
\ead{tr74@nyu.edu}

\cortext[cor1]{Corresponding author}

\address[nyuad]{New York University, NYUAD, Division of Science.}




\begin{abstract}
\hl{Advanced Persistent Threats (APT) pose a major cybersecurity challenge due to their stealth, persistence, and adaptability. Traditional machine learning detectors struggle with class imbalance, high-dimensional features, and scarce real-world traces. They often lack transferability—performing well in the training domain but degrading in novel attack scenarios. We propose a hybrid transfer framework that integrates Transfer Learning, Explainable AI (XAI), contrastive learning, and Siamese networks to improve cross-domain generalization. An attention-based autoencoder supports knowledge transfer across domains, while Shapley Additive exPlanations (SHAP) select stable, informative features to reduce dimensionality and computational cost. A Siamese encoder trained with a contrastive objective aligns source and target representations, increasing anomaly separability and mitigating feature drift. We evaluate on real-world traces from the DARPA Transparent Computing (TC) program and augment with synthetic attack scenarios to test robustness. Across source to target transfers, the approach delivers improved detection scores with classical and deep baselines, demonstrating a scalable, explainable, and transferable solution for APT detection.}\\
\end{abstract}


\end{frontmatter}

\section{Introduction}

``The greatest victory is that which requires no battle." – Sun Tzu, \textit{The Art of War}. Threats in warfare have evolved from the battlefield to the digital realm, where cyber adversaries deploy sophisticated and persistent attack techniques \citep{xuan2024novel}. In today's hyper-connected world, cybersecurity has become a new frontier of warfare, where Advanced Persistent Threats (APTs) are among the most formidable adversaries \citep{Sakthivelu23}. APTs represent an unprecedented challenge to traditional cybersecurity defenses, given their stealth, adaptability, and ability to remain hidden within networks for prolonged periods. Unlike conventional malware, which can often be detected through signature-based methods, APTs do not rely on predictable attack patterns \citep{arefin2024understanding}. Instead, they employ multi-stage, long-term attack strategies, leveraging social engineering, zero-day exploits, privilege escalation, and lateral movement to infiltrate target networks \citep{che2024systematic}. Once inside, they establish a persistent foothold, allowing them to execute their objectives—typically intelligence gathering, data exfiltration, or critical system disruption—with minimal risk of detection \citep{buchta2024advanced}.

Despite the existence of numerous cybersecurity solutions, current detection mechanisms remain largely inadequate against APTs \citep{DBLP:journals/fgcs/BenabderrahmaneHVCR24}. Traditional rule-based systems struggle to identify novel or highly customized attack behaviors, as they primarily rely on known indicators of compromise (IoCs) \citep{roesch1999snort}. Meanwhile, supervised machine learning approaches face severe limitations, as they depend on large, labeled datasets that are rarely available for APTs. Threat actors continuously modify their attack signatures, rendering static detection ineffective and necessitating more adaptable detection methods that can learn to recognize evolving attack patterns \citep{arefin2024understanding}. Moreover, conventional feature-based learning approaches often fail to generalize across attack scenarios due to inconsistencies in feature representations between different datasets. Addressing this limitation requires an advanced feature selection and alignment mechanism to enhance the quality of learned representations \citep{inbookKong20}.

Another fundamental challenge with APT detection is the targeted nature of these attacks. Unlike mass-distributed malware that follows a relatively standardized pattern, APT tactics are highly dependent on the victim organization’s industry, infrastructure, and security posture \citep{al2023machine}. This results in attack traces that are inconsistent across different datasets, making it extremely difficult to develop generalizable detection models. Consequently, cybersecurity researchers struggle to build machine learning models that perform reliably across varied attack environments, as training on one dataset often does not transfer well to new attack scenarios \citep{wang2025novel}. This challenge necessitates a robust transfer learning approach that can effectively adapt knowledge learned from one attack scenario to another, ensuring cross-scenario generalization.

Moreover, APT cybersecurity datasets are notoriously scarce, highly imbalanced, and exhibit heterogeneous feature representations, further exacerbating the difficulties of APT detection \citep{chen2022machine}. Anomaly detection models trained on limited and skewed data often yield high false positive rates, reducing their operational viability \citep{chavan2023review}, \citep{chaudhary2023machine}. Traditional intrusion detection systems, which primarily depend on static indicators of compromise (IoCs) or signature-based detection, are inherently incapable of adapting to the dynamic evolution of APT campaigns \citep{shi2018malicious}. Therefore, more sophisticated, learning-driven techniques are required—ones that can generalize knowledge from limited attack traces, align feature representations across datasets, and effectively distinguish between benign and malicious activities. Beyond dataset scarcity, another critical issue is the presence of redundant and uninformative features, which increase computational complexity and reduce model effectiveness. Thus, a well-defined feature filtering mechanism is needed to retain only the most relevant features while improving detection performance \citep{zuhairi2024realtime}.

\hl{Despite substantial progress, existing machine learning-based APT detection approaches fail to holistically address the multifaceted challenges posed by the real-world cyber threats. Many proposed anomaly detection models suffer from key drawbacks. A common limitation is their lack of \textit{transferability}, as they are typically trained in a closed-world setting—performing well on the training dataset but failing to generalize to new, unseen attack scenarios.  Additionally, different cybersecurity datasets present an inability to handle feature space misalignment since they often have different feature representations due to variations in logging, network configurations, or monitoring tools, and traditional models do not account for this discrepancy, leading to poor generalization across datasets} \citep{pandey2025accidental}.\\ \hl{Supervised machine learning models further struggle due to their dependency on labeled datasets, which are often incomplete or fail to capture the full life-cycle of an APT. As a result, such models tend to bias toward early-stage attack indicators while missing later-stage tactics such as persistence and data exfiltration, thus failing to capture long-term attack behaviors. Compounding these limitations, many prior works rely heavily on evaluation metrics such as AUC (Area Under the ROC Curve), which are not always suitable for cybersecurity applications. In practice, ranking the most severe threats is often more critical than binary classification, and many existing models lack the ability to prioritize the most dangerous threats effectively. Lastly, the absence of \textit{explainability} in many anomaly detection pipelines remains a significant barrier to operational deployment. Without insight into how features influence model decisions, it becomes challenging for security analysts to interpret and trust automated alerts }\citep{li2023survey}, \citep{li2024survey}.

To overcome these critical shortcomings, we propose a novel hybrid learning framework that integrates {feature selection}-driven {transfer learning}, {explainable AI (XAI)}-guided {feature importance ranking, contrastive learning}, and {Siamese networks} to enhance APT anomaly detection in imbalanced datasets. Our approach leverages transfer learning \citep{wang2025novel} to address the scarcity of APT datasets by training on a known attack dataset and applying the learned knowledge to new attack scenarios, including synthetic attacks generated using cGANs and VAEs. This enables cross-scenario generalization, improving model robustness without requiring large labeled datasets for every new APT. To optimize transfer learning performance, we incorporate an explainable feature selection mechanism that ranks feature importance using a combined metric based on SHAP values, reconstruction error from an attention-based autoencoder, and entropy-based feature informativeness. By filtering out irrelevant features, we significantly enhance model efficiency and focus learning on the most critical attack indicators.

In addition, contrastive learning is integrated to refine feature representations, ensuring that similar attack behaviors remain close in feature space while distinct anomalies remain separated \citep{hu2024comprehensive}. Contrary to traditional anomaly detection models that struggle with inconsistencies in feature spaces between different datasets, contrastive learning in our model would improve the ranking of detected anomalies, as evidenced by higher scores. Furthermore, we leverage contrastive learning to enforce consistency in the refined feature space, preventing feature drifts between attack datasets.

Moreover, when applying transfer learning, feature space misalignment between source and target datasets can degrade model effectiveness. To mitigate this, we introduce a Siamese Network-based feature alignment mechanism, which explicitly measures feature similarity between different attack datasets \citep{fedele2024explaining}. The resulting Siamese Distance metric provides an indicator of transferability, ensuring better knowledge transfer and improved anomaly detection performance. Given the rarity of real-world APT data, we further employ generative adversarial networks (cGANs) and variational autoencoders (VAEs) to generate synthetic attack scenarios. These synthetic datasets allow us to expand our training set, providing more diverse attack patterns. By aligning features using Siamese networks, we ensure that knowledge transfer across datasets is seamless, allowing for a more robust anomaly detection system that effectively adapts to different attack landscapes.

To enhance interpretability, we provide feature space visualizations before and after Siamese Network alignment, demonstrating how our approach refines the decision boundaries and correlates feature alignment with improved anomaly ranking. Additionally, we evaluate our framework using both AUC and nDCG scores, ensuring that the ranking of detected threats remains effective even when raw classification metrics do not fully capture attack severity.

To the best of our knowledge, this is the first work to unify feature selection-based transfer learning, contrastive learning, and Siamese network-based feature space alignment into a single coherent framework for Advanced Persistent Threat (APT) detection. Our contributions bridge critical gaps between knowledge transferability, feature representation alignment, and anomaly ranking. Furthermore, the integration of explainable feature selection with synthetic data augmentation via cGANs and VAEs addresses an underexplored area in cybersecurity, offering a novel perspective on AI-driven threat detection. By combining these complementary techniques into a unified architecture, we substantially improve the detection of APTs in imbalanced and complex cybersecurity environments.

\hl{This paper makes the following contributions:}
\begin{itemize}
  \item \hl{Siamese contrastive transfer: a framework that couples contrastive learning with a Siamese network to align source $\rightarrow$ target embeddings and mitigate feature drift.}
  \item \hl{XAI-guided feature selection: a SHAP, entropy, reconstruction criterion that preserves stable, informative features and reduces dimensionality and compute.}
  \item \hl{Attention-based autoencoder backbone: robust latent representations enabling label-efficient transfer with minimal target supervision.}
  \item \hl{Evaluation with ranking metrics and stats: results in AUC and nDCG with Friedman/Wilcoxon tests across transfers.}
  \item \hl{Ablations and strong baselines: component ablations and comparisons to classical and deep baselines under matched budgets.}
\end{itemize}

The remainder of the paper is organized as follows. Section 2 presents the existing methods, while Section 3 describes the proposed framework by presenting the concept of transfer learning, contrastive learning, and the Siamese network architecture. Section 4 presents the experimental settings, results, and discussions. Finally, Section 5 sums up the main results of this work, followed by some future perspectives.
\section{Related Work}
\subsection{Advanced Persistent Threats}
Advanced Persistent Threats (APTs) have been responsible for some of the most devastating cyberattacks in history, targeting critical infrastructure, government institutions, and major corporations \citep{che2024systematic}, \citep{talib2022apt}. Notable APT incidents include Stuxnet, a sophisticated malware attack that targeted Iran’s nuclear facilities in 2010, causing substantial physical damage to centrifuges through cyber-physical sabotage \citep{kumar2022apt}. Another infamous attack, APT29 (Cozy Bear), attributed to Russian state-sponsored actors, infiltrated the U.S. Democratic National Committee (DNC) in 2016, leading to major geopolitical consequences \citep{stojanovic2020apt}. The APT10 (Cloud Hopper) campaign, linked to Chinese espionage groups, compromised major managed service providers (MSPs), affecting numerous global enterprises and leading to vast intellectual property theft. More recently, the SolarWinds attack in 2020, attributed to APT29, saw the compromise of thousands of organizations, including government agencies and Fortune 500 companies, by injecting malicious updates into a widely used IT monitoring tool. These attacks highlight the growing sophistication and persistence of APT campaigns, demonstrating the inadequacy of traditional cybersecurity defenses in detecting and mitigating such threats \citep{alkhadra2021solar}.

 \hl{Following SolarWinds, multiple state-aligned APT operations underscored the shift toward supply-chain and “living-off-the-land” tradecraft. In early 2021, HAFNIUM exploited four zero-days in on-premises Microsoft Exchange to establish web shells and long-term access across thousands of organizations} \citep{Microsoft2021Hafnium,CISA2021Exchange}. \hl{In 2023, UNC4841 abused a zero-day (CVE-2023-2868) in Barracuda Email Security Gateway appliances to deploy bespoke backdoors and conduct intelligence collection, with activity traced back to late 2022} \citep{CISA2023Barracuda,Mandiant2023UNC4841}. \hl{The same year, a software supply-chain compromise of the 3CX desktop app—ultimately linked to DPRK/Lazarus—propagated trojanized installers to enterprise endpoints (a “double” supply-chain incident)} \citep{Google20233CX,Avertium20233CX}. \hl{In 2023–2024, “Volt Typhoon,” assessed as PRC-sponsored, infiltrated U.S. critical-infrastructure networks using valid accounts and living-off-the-land techniques, with government advisories warning about pre-positioning for potential disruption} \citep{CISA-VoltTyphoon-2024}. \hl{Over 2024–2025, “Midnight Blizzard” (APT29/NOBELIUM) targeted Microsoft’s corporate email and downstream customers, illustrating persistent espionage against cloud and identity platforms} \citep{MSRC2024Midnight,Reuters2024Midnight}.

\subsection{\texorpdfstring{\hl{Machine- and Deep-Learning for APT Detection}}{Machine- and Deep-Learning for APT Detection}}

The detection of Advanced Persistent Threats (APTs) remains a critical challenge in cybersecurity \citep{mei2022review}. Unlike traditional cyberattacks, APTs are highly sophisticated, employing stealthy, multi-stage attack strategies designed to evade conventional intrusion detection systems (IDS) and security measures \citep{muneer2024critical}. Traditional rule-based detection mechanisms and supervised learning approaches struggle to adapt to the dynamic and evolving nature of APTs \citep{abdulganiyu2024towards}. This difficulty is further compounded by the extreme scarcity of labeled APT datasets and the significant imbalance between normal and malicious activities, leading to high false positive and false negative rates in anomaly detection \citep{vibhute2024deep}.
Several anomaly detection approaches have been explored for detecting Advanced Persistent Threats (APTs), each with varying degrees of success \citep{wang2024combating,Benabderrahmane21,benabderrahmane_2019,DBLP:journals/fgcs/BenabderrahmaneHVCR24,jia2024magic}. Rule-based and signature-based methods, such as Snort and Suricata, rely on predefined attack signatures and heuristics to identify malicious activity \citep{roesch1999snort}. However, these approaches are ineffective against novel or evolving APTs that employ zero-day exploits and sophisticated evasion techniques. Statistical anomaly detection methods, including Principal Component Analysis (PCA) and clustering-based models like k-means, attempt to establish a baseline of normal behavior and flag deviations \citep{fahrmann2024anomaly}. Other statistical methods, such as Hidden Markov Models (HMMs) and Gaussian Mixture Models (GMMs), have been applied to model system behavior and identify deviations, but they often struggle to capture the complexities of evolving cyber threats \citep{alavizadeh2024markov}. Although these techniques can detect unknown threats, they frequently suffer from high false-positive rates due to the dynamic nature of enterprise networks.

\hl{Machine learning–based approaches have gained traction in APT detection, leveraging algorithms such as Isolation Forest (IF), One-Class SVM (OC SVM), and k-Nearest Neighbors (KNN) to identify deviations from normal activity} \citep{alshuaibi2025machine}. \hl{Isolation Forest scores anomalies by how quickly points are isolated in random partitioning trees—processes with sparse, atypical event patterns yield shorter path lengths and thus higher anomaly scores; it is label-free and scales to binary provenance features, but can over-flag rare benign behaviors and is sensitive to the contamination and tree-depth settings} \citep{ripan2022effectively}. \hl{One-Class SVM instead learns a tight boundary around “normal” activity and flags points outside; it works well when trained on mostly benign data, but performance can degrade under domain shift and in very high-dimensional sparse spaces without careful scaling/tuning} \citep{cheng2025tagapt,benabderrahmane2025adversarial}. \hl{KNN-based anomaly scoring uses distances to the $K$-th neighbor to expose unusual process behaviors yet it is sensitive to the choice of $K$, suffers from dimensionality, and can be unstable under severe class imbalance} \citep{ibrahim2025optimized}. \hl{Frequent and rare itemset mining over provenance-derived feature vectors have been used to expose both common attack motifs and low-support, high-suspicion behaviors} \citep{Benabderrahmane21}. Additionally, clustering-based approaches, such as DBSCAN and hierarchical clustering, have been employed to group similar behaviors and detect outliers indicative of APT activity. However, clustering methods are highly sensitive to parameter tuning and often misclassify benign deviations as anomalies \citep{karim2025slf}. While these models can improve detection rates, they often struggle with class imbalance, lack of labeled data, and inconsistencies in feature representation across different attack environments.

Deep learning techniques have further advanced anomaly detection by learning complex feature representations directly from data \citep{dehghan2025proapt,cuong2025novel,benabderrahmane2025attackers}. Recurrent Neural Networks (RNNs), particularly Long Short-Term Memory (LSTM) networks, have been applied to sequential log data to detect anomalies in user and system behavior \citep{DBLP:journals/fgcs/BenabderrahmaneHVCR24, saeed2025enhanced}. Transformers and self-attention mechanisms have also been explored for capturing long-range dependencies in attack sequences \citep{BenabderrahmaneLLM}. Graph-based models, such as Graph Neural Networks (GNNs), have been leveraged to analyze entity relationships and detect lateral movement within compromised networks. While deep learning techniques show promise, their applicability to APT detection is hindered by the need for extensive labeled datasets, computational complexity, and their tendency to overfit to specific attack patterns \citep{jin2025pdcleaner}.

Despite these advancements, a fundamental challenge remains: most existing anomaly detection models lack transferability, meaning that models trained on one dataset often fail to generalize when applied to new, unseen attack scenarios. This limitation arises due to several key factors. First, cybersecurity datasets exhibit significant variability in their feature space, as different organizations log security events using diverse tools, leading to inconsistencies in recorded attack traces. Second, APTs are highly adaptive, modifying their techniques based on the target environment, which prevents static detection models from maintaining effectiveness across different contexts. Third, many anomaly detection models rely on supervised learning, which requires labeled datasets that are rarely available in cybersecurity, particularly for novel APT campaigns. Even when labeled data exists, the extreme class imbalance between benign and malicious instances further degrades model performance.

These challenges highlight the urgent need for advanced approaches that enhance model generalization across heterogeneous attack environments. Transfer learning provides a compelling solution by allowing models trained on one attack dataset to adapt their knowledge to new scenarios, mitigating the problem of dataset scarcity. However, simple transfer learning is often insufficient due to feature space misalignment between datasets, which can lead to suboptimal performance. To address this, contrastive learning can be leveraged to refine feature representations by pulling similar attack patterns closer while pushing dissimilar behaviors apart, thereby improving anomaly ranking. Additionally, Siamese networks play a crucial role in explicitly aligning feature spaces, ensuring that embeddings from different datasets remain comparable. This combination of transfer learning, contrastive learning, and feature space alignment represents a novel and effective strategy for improving APT detection across diverse and evolving cyber threats.
\section{Materials and Methods}
\subsection{Overview and Motivation}
To establish a strong proof of concept for the feasibility of transfer learning in Advanced Persistent Threat (APT) detection, it is essential to justify both the necessity and the effectiveness of transfer learning in this context. As explained before, one of our primary motivations for using transfer learning in APT detection is the scarcity of labeled data. APT attacks are rare, making it impractical to train machine learning models from scratch due to the limited availability of annotated datasets. Additionally, while APT campaigns differ based on their targets, tools, and techniques, there exist underlying attack patterns that can be transferred across different attack environments. However, a major challenge in cybersecurity datasets is feature space misalignment, where logs, monitoring systems, and extracted features differ significantly across datasets. Transfer learning provides a mechanism to bridge these differences, allowing models trained on one attack scenario to generalize to others.

From a practical standpoint, transfer learning offers significant advantages in real-world cybersecurity applications. One of its primary benefits is computational efficiency, as pre-trained models can be rapidly adapted to new attack scenarios without requiring full retraining. This substantially reduces training time and computational costs, making transfer learning a scalable solution for APT detection. Furthermore, as APTs continuously evolve, transfer learning allows models to be incrementally updated with new attack data, enhancing robustness against emerging threats. A timeline-based evaluation could illustrate how models trained on historical APT data remain effective in detecting newer attack variants. Lastly, transfer learning offers operational feasibility in Security Operations Centers (SOCs) by reducing reliance on manual rule updates and allowing seamless integration into Security Information and Event Management (SIEM) systems. This minimizes human intervention while improving the detection of evolving cyber threats.
\subsection{Global Architecture}
\begin{figure}[!th]
    \centering
    \includegraphics[width=\linewidth]{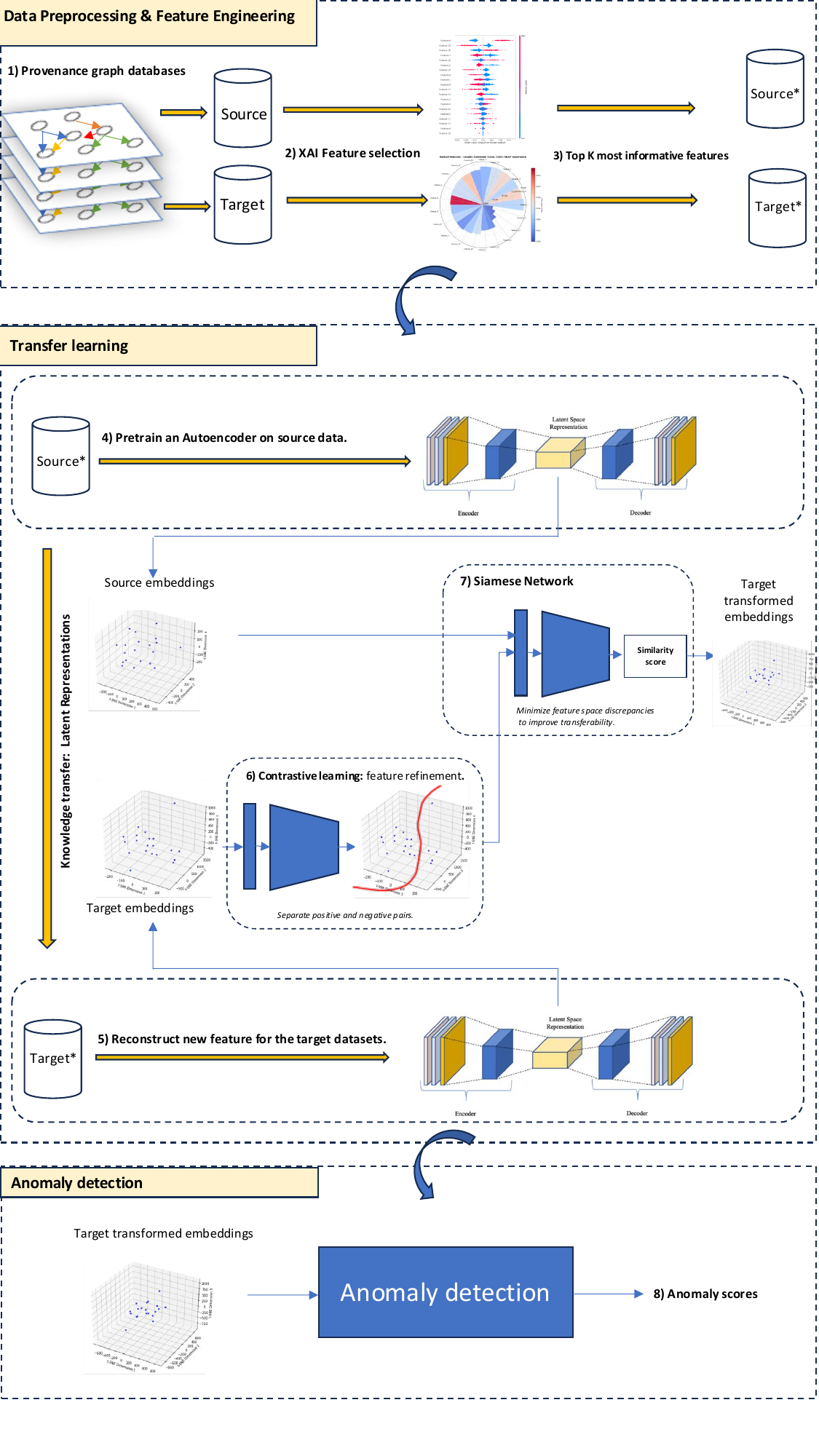}
    \caption{Pipeline for Transfer Learning in APT Detection: The framework begins with provenance graph databases and explainable AI (XAI) feature selection to extract the most informative features. An autoencoder is pretrained on source attack data to learn latent representations, which are then transferred to the target domain. Feature space refinement is achieved through contrastive learning, ensuring better separation between positive and negative pairs. A Siamese network further aligns source and target embeddings by minimizing feature space discrepancies. The final transformed embeddings are used for anomaly detection, producing anomaly scores that enhance cybersecurity threat identification.}
    \label{fig:pipeline}
\end{figure}

The global architecture of our framework is illustrated in Figure \ref{fig:pipeline}.
The pipeline integrates transfer learning, explainable feature selection, contrastive learning, and Siamese networks to enhance the detection of Advanced Persistent Threats (APTs) across different attack scenarios. Below in algorithm \ref{alg:pipeline}, we provide a detailed step-by-step breakdown of our methodology, including formal mathematical definitions where applicable.
%

\begin{algorithm}
\caption{Contrastive Transfer Learning with Siamese Networks for APT Detection}
\label{alg:pipeline}
\begin{algorithmic}[1]
    \State \textbf{Input:} Source dataset $\Ds$, Target dataset $\Dt$
    \State \textbf{Output:} Anomaly scores for $\Dt$
    \State \textbf{Step 1: Data Preprocessing }
    \State \hl{Load provenance graphs into data frames and vectorize into matrices $\mX_s \in \R^{N_s \times d}$ and $\mX_t \in \R^{N_t \times d}$.}
    \State Augment data with CGAN and VAE, producing \hl{$\widetilde{\mX}_s, \widetilde{\mX}_t$}.
   \State \textbf{Step 2: Feature Selection Using Explainable AI (XAI)}
    \State Compute per-feature scores $S_j$ for $j=1,\dots,d$:
    \begin{equation*}
  S_j \;=\; \alpha\, \mathrm{RE}_j \;+\; \beta\, \mathrm{Ent}_j \;+\; \gamma\, \mathrm{SHAP}_j
\end{equation*}
%
    \State \hl{Select the top-$K$ features and restrict $\mX_s,\mX_t$ to those columns.}

    \State \textbf{Step 3: Train Autoencoder on Source Dataset $\Ds$}
    \State \hl{Train AAE parameters $\vtheta_{\mathrm{enc}},\vtheta_{\mathrm{dec}}$ by minimizing}
    \[
      L_{\mathrm{AAE}} \;=\; \big\|\mX_s - \widehat{\mX}_s\big\|_F^2 \;=\; \sum_{i=1}^{N_s}\big\|\vx^{(i)}_s - \hat{\vx}^{(i)}_s\big\|_2^2,
    \]
    \Statex \hspace{1.5em} \hl{and extract latent embeddings $\mZ_s = f_{\mathrm{enc}}(\mX_s;\vtheta_{\mathrm{enc}})$ with rows $\vz^{(i)}_s$.}

    \State \textbf{Step 4: Transfer Learning to Target Dataset $\Dt$}
    \State Fine-tune $AAE$ on $\mX_t$ using transfer learning objective:
       \[
      \min_{\vtheta}\; \sum_{i=1}^{N_t}\big\|\vx^{(i)}_t-\hat{\vx}^{(i)}_t\big\|_2^2
      \;+\; \lambda\,\sum_{i=1}^{N_s}\big\|\vx^{(i)}_s-\hat{\vx}^{(i)}_s\big\|_2^2 .
    \]
    \State \hl{Obtain $\mZ_t = f_{\mathrm{enc}}(\mX_t;\vtheta_{\mathrm{enc}})$.}
    
    \State \textbf{Step 5: Contrastive Learning for Feature Space Refinement}
    \State Train contrastive loss to refine embeddings:
    \begin{equation*}
        L_{contrast} = - \sum_{i=1}^{N} \log \frac{ \exp(\text{sim}(Z_i, Z_j) / \tau) }{ \sum_{k=1}^{N} \exp(\text{sim}(Z_i, Z_k) / \tau) }
    \end{equation*}
    
    \State \textbf{Step 6: Feature Space Alignment Using Siamese Network}
    \State Train Siamese Network to align embeddings across datasets:
    $        L_{siamese} = (1 - y) \frac{1}{2} D^2 + y \frac{1}{2} \max(0, m - D)^2$
    \State Compute Siamese Distance to measure alignment

    \State \textbf{Step 7: Anomaly Detection on Target Dataset}
    \State Compute anomaly scores:
    $\mathrm{score}^{(i)} \;=\; \big\|\vx^{(i)}_t - \hat{\vx}^{(i)}_t\big\|_2$ 
    \State Apply anomaly detection models (Isolation Forest, One-Class SVM, Local Outlier Factor, KNN, DBSCAN)
    
    \State \textbf{Evaluation Metrics (nDCG, AUC)}
    \State Compute nDCG score: $\text{nDCG} = \frac{1}{Z} \sum_{i=1}^{p} \frac{2^{rel_i} - 1}{\log_2(i + 1)}$ 
    \State Compute AUC score for classification performance

    \State \textbf{Feature Space Visualization}
    \State Apply UMAP/t-SNE to visualize feature space alignment

    \State \textbf{Statistical Validation}
    \State Conduct statistical test to verify performance gains:
    
    \State If $p < 0.05$, conclude significant improvement.

    \State \textbf{Return:} Anomaly scores and evaluation results.
\end{algorithmic}
\end{algorithm}
\subsection*{\textbf{\hl{Step 1: Data Preprocessing and Feature Selection}}}
The first stage of our pipeline involves data preprocessing and feature engineering, where we transform raw cybersecurity datasets recorded as graph provenance traces, such as those from the DARPA Transparent Computing program. These datasets are structured as binary dataframes, where each row represents a unique process, and the columns correspond to specific system events (e.g., file accesses, network connections, privilege escalations). The presence or absence of an event for a given process is encoded as a binary value, creating a structured representation of process activity.

To endorse the analysis, we augment both source and target datasets with synthetic traces generated using cGANs and VAEs, ensuring a more diverse and representative set of attack scenarios. Conditional Generative Adversarial Networks (cGANs) operate by training a generator to create synthetic attack traces conditioned on specific labels or constraints, ensuring realistic variations that resemble real-world threats. Meanwhile, Variational Autoencoders (VAEs) generate additional attack instances by learning a latent representation of the existing data and sampling from this distribution to create new examples, effectively introducing variability while maintaining the statistical properties of the original dataset. This augmentation strategy helps mitigate data scarcity issues and enhances the robustness of the learning model.

The first stage then begins by loading the raw logs and converting them into a format suitable for analysis. Since different datasets may have varying structures and levels of granularity, feature extraction techniques are applied to ensure consistency across sources. Additionally, to enhance model performance, we normalize the binary feature space, ensuring that the data is well-prepared for subsequent learning tasks.
\subsection*{\textbf{\hl{Step 2: Explainable Feature Selection Using XAI Techniques}}}
To optimize transfer learning, we introduce feature selection using SHAP (Shapley Additive Explanations) values and entropy-based scoring. A key aspect of this process is the use of a surrogate autoencoder, which is specifically trained to compute reconstruction errors for feature selection. Unlike the primary autoencoder used for transfer learning, the surrogate autoencoder is trained independently to capture the reconstruction difficulty of each feature, serving as a proxy for feature importance.

The importance score for each feature $j$ is computed as:

\begin{equation} 
  S_j \;=\; \alpha\, \mathrm{RE}_j \;+\; \beta\, \mathrm{Ent}_j \;+\; \gamma\, \mathrm{SHAP}_j
\end{equation}

where the reconstruction error \hl{$\mathrm{RE}$} is derived from the surrogate autoencoder, reflecting how well each feature is reconstructed by the learned latent representations. Features with higher reconstruction errors indicate that they contain unique or difficult-to-model information, making them potentially valuable for anomaly detection.

Entropy \hl{$\mathrm{Ent}$} is further incorporated into the scoring process to quantify the variability of each feature and capture rare events that might be potentially anomalous:

\begin{equation} Entropy(x_j) = -\sum p(x_j) \log p(x_j) \end{equation}

The coefficients $\alpha$, $\beta$, and $\gamma$ control the relative contribution of each criterion: the reconstruction error from the surrogate autoencoder, entropy-based feature variability, and SHAP-based explainability, respectively. In our implementation, we set the default weights to $\alpha$=0.5, $\beta$=0.5, and $\gamma$=0.5, ensuring equal consideration of each factor in the feature selection process. However, these values are not fixed, and they can be tuned based on the specific characteristics of the dataset or the operational requirements of the detection system. Adjusting these parameters allows for a flexible calibration of the algorithm, where a higher $\alpha$ value would prioritize features that are difficult to reconstruct, a higher $\beta$ would emphasize features with greater variability, and a higher $\gamma$ would favor features identified as important by the SHAP model. This adaptability ensures that the feature selection process remains robust across different attack scenarios and cybersecurity environments.\\
By integrating reconstruction errors from the surrogate autoencoder, entropy measures, and SHAP-based explainability, we ensure that only the most informative and discriminative features are retained. We retain only the top $K$ features to reduce computational cost and improve transferability, ensuring that the model focuses on the most relevant attributes while discarding redundant or noisy dimensions. To determine the optimal number of selected features $K$, we employ a \emph{Cumulative Contribution Method}, ensuring that the most informative features are retained while filtering out less relevant ones. We define $K$ such that the cumulative contribution of the selected features exceeds a predefined percentage $\xi$ (e.g. between 60\% to 95\% of the total feature importance). Formally, this is expressed as:
\begin{equation}
\sum_{j=1}^{K} S_j \geq \xi \sum_{j=1}^{d} S_j
\end{equation}
where:
$S_j$ represents the feature importance score of the $j$-th feature. $d$ is the total number of features in the dataset.

The threshold $\xi$ (e.g., 60\%) ensures that the majority of the dataset's informative content is preserved.
This method provides an adaptive mechanism for feature selection, ensuring that only the most significant features are retained while maintaining a balance between computational efficiency and detection performance.

This selection is driven by an Explainable AI (XAI)-based feature importance framework, which enhances model interpretability, facilitates knowledge transfer between domains, and mitigates feature space inconsistencies. By leveraging a dedicated surrogate autoencoder for feature selection, we ensure that the chosen features not only improve the efficiency of transfer learning but also enhance the robustness and scalability of APT detection.
\subsection*{\textbf{\hl{Step 3: Pretraining the AutoEncoder on Source Dataset}}}

The next stage of our pipeline focuses on transfer learning using an attention-based autoencoder (AAE) to effectively adapt knowledge from one attack domain to another. It begins by pretraining an Attention-Enhanced Autoencoder on the source dataset. The autoencoder is designed to extract latent representations that capture essential patterns and structures within the attack data. Once the pretraining phase is completed, the model is fine-tuned on the target datasets (including real and synthetic attack traces generated by cGANs and VAEs). This fine-tuning process ensures that the autoencoder adapts to variations in attack behaviors while leveraging prior knowledge from the source dataset. Finally, we use the trained autoencoder to reconstruct new feature embeddings for the target datasets, effectively transferring learned knowledge across domains while maintaining the integrity of critical attack patterns.

The objective behind training the \emph{attention-based autoencoder} on a source attack dataset is to extract meaningful representations of normal and anomalous behavior. 

The Attention-Based Autoencoder (AAE) is an enhanced version of a baseline autoencoder that incorporates an attention mechanism to selectively focus on the most relevant features in the input space. This improves feature extraction, making it particularly useful in transfer learning for APT detection.

Basically, the autoencoder consists of two primary components: an encoder $f_{enc}$ that compresses input data into a lower-dimensional representation and a decoder $f_{dec}$ that reconstructs the original input from the latent space.

Given an input source sample \hl{$\mX_s \in \mathbb{R}^{d}$}, the autoencoder learns a compressed latent representation $\mZ_s$ as follows:

\begin{equation}
\mZ_s = f_{enc}(\mX_s;\vtheta_{\mathrm{enc}}) = \sigma(\mW_{enc}\mX_s + b_{enc})
\end{equation}

where:
\begin{itemize}
\item \hl{$\mW_{enc} \in \mathbb{R}^{h \times d}$ and $b_{enc} \in \mathbb{R}^{h}$ are the encoder weight matrix and bias.}
\item $h$ is the dimension of the latent space (h$\ll$d).
\item $\sigma(\cdot)$ is the activation function ( ReLU).
\end{itemize}

The decoder reconstructs the input from $Z$:

\begin{equation}
\hat{X_s} = f_{dec}(\mZ_s;\vtheta_{\mathrm{dec}}) = \sigma(W_{dec}\mZ_s + b_{dec})
\end{equation}

where\hl{ $\mW_{dec} \in \mathbb{R}^{d \times h}$ and $b_{dec} \in \mathbb{R}^{d}$ are the decoder parameters.}

The autoencoder is trained by minimizing the reconstruction loss:

\begin{equation}
L_{AE} = \sum_{i=1}^{N} | x_s^{(i)} - \hat{x_s}^{(i)} |^2
\end{equation}

where $N$ is the number of training samples.

To enhance transferability, we use an attention mechanism that assigns different importance scores to each feature before encoding. The attention mechanism learns a set of attention weights $\alpha$ such that:

\begin{equation}
\alpha_j = \frac{\exp(\mW_{att} \cdot X_j)}{\sum_{k=1}^{d} \exp(\mW_{att} \cdot X_k)}
\end{equation}

where:
\begin{itemize}
\item \hl{$\mW_{att}$ is a learnable parameter matrix for attention scoring.}
\item $\alpha_j$ represents the importance of feature $j$, ensuring that informative features contribute more to the latent space.
\end{itemize}

The attended input is then computed as:

\begin{equation}
X_{att} = \alpha \odot X_s
\end{equation}

where $\odot$ represents the element-wise multiplication. The encoder now processes $\mX_{att}$ instead of $\mX$:

\begin{equation}
\mZ_s = f_{enc}(\mX_{att};\vtheta_{\mathrm{enc}})
\end{equation}

This modified encoding process allows the model to focus on the most relevant features, improving the efficiency of knowledge transfer across attack scenarios.

The final objective function includes both reconstruction loss and attention regularization:

\begin{equation}
L_{AAE} = L_{AE} + \lambda \sum_{j=1}^{d} |\alpha_j - \frac{1}{d}|^2
\end{equation}

where $\lambda$ controls the regularization strength, ensuring that attention scores do not collapse to trivial values.

By incorporating attention, the autoencoder improves feature representation quality, ensuring better transfer learning performance for APT detection.

%
%
%
%
%
%
\subsection*{\textbf{\hl{Step 4: Transfer Learning for Knowledge Adaptation}}}
In APT detection, transfer learning is crucial due to:
\begin{itemize}
\item Feature Space Misalignment: Different cybersecurity datasets have distinct log formats and extracted features.
\item Lack of Labeled Data: APT attacks are rare, making fully supervised learning impractical.
\item Targeted Nature of APTs: Attack strategies depend on the victim's infrastructure, causing distribution shifts between datasets.
\end{itemize}
In transfer learning, we have two domains: the source domain ($\mathcal{D}_s$) where the model is initially trained, and the target domain ($\mathcal{D}_t$) where we apply the learned knowledge. Each domain consists of:
\begin{itemize}
\item \hl{A feature space (or variables) ($\mX_s$ or $\mX_t$) representing the extracted features.}
\item \hl{A probability distribution ($P(\mX_s)$ or $P(\mX_t)$) describing how data points are distributed in each domain, which the model aims to learn.}
\item A learning task ($\mathcal{T}_s$ or $\mathcal{T}_t$), which includes:
\begin{itemize}
\item A label space ($Y_s$ or $Y_t$), representing possible outputs (e.g., benign vs. malicious processes).
\item A task function ($f_s(\cdot)$ or $f_t(\cdot)$), the mapping from features to labels.
\end{itemize}
\end{itemize}

By breaking down the above definitions, the \textit{source domain} is defined as: 
\begin{equation}
    \mathcal{D}_s = \{\mX_s, P(\mX_s)\}
\end{equation}

This represents the domain where we train the attention autoencoder model. It consists of:
\begin{itemize}
    \item  \hl{$\mX_s$} (Source Feature Space):
    The set of features describing the data in the source domain.
    In our case (APT detection), \hl{$X_s$} represents a binary feature space where each row corresponds to a process, and each column represents an action (event) the process performed.
    \item \hl{$P(\mX_s)$} (Source Data Distribution):
    The probability distribution over the source feature space.
    This describes how often certain patterns appear in the source dataset (e.g., frequency of file access events in the dataset).
\end{itemize}

The second important parameter is the \textit{source learning task} defined as:
\begin{equation}
    \mathcal{T}_s = \{Y_s, f_s(\cdot)\}
\end{equation}

This defines the learning objective in the source domain:
\begin{itemize}
    \item $Y_s$ (Source Label Space):
    The possible output labels in the source dataset.
    In our case, since we have a binary anomaly classification, $Y_s$ could be: {Anomalous (1), Normal (0)}.
        \item $f_s(\cdot)$ (Source Task Function):
    The model function trained on the source dataset to predict labels.
    In our case, $f_s(\cdot)$ is the AAE autoencoder’s reconstruction loss, which learns to detect anomalies in the source dataset.
\end{itemize}
    
Similarly, the target domain and task are defined as follows:
\begin{align}
\mathcal{D}_t &= \{\mX_t, P(\mX_t)\}, &\mathcal{T}_t &= \{Y_t, f_t(\cdot)\}
\end{align}

This represents the new domain where we apply transfer learning. It consists of:
\begin{itemize}
    \item $\mX_t$ (Target Feature Space):
The set of features describing data in the target domain. 
This feature space is often similar but not necessarily identical to $\mX_s$, meaning: 
\begin{itemize}
    \item The features may be slightly different due to logging variations (e.g., different OS environments).
    \item The distribution of values may be different (e.g., different attack frequencies).
\end{itemize}
    \item $P(\mX_t)$ (Target Data Distribution):
The probability distribution over the target feature space.
\begin{itemize}
    \item    If the data distribution changes significantly, direct application of $f_s(\cdot)$ may fail.
  \item  Our goal in transfer learning is to align $P(\mX_s)$ and $P(\mX_t)$ so that knowledge can be transferred.
\end{itemize}
\end{itemize}
 
The target Task $\mathcal{T}_t = \{Y_t, f_t(\cdot)\}$ defines the learning objective in the target domain:
\begin{itemize}
    \item  $Y_t$ (Target Label Space):
    The labels in the target domain, which may differ slightly from the source domain.
        For example, in the target, we may have different types of anomaly compared to the source.
        \item $f_t(\cdot)$ (Target Task Function):
    The function we want to learn in the target domain.
    \begin{itemize}
        \item  If $f_s(\cdot) \neq f_t(\cdot)$, direct application of the source model could fail.
        \item  Transfer learning ensures that the knowledge from $f_s(\cdot)$ is adapted to help learn $f_t(\cdot)$ efficiently.
    \end{itemize}
\end{itemize}

%
The goal of Transfer Learning is to improve learning in $\mathcal{D}_t$ using knowledge from $\mathcal{D}_s$, even when:
\begin{itemize}
\item The feature spaces are different: $\mX_s \neq \mX_t$.
\item The data distributions differ: $P(\mX_s) \neq P(\mX_t)$.
\item The learning tasks are not identical: $f_s(\cdot) \neq f_t(\cdot)$.
\end{itemize}

Mathematically, this is formulated as minimizing the generalization error in the target domain:
\begin{equation}
\min_{\theta} \sum_{(x_t, y_t) \in \mathcal{D}t} L(f_t(x_t; \vtheta), y_t) + \lambda \sum_{(x_s, y_s) \in \mathcal{D}_s} L(f_s(x_s; \vtheta), y_s)
\end{equation}
where:
\begin{itemize}
\item $L(\cdot)$ is a loss function (AAE reconstruction loss).
\item $\vtheta$ represents the model parameters.
\item $\lambda$ controls the contribution of the source knowledge.
\end{itemize}
%
%
%
%
%
%
%
%
%
%
%
%
%
\subsection*{\textbf{\hl{Step 5: Contrastive Learning for Feature Space Refinement}}}
\begin{figure}[h!]
    \centering
    \includegraphics[width=1\linewidth]{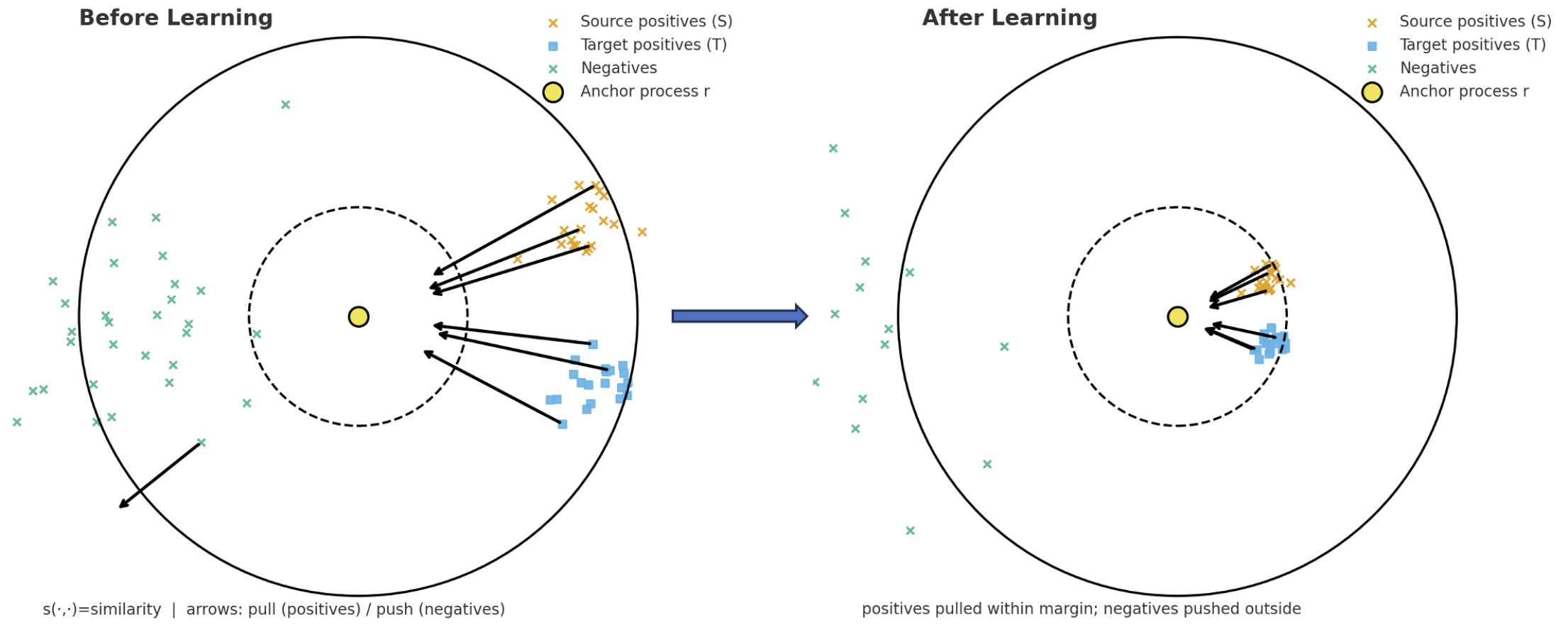}
    \caption{Principle of contrastive learning: 
For an anchor process $r$ (yellow), source and target positives (orange x and blue $\blacksquare$) are pulled closer while negatives (green ×) are pushed away. Before learning (left), embeddings are dispersed; after optimizing a contrastive loss (right), positives are pulled within a similarity margin (dashed circle) and negatives are driven outside the margin, reducing intra-class distance, increasing inter-class separation, and aligning source–target representations for improved anomaly ranking.}
    \label{fig:contrastive}
\end{figure}

To refine feature representations and enhance the transferability of learned embeddings, we incorporate \emph{Contrastive Learning} mechanism. \hl{As illustrated in Figure} \ref{fig:contrastive}, this technique ensures that embeddings from similar attack instances remain close in the latent space while pushing apart representations of dissimilar samples. Such an approach is crucial in APT detection, where subtle variations in attack behaviors can lead to misclassification if not properly aligned.

\paragraph{Contrastive Loss Formulation:} 
Given the encoded feature space $\mathcal{Z}$, we define a set of anchor-positive pairs $(Z_i, Z_j)$, where $Z_i$ and $Z_j$ are feature embeddings of instances belonging to the same attack type or behavioral cluster. Similarly, we define anchor-negative pairs $(Z_i, Z_k)$, where $Z_k$ belongs to a different category (e.g., benign vs. malicious or different APT campaigns). The contrastive loss function, is given by:

\begin{equation} L_{\text{contrast}} = - \sum_{i=1}^{N} \log \frac{ \exp(\text{sim}(Z_i, Z_j) / \tau) }{ \sum_{k=1}^{N} \exp(\text{sim}(Z_i, Z_k) / \tau) } \end{equation}

where:
\begin{itemize}
    \item  $\text{sim}(Z_i, Z_j)$ represents a similarity function (e.g., cosine similarity), \begin{equation} \text{sim}(Z_i, Z_j) = \frac{Z_i \cdot Z_j}{|Z_i| |Z_j|} \end{equation}.
    \item $\tau$ is a temperature scaling parameter that controls the sharpness of similarity distributions (here $\tau = 0.1$).
    \item The denominator sums over all possible negative pairs, enforcing discrimination between similar and dissimilar feature representations.
\end{itemize}
   
\paragraph{Contrastive Learning in Transfer Learning for APT Detection:}

In the context of APT detection with transfer learning, contrastive learning plays a key role in ensuring that knowledge from the source dataset is efficiently adapted to the target dataset.\\ Specifically, we apply contrastive learning to:
\begin{itemize}
    \item Align Embeddings Across Domains: By leveraging contrastive loss, the model learns a structured latent space where embeddings from similar attack patterns in the source and target domains are positioned closer together.
    \item Improve Feature Generalization: The method enhances feature robustness, ensuring that the model can distinguish between benign processes and anomalies across different datasets.
    \item  Mitigate Catastrophic Forgetting: In scenarios where new data is incrementally introduced, contrastive learning prevents the model from entirely forgetting previously learned representations.
\end{itemize}

\paragraph{Hard Negative Mining:}
To further improve feature space alignment, we introduce hard negative mining, which focuses training on instances where the model is most uncertain. Instead of randomly selecting negative examples, we prioritize cases where:

\begin{equation} \text{sim}(Z_i, Z_k) > \delta \end{equation}

where $\delta$ is a threshold for similarity between an anchor and a negative sample (here $\delta = 0.8$). If a negative sample is too similar to the anchor, it is more informative for training since it forces the model to learn finer-grained distinctions.
%
%
%
%
%
%
%
%
\subsection*{\textbf{\hl{Step 6: Feature Space Alignment Using Siamese Networks}}}
In the context of APT detection and transfer learning, a major challenge arises from the feature space misalignment between the source and target domains. This misalignment occurs due to differences in logging mechanisms, monitoring tools, and attack behaviors recorded across different datasets. If not properly addressed, such inconsistencies degrade the performance of machine learning models when transferring knowledge from one attack domain to another.

\begin{figure}[h!]
    \centering
    \includegraphics[width=1\linewidth]{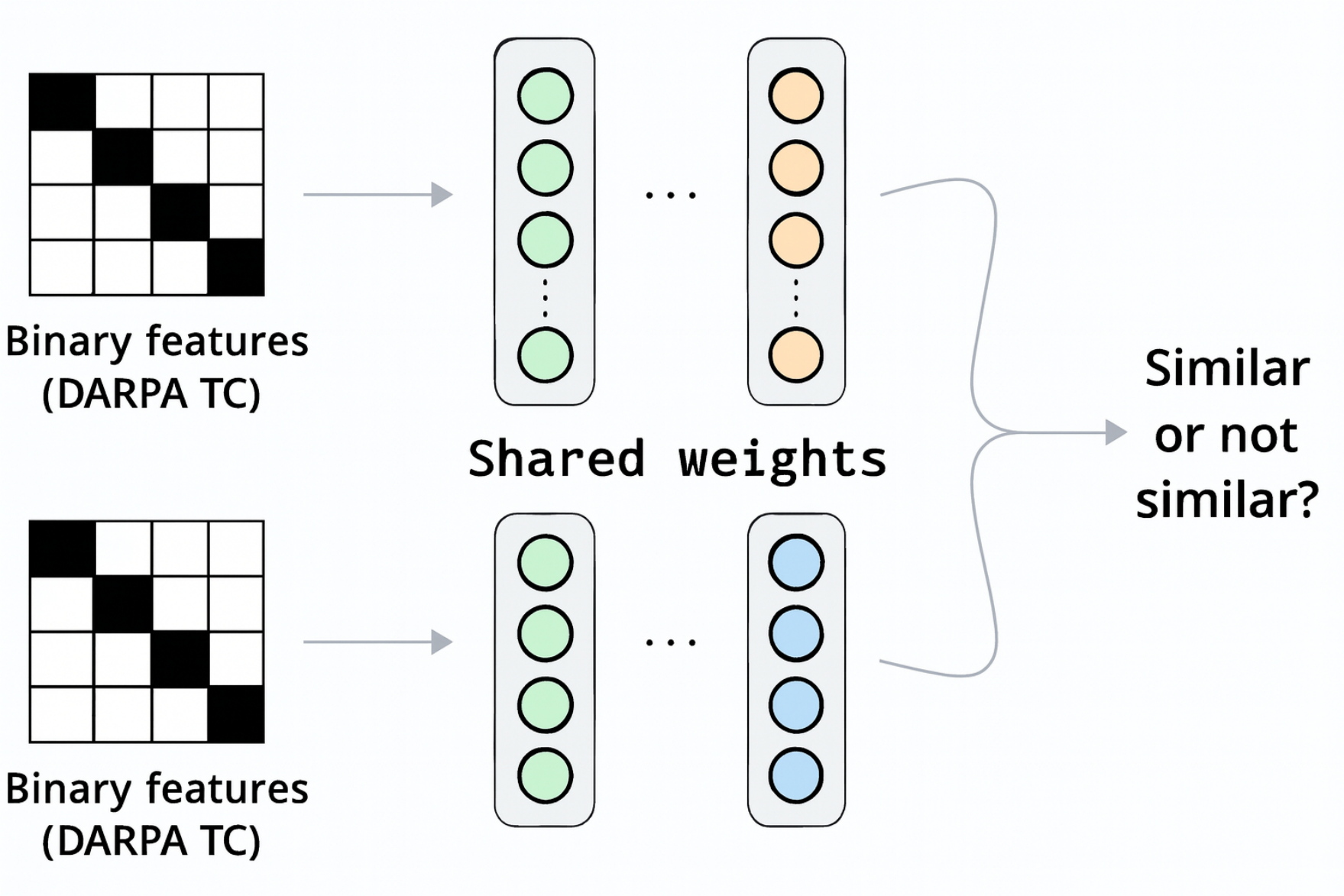}
    \caption{\hl{Siamese Network Architecture for Transfer Learning in APT Detection.
Two input feature vectors are processed through identical neural network branches with shared weights. The resulting embeddings are compared using a similarity function to determine whether the inputs exhibit similar behavior, enabling cross-domain generalization for anomaly or threat detection.}}
    \label{fig:siamese}
\end{figure}

To mitigate this issue, we employ Siamese Networks, a type of neural network architecture designed to measure similarity between two inputs (Figure \ref{fig:siamese}). Unlike traditional neural networks, Siamese Networks do not focus on classifying individual samples. Instead, they learn a distance function that quantifies how similar or different two input feature representations are. This property makes them highly effective in feature space alignment for transfer learning, as they enable the model to adjust embeddings from different datasets to be more coherent and comparable.
\paragraph{Siamese Network Architecture:}
A Siamese Network consists of two identical subnetworks (parameter-sharing deep neural networks), each processing one of two input samples. Given two input feature vectors, $x_1$ from the source dataset and $x_2$ from the target dataset, the network generates their respective embeddings, $h_1$ and $h_2$, using a shared encoding function $f_{\theta}$:

\begin{equation} h_1 = f_{\theta}(x_1), \quad h_2 = f_{\theta}(x_2) \end{equation}

where $f_{\theta}$ represents a shared neural encoder parameterized by $\theta$. The similarity or distance between the two embeddings is then computed using the Euclidean distance:

\begin{equation} D(x_1, x_2) = | h_1 - h_2 |_2 \end{equation}

This distance measure serves as the basis for optimizing feature space alignment between the source and target domains.

\paragraph{Contrastive Loss for Siamese Networks:}
The training objective for Siamese Networks is to ensure that similar samples (e.g., attacks from different datasets with similar behavior) have small distances, whereas dissimilar samples have large distances. To achieve this, we use the contrastive loss function, which is formulated as follows:

\begin{equation*} L_{\text{siamese}} = (1 - y) \frac{1}{2} D^2 + y \frac{1}{2} \max(0, m - D)^2 \end{equation*}

where:
\begin{itemize}
    \item $D$ is the Euclidean distance between the two feature embeddings.
    \item $y$ is the label indicating whether the two samples belong to the same class ($y=0$ for similar pairs, $y=1$ for dissimilar pairs).
    \item $m$ is a margin parameter that defines the minimum separation between dissimilar pairs to prevent trivial solutions.
    \item The first term minimizes the distance for similar pairs, while the second term ensures that dissimilar pairs remain separated by at least $m$.
\end{itemize}
By minimizing this loss, the model gradually adjusts the feature spaces of the source and target datasets, improving alignment and reducing domain shift.

\paragraph{Role of Siamese Networks in Transfer Learning for APT Detection:}
In our APT detection pipeline, Siamese Networks serve three critical roles:
\begin{itemize}
    \item Feature Space Normalization Across Datasets: The network forces the embeddings from different datasets to become comparable, enabling the anomaly detection models to generalize across multiple attack environments.
    \item  Knowledge Transfer Optimization: By refining the latent space, Siamese Networks help improve the effectiveness of transfer learning, ensuring that models trained on the source dataset can be successfully applied to the target dataset.
    \item   Robustness to Dataset Variability: Since APT datasets often have imbalanced distributions, using Siamese Networks allows us to learn a similarity function that is less sensitive to feature noise and variations between training and testing distributions.
\end{itemize}

\paragraph{Joint Training with Autoencoder and Contrastive Learning:}
Our framework integrates Siamese Networks with autoencoder embeddings and contrastive learning to enhance feature alignment. The training is performed jointly with the following components:
\begin{itemize}
    \item Autoencoder Embeddings ($Z_s$ and $Z_t$): The Siamese Network operates on latent feature embeddings extracted by the attention-based autoencoder, rather than raw feature vectors, ensuring that only meaningful features are compared.
\item Contrastive Learning for Distance Optimization: The Siamese Network works alongside contrastive learning, which refines the decision boundaries by pulling similar instances closer together and pushing dissimilar ones apart.

\item XAI-Guided Feature Selection: To ensure the alignment process focuses on the most informative features, we apply feature selection using SHAP values, entropy measures, and reconstruction errors, removing redundant or irrelevant features.
\end{itemize}

 The final optimization function for training the entire pipeline is:

\begin{equation} L_{\text{total}} = \lambda_1 L_{\text{contrast}} + \lambda_2 L_{\text{AAE}} + \lambda_3 L_{\text{siamese}} \end{equation}

where:
\begin{itemize}
    \item $L_{\text{contrast}}$ improves the separability of anomalies in feature space.
\item $L_{\text{AAE}}$ ensures that extracted feature embeddings retain relevant information.
\item $L_{\text{siamese}}$ explicitly optimizes feature space alignment between source and target datasets.
\item To ensure all three loss terms are comparable in scale, we set $\lambda_1$, $\lambda_2$, $\lambda_3$ to 1.
\end{itemize}

\paragraph{Performance Impact on APT Detection:}

The integration of Siamese Networks within our framework results in:
\begin{itemize}
    \item Improved Transferability: Models trained on one dataset generalize better to unseen attack scenarios.
    \item   Reduced False Positives: Feature space alignment ensures that anomalies are more accurately ranked.
    \item   Better Feature Representation: The joint optimization framework allows the model to extract transferable, interpretable, and semantically meaningful attack features.
\end{itemize}

By leveraging Siamese Networks, we provide a scalable and explainable approach to APT detection, ensuring that machine learning models remain effective across evolving cyber threats.
\subsection*{\textbf{\hl{Step 7: Anomaly Detection on Target Dataset}}}
\hl{After knowledge transfer and feature space alignment, the anomaly score for each process in our model is computed using the per-sample reconstruction error:}

\begin{equation}
\text{Anomaly Score}^{(i)} = \|\vx^{(i)}_t - \hat{\vx}^{(i)}_t\|_2
\end{equation}

\hl{where $\vx^{(i)}_t$ represents the original feature vector of the $i$-th process in the target dataset and $\hat{\vx}^{(i)}_t$ denotes its reconstruction generated by the transferred autoencoder.

Specifically, $\vx^{(i)}_t$ corresponds to the input raw feature vector of a target dataset process, while $\hat{\vx}^{(i)}_t$ corresponds to its reconstruction from the latent representations learned via the autoencoder pre-trained on the source domain and fine-tuned on the target domain. The magnitude of the reconstruction error for each data point thus quantifies its anomalousness in relation to patterns learned previously.}
\subsection{\texorpdfstring{\hl{Comparison with baseline methods:}}{Comparison with baseline methods:}}
We benchmark our approach against widely used classical anomaly detectors and three representative deep learning baselines. \textit{Isolation Forest}\citep{xue2025novel} isolates anomalies via random partitioning, where anomalous processes are separated with fewer splits. \textit{One-Class SVM}\citep{liu2025boosting} learns a tight boundary around normality and flags points outside it, a common one-class baseline for rare APT behaviors. \textit{Local Outlier Factor (LOF)}\citep{gholizade2025review} detects local density deviations indicative of stealthy, neighborhood-level anomalies. \textit{K-Nearest Neighbors (KNN)}\citep{nagendrudu2025detection} anomaly scoring uses distances to the $k$-th neighbor to flag unusual processes. \textit{DBSCAN}\citep{xavier2025trunc} clusters by density and marks nonclustered points as outliers—useful for capturing isolated attack traces. \hl{\textit{Deep SVDD}}\footnote{\url{https://github.com/lukasruff/Deep-SVDD-PyTorch}}~\citep{pmlr-v80-ruff18a} is an unsupervised deep one-class method that maps embeddings into a compact hypersphere; anomalies are those far from the center. \hl{\textit{DevNet}}\footnote{\url{https://github.com/GuansongPang/deviation-network}}~\citep{devnet19} is a supervised deep anomaly detector trained with a deviation-based loss to enlarge the margin between normal and anomalous samples when limited labels exist. Finally, a \hl{\textit{baseline Autoencoder}}~\citep{DBLP:journals/fgcs/BenabderrahmaneHVCR24} models per-process event sequences reconstructed from provenance; anomalies are detected via reconstruction error, providing a temporal deep baseline complementary to representation learning.
%
\subsection{\texorpdfstring{\hl{Evaluation using nDCG and AUC:}}{Evaluation using nDCG and AUC:}}
To measure performance, we use \textit{nDCG} scores to assess anomaly ranking and \textit{AUC} to evaluate classification. Normalized Discounted Cumulative Gain (nDCG) is a widely-used ranking metric originally designed to evaluate information retrieval systems by assessing how effectively ranked results match the expected relevance. In anomaly detection, especially in cybersecurity contexts such as Advanced Persistent Threat detection, accurately ranking the severity or likelihood of anomalies is often more critical than merely binary classification performance. Unlike conventional metrics, such as the Area Under the Curve (AUC), which only measure binary classification effectiveness, nDCG considers the relative positions of anomalies in a ranked list, giving higher importance to anomalies detected earlier. Formally, nDCG computes the cumulative gain (CG) by summing the graded relevance scores of detected anomalies, discounted by their logarithmic position in the ranked list, and normalizes this by the ideal discounted cumulative gain (IDCG). Therefore, higher nDCG values indicate that the model effectively prioritizes significant anomalies at the top ranks. This property makes nDCG particularly suitable for evaluating anomaly detection in cybersecurity scenarios, where the critical objective is not merely classification accuracy but rather ensuring the most severe and actionable anomalies are quickly identified for timely mitigation.

The nDCG score is defined as:

\begin{equation}
\text{nDCG} = \frac{1}{Z} \sum_{i=1}^{p} \frac{2^{rel_i} - 1}{\log_2(i + 1)}
\end{equation}
\hl{where $p$ is the number of top-ranked items considered in the evaluation, $rel_i$ is the graded relevance score of the item at position $i$, and $Z$ is a normalization constant representing the Ideal DCG (IDCG) for perfect ranking. Higher nDCG values indicate that the model effectively prioritizes significant anomalies at the top ranks, which is crucial for cybersecurity applications where timely detection of severe threats is paramount.}
\subsection{\texorpdfstring{\hl{Complexity Analysis}}{Complexity Analysis}}
\label{sec:complexity}
\hl{Below we quantify the asymptotic costs of our method, specifying per-stage time/memory requirements and highlighting how implementation choices (batch size, negatives, indexing) affect runtime in practice, with $N_s,N_t$ the numbers of source/target samples, $d$ features,
$h$ latent size, $E$ epochs, $B$ batch size, and $K$ selected features, and $M$ mined pairs per epoch (Siamese; includes hard negatives).}

\medskip
\noindent\hl{\textit{Step 1 — Preprocessing.}
Vectorization/normalization is $O(N_s d)$ on the source and $O(N_t d)$ on the target.}

\medskip
\noindent\hl{\textit{Step 2 — XAI Feature Selection.}
We compute per-feature scores combining a surrogate autoencoder reconstruction error, entropy, and SHAP:}
\begin{itemize}
    \item \hl{Surrogate AE passes over the source: $O(E N_s d)$.}
    \item \hl{Entropy per feature: $O(N_s d)$.}
    \item \hl{SHAP with a lightweight surrogate (tree/kernel) typically $O(N_s d)$ to $O(N_s d \log d)$, depending on the explainer.}
    \item \hl{Ranking and selecting top-$K$: $O(d \log d)$.}
\end{itemize}

\medskip
\noindent\hl{\textit{Step 3 — Attention-based Autoencoder (pretrain on $D_s$).}
With dense layers, a forward+backward pass per epoch is $O(N_s d h)$, hence pretraining costs $O(E N_s d h)$.}

\medskip
\noindent\hl{\textit{Step 4 — Transfer to $D_t$.}
Fine-tuning on the target mirrors Step 3 with $N_t$: $O(E N_t d h)$.
}

\medskip
\noindent\hl{\textit{Step 5 — Contrastive Learning (InfoNCE).}
With batch size $B$, computing all pairwise similarities in a batch is $O(B^2 h)$ per batch.
Across $E$ epochs and $N_t/B$ batches per epoch:}
\[
O\!\left(E \cdot \tfrac{N_t}{B} \cdot B^2 h\right) = O(E N_t B h).
\]

\medskip
\noindent\hl{\textit{Step 6 — Siamese Alignment.}
A forward distance for a pair is $O(h)$; with $M$ mined pairs per epoch (including hard negatives), the training cost is $O(E M h)$.}

\medskip
\noindent\hl{\textit{Step 7 — Anomaly Scoring.}
Per-sample reconstruction error $\|X_t - \hat{X}_t\|_2$ is $O(d)$, hence $O(N_t d)$ for all target samples.}


\medskip
\noindent\hl{\textit{Memory.}
AAE/contrastive/siamese store activations $O(B h)$ and parameters $O(d h + h^2)$.
}

\medskip
\subsection{\texorpdfstring{\hl{Feature Embeddings Visualization:}}{Feature Embeddings Visualization:}}
We visualize feature embeddings using \textit{t-SNE and UMAP} to verify transfer effectiveness. By projecting high-dimensional latent representations into two-dimensional or three-dimensional spaces, these visualization techniques allow us to qualitatively assess the alignment and overlap of source and target dataset embeddings before and after transfer learning. Improved clustering or proximity of source and target data points after the transfer indicates effective feature space alignment and enhanced generalization capability across different APT scenarios, providing intuitive validation of our method's effectiveness.
\subsection{\texorpdfstring{\hl{Experimental Setup for Validating Transfer Learning Effectiveness:}}{Experimental Setup for Validating Transfer Learning Effectiveness:}}

%
To thoroughly assess the effectiveness of our proposed APT detection framework, we define four distinct evaluation protocols, capturing both in-domain and cross-domain generalization scenarios:

\begin{itemize}
    \item \emp{Protocol $P_{0}$ — In-Domain Evaluation (No Transfer, Same Domain):} \\
    Each anomaly detection method, including our proposed Attention-based Autoencoder (AAE), is trained and tested on the \emph{same source dataset}. This protocol captures the model’s performance in an ideal setting where no domain shift occurs and serves as a baseline for in-domain detection.

    \item \emp{Protocol $P_{1}$ — Cross-Domain Evaluation (No Transfer, Direct Testing):} \\
    Methods are trained on the source dataset and directly evaluated on the target dataset \emph{without any adaptation}. This protocol measures the ability of models to generalize across domains in the absence of transfer learning mechanisms.

    \item \emp{Protocol $P_{2}$ — Cross-Domain with Method-Specific Adaptation (Traditional Transfer):} \\
    This protocol introduces method-specific transfer learning by fine-tuning model parameters on the target dataset. For classical anomaly detection methods (e.g., Isolation Forest, DBSCAN, OC-SVM), this involves adjusting hyperparameters (e.g., number of trees, contamination ratio, kernel types, or neighborhood sizes). \\
    \textit{For our proposed AAE model, this serves as an ablation study}:
    \begin{itemize}
        \item SHAP- and entropy-based feature selection is \emp{not applied};
        \item Siamese network-based feature space alignment is \emp{excluded};
        \item The model is regularized (e.g., with dropout) and fine-tuned directly on the target dataset.
    \end{itemize}

    \item \emp{Protocol $P_{3}$ — Full Proposed Transfer Framework:} \\
    This protocol implements the complete transfer learning pipeline:
    \begin{itemize}
        \item Train the AAE on the source dataset;
        \item Apply explainable feature selection using SHAP and entropy;
        \item Align source and target feature spaces via Siamese networks;
        \item Perform anomaly detection using Isolation Forest, OC-SVM, LOF, KNN, DBSCAN, Deep SVDD, DevNet, AE and our AAE on the resulting latent embeddings.
    \end{itemize}
\end{itemize}

The four evaluation protocols are summarized in Table \ref{tab:evaluation_scenarios}.

\begin{table*}
\small
\centering
\caption{Detailed evaluation protocols for anomaly detection methods including the proposed Attention-based Autoencoder (AAE).}
\resizebox{\textwidth}{!}{
\begin{tabular}{@{}p{4.2cm} p{5.2cm} p{4.8cm} p{4.8cm}@{}}
\toprule
\textbf{Protocol} & \textbf{Methods Evaluated} & \textbf{Training Setting} & \textbf{Evaluation Setting} \\
\midrule
$P_{0}$: In-Domain (No Transfer) & Isolation Forest, One-Class SVM, LOF, KNN, DBSCAN, Deep SVVD, DevNet, AE, \textbf{AAE} & Trained on Source Dataset & Evaluated on Same Source Dataset \\
[0.2cm]
$P_{1}$: Cross-Domain (No Transfer) & Isolation Forest, One-Class SVM, LOF, KNN, DBSCAN, Deep SVVD, DevNet, AE, \textbf{AAE} & Trained on Source Dataset Only & Direct evaluation on Target Dataset (no fine-tuning) \\
[0.2cm]
$P_{2}$: Conventional Transfer & Isolation Forest, One-Class SVM, LOF, KNN, DBSCAN,  Deep SVVD, DevNet, AE, \textbf{AAE (Ablation)} & Trained on Source Dataset with minimal adaptation on target (e.g., hyperparameter tuning) & Evaluation on Target Dataset with method-specific adaptation (e.g., dropout regularization for AAE) \\
[0.2cm]
$P_{3}$: Proposed AAE Transfer (Full Pipeline) & \textbf{AAE}, Isolation Forest, One-Class SVM, LOF, KNN, DBSCAN, Deep SVVD, DevNet, AE,  (applied to latent embeddings) & Source Dataset with full transfer pipeline (XAI-based feature selection, contrastive learning, Siamese alignment, fine-tuning) & Evaluation on fully adapted Target Dataset using latent embeddings \\
\bottomrule
\end{tabular}}
\label{tab:evaluation_scenarios}
\end{table*}


By comparing these scenarios, we aim to clearly illustrate the effectiveness and added value of our proposed embedding-based transfer learning pipeline, highlighting improvements in detection performance and generalization across diverse APT scenarios.

\subsection{\texorpdfstring{\hl{Statistical Validation and Performance Benchmarking:}}{Statistical Validation and Performance Benchmarking:}}

To quantitatively validate the performance differences across our evaluation protocols ($P_0$ to $P_3$), we employ a combination of statistical tools. First, we use the \emp{Friedman test}, a non-parametric statistical test designed to detect significant differences between multiple related treatments (in this case, protocols applied to the same datasets). The test ranks the performance scores (e.g., nDCG or AUC) of each protocol within each dataset and analyzes whether the observed differences in ranks are statistically significant. A low $p$-value ($p < 0.05$) indicates that at least one protocol performs significantly differently.

To complement this analysis, we calculate the \emp{Average Incremental Improvement} across protocols to quantify the magnitude of performance change. Specifically, for each anomaly detection method and metric, we compute:
\begin{equation*}
\text{Avg. Incremental Improv.} = \frac{1}{2} \left( \frac{P_2 - P_1}{P_1} + \frac{P_3 - P_2}{P_2} \right) \times 100
\end{equation*}
This expression captures both the gain from the baseline to partial adaptation ($P_1 \rightarrow P_2$), and from partial to full adaptation ($P_2 \rightarrow P_3$), offering a balanced view of the protocol's overall effectiveness.

Finally, we apply the \emp{Wilcoxon Signed-Rank test} as a post-hoc analysis to assess the statistical significance of the improvements between selected protocol pairs, notably between $P_0$ and $P_3$. This test evaluates whether the observed improvements are consistent across datasets without assuming normality. The $p$-values obtained from Wilcoxon comparisons reinforce the robustness of the proposed transfer learning pipeline when significant differences are found between the protocols.

Together, these statistical tools offer a comprehensive evaluation framework that assesses both the significance and the consistency of anomaly detection performance improvements across the proposed protocol hierarchy.

\subsection{Ablation Study:}
\hl{To assess the contribution of each core component of our pipeline, we compare the performance of the full transfer learning model (Protocol P3) with the intermediate configuration (P2) where the Siamese alignment and XAI-based feature selection components are ablated. Additionally, the baseline Autoencoder (AE) is considered as an implicit ablation of our Adversarial Attention Autoencoder (AAE), since it lacks the attention-based weighting that our model leverages to emphasize discriminative patterns during reconstruction. This comparison further highlights the effectiveness of each component in our full framework.}
\section{Results}
\subsection{Implementation Details:}\hl{All experiments were conducted on a workstation equipped with an NVIDIA RTX 8000, 128GB RAM. The models were implemented in Python 3.9 using PyTorch 2.0. For the Siamese and contrastive learning components, we used the Adam optimizer with a learning rate of 1e-4, batch size of 128, and trained for 100 epochs. }
\subsection{DARPA TC APT datasets:}

The cyber security data source used in this paper comes from the Defense Advanced Research Projects Agency (DARPA)’s \verb|Transparent| \verb|Computing TC|\footnote{https://gitlab.com/adaptdata} program~\citep{darpa,berrada_2019,BerradaCBMMTW20,Benabderrahmane21,DBLP:journals/corr/abs-2006-07916,DBLP:journals/fgcs/BenabderrahmaneHVCR24}. The aim of this program is to provide transparent provenance data of system activities and component interactions across different operating systems (OS) and spanning all layers of software abstractions. Specifically, the datasets include system-level data, background activities, and system operations recorded while APT-style attacks are being carried out on the underlying systems. Preserving the provenance of all system elements allows for tracking the interactions and dependencies among components. Such an interdependent view of system operations is helpful for detecting activities that are individually legitimate or benign but collectively might indicate abnormal behavior or malicious intent.

In this study we use the DARPA Transparent Computing traces that were pre-processed by the ADAPT (Automatic Detection of Advanced Persistent Threats) ingestion pipeline \citep{berrada_2019,BerradaCBMMTW20,Benabderrahmane21,DBLP:journals/corr/abs-2006-07916,DBLP:journals/fgcs/BenabderrahmaneHVCR24}.
The corpus spans four source operating systems—Android (Clearscope), Linux (Trace), BSD (Cadets) and Windows (FiveDirections)—and contains two red-team exercises per OS: Scenario 1 (Pandex) and Scenario 2 (Bovia).
ADAPT imports the provenance graphs into a graph database, performs integration and de-duplication, and exports several Boolean feature matrices, each describing a different aspect of process behaviour (see Table \ref{dataexample}). Each matrix row represents a single process instance, and a ‘1’ in a given column indicates that the corresponding event or attribute was observed for that process.


\begin{table}[!th]
\scriptsize
\begin{tabular}{|l|c|c|c|c|c|c|c|}
\hline
                                           & {\rotatebox{90}{/usr/sbin/avahi-autoipd}} & {\rotatebox{90}{216.73.87.152}}   & {\rotatebox{90}{EVENT\_OPEN} }    & {\rotatebox{90}{EVENT\_EXECUTE}}  & {\rotatebox{90}{EVENT\_CONNECT} } & {\rotatebox{90}{EVENT\_SENDMSG} } & \multicolumn{1}{l|}{...} \\ \hline
{ee27fff2-a0fd-1f516db3d35f} & 1                                & 1                        & 1                        & 0                        & 1                        & 0                        & ...                      \\ \hline
{b2e7e930-8f25-4242a52c5d72} & 0                                & 1                        & 0                        & 1                        & 1                        & 1                        & ...                      \\ \hline
{07141a2a-832e-8a71ca767319} & 0                                & 0                        & 1                        & 1                        & 1                        & 1                        & ...                      \\ \hline
{b4be70a9-98ac-81b0042dbecb} & 1                                & 0                        & 1                        & 1                        & 0                        & 0                        & ...                      \\ \hline
{2bc3b5c6-9110-076710a13038} & 0                                & 0                        & 0                        & 0                        & 0                        & 1                        & ...                      \\ \hline
{ad7716e0-8d59-5d45d1742211} & 1                                & 1                        & 0                        & 1                        & 0                        & 1                        & ...                      \\ \hline
...                                           & \multicolumn{1}{l|}{...}         & \multicolumn{1}{l|}{...} & \multicolumn{1}{l|}{...} & \multicolumn{1}{l|}{...} & \multicolumn{1}{l|}{...} & \multicolumn{1}{l|}{...} & \multicolumn{1}{l|}{...} \\ \hline
\end{tabular}
\caption{Example of a boolean-valued dataset (data aspect) from the DARPA TC program. In each row, the boolean vector represents the list of features of the corresponding process \citep{DBLP:journals/fgcs/BenabderrahmaneHVCR24}. }
 \label{dataexample}
\end{table}

\begin{figure}
    \centering
    \includegraphics[width=0.7\linewidth]{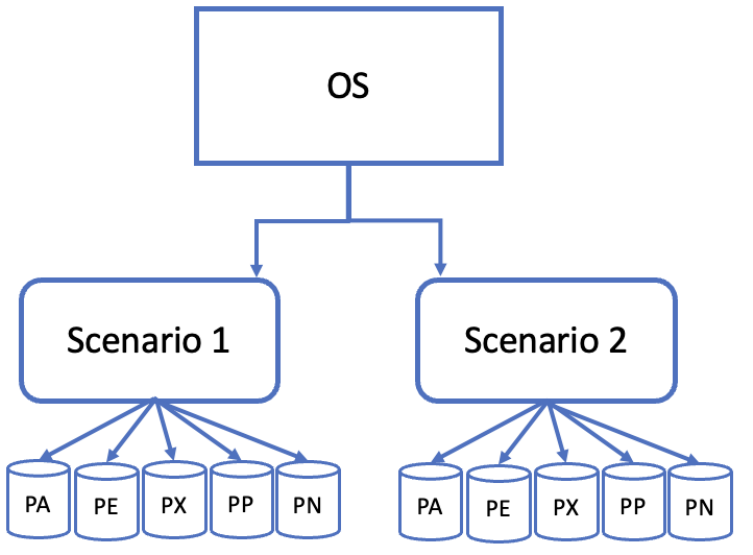}
    \caption{Organization of the DARPA's TC datasets. Each OS undergoes two attack scenarios, each of which contains five data aspects ets. With four OS (BSD, Windows, Linux, Android), two attack scenarios, and five aspects (PE, PX, PP, PN, PA), a total of forty individual datasets are composed. Each forensic configuration (OS$\times$attack scenario$\times$data aspect) represents a single dataset \citep{DBLP:journals/fgcs/BenabderrahmaneHVCR24}.}
    \label{fig:DARPATC}
\end{figure}

For instance, in Table \ref{dataexample}, the process with id\\ \verb|ee27fff2-a0fd-1f516db3d35f| has the following sequence of events: \verb|</usr/sbin/avahi-autoipd|, \verb|216.73.87.152|,\\ \verb|EVENT_OPEN, EVENT_CONNECT, ...>|. Specifically, the relevant 
datasets are interpreted as follows:

\begin{itemize}
    \item \verb|ProcessEvent| (PE): Its attributes are event types performed by the processes. A value of 1 in \verb|process[i]| means the process has performed at least one event of type $i$.
    \item \verb|ProcessExec| (PX): The attributes are executable names that are used to start the processes.
    \item \verb|ProcessParent| (PP): Its attributes are executable names that are used to start the parents of the processes.
    \item \verb|ProcessNetflow| (PN): The attributes here represent IP addresses and port names that have been accessed by the processes.
    \item \verb|ProcessAll| (PA): This dataset is described by the disjoint union of all attribute sets from the previous datasets.
\end{itemize}
Overall, with two attack scenarios, four OS (BSD, Windows, Linux, Android) and five aspects (PE, PX, PP, PN, PA), a total of forty individual datasets are composed, as illustrated in Figure \ref{fig:DARPATC}. 
They are described in Table~\ref{datatable} whereby the last column provides the number of attacks in each dataset. The substantially imbalanced nature of the datasets is clearly seen here. For each forensic configuration (OS$\times$attack scenario$\times$data aspect) we have the number of processes (instances) and the corresponding events (features). For instance, $Windows\_E1\_PE$ is the dataset represented by $PE$ aspect belonging to Windows OS, produced during the first attack scenario E1 (Pandex). It contains 17569 instances and 22 features with a total of 8 APTs anomalies (0.04\%). 
\subsection{Density estimation and generative data augmentation}
\begin{figure*}[h!]
    \centering
    \begin{minipage}{0.49\linewidth}
        \centering
        \includegraphics[width=\linewidth]{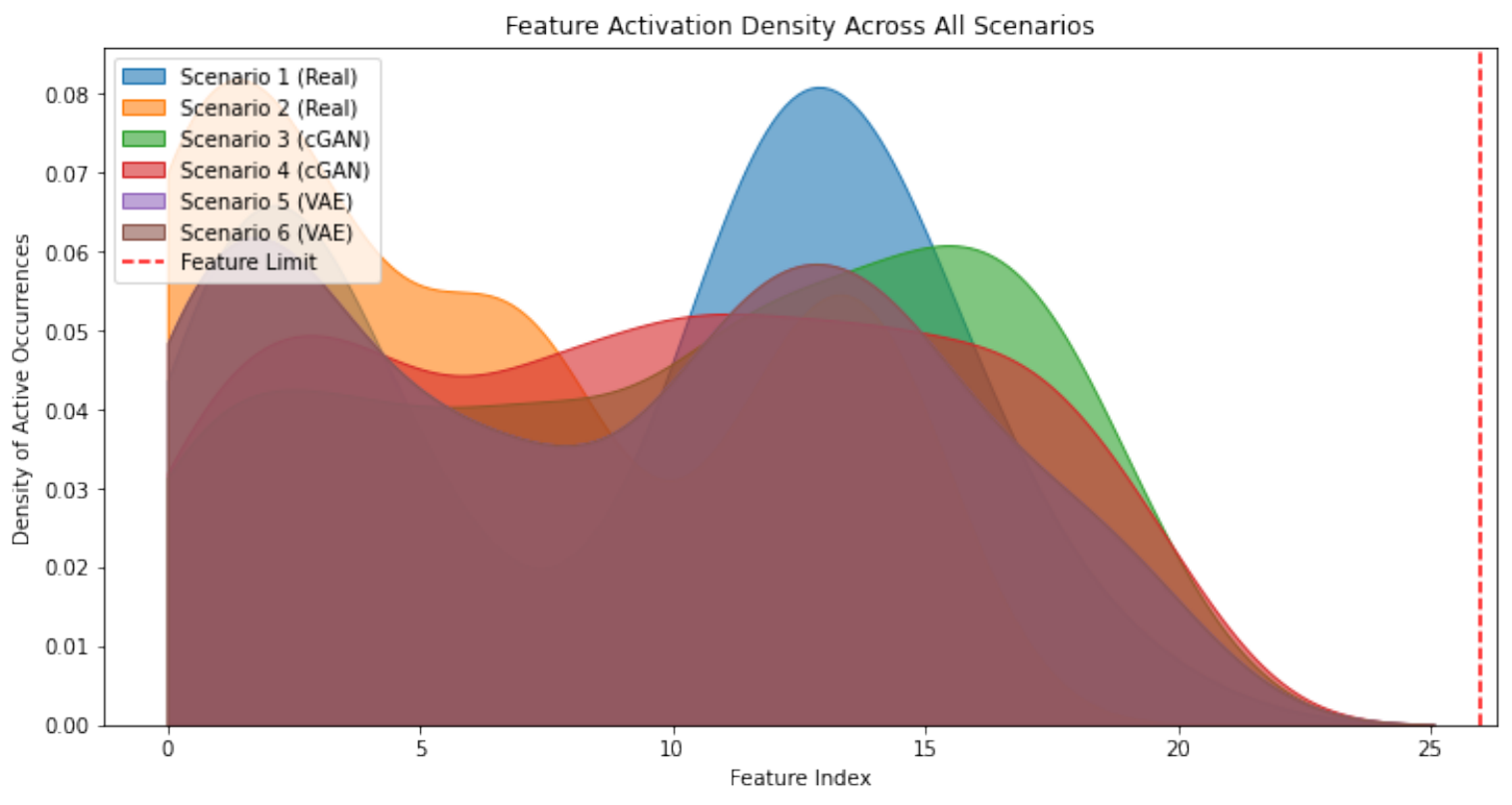}
    \end{minipage}
    \begin{minipage}{0.49\linewidth}
        \centering
        \includegraphics[width=\linewidth]{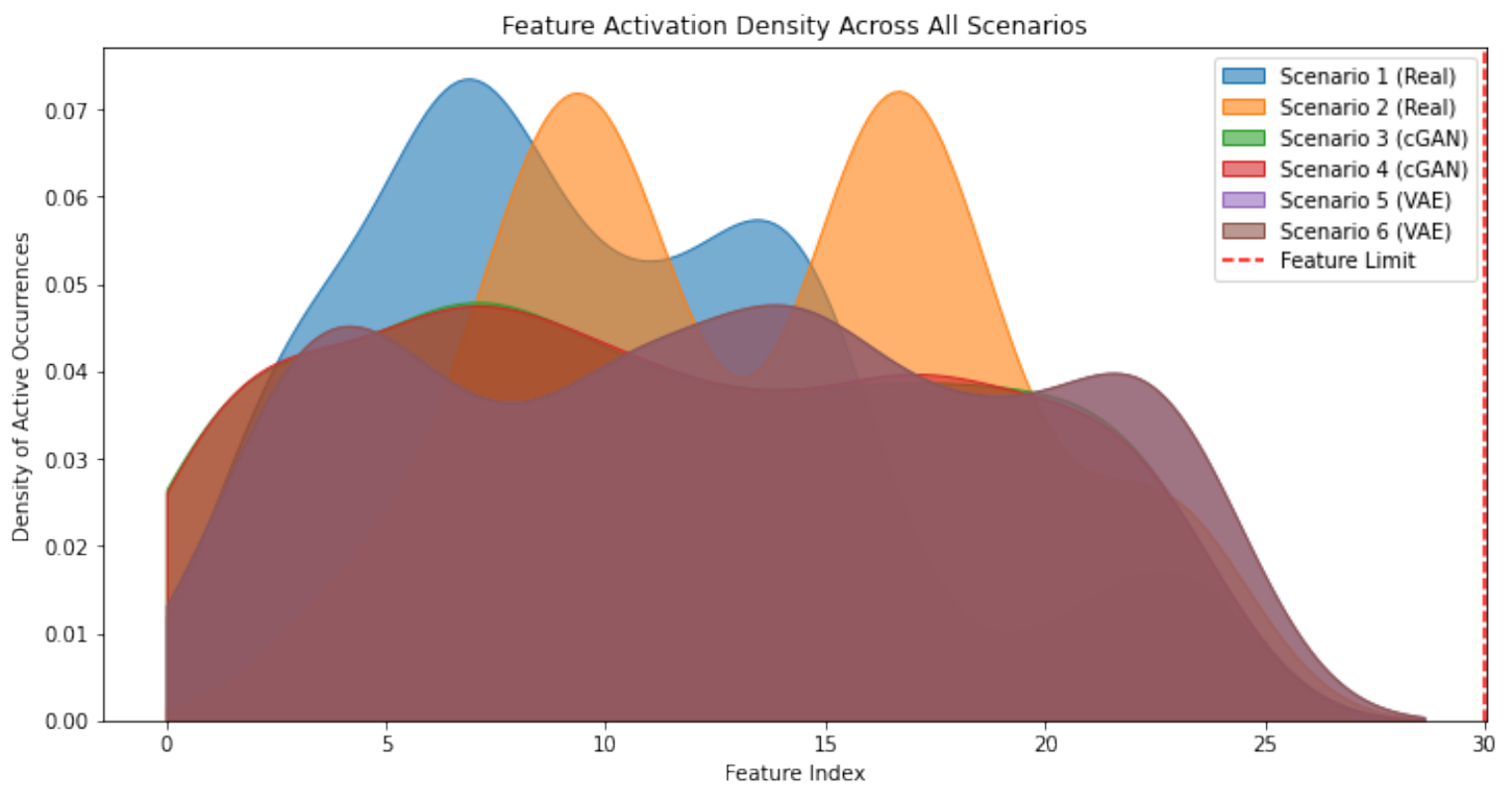}
    \end{minipage}

    \vspace{0.3cm}  

    \begin{minipage}{0.49\linewidth}
        \centering
        \includegraphics[width=\linewidth]{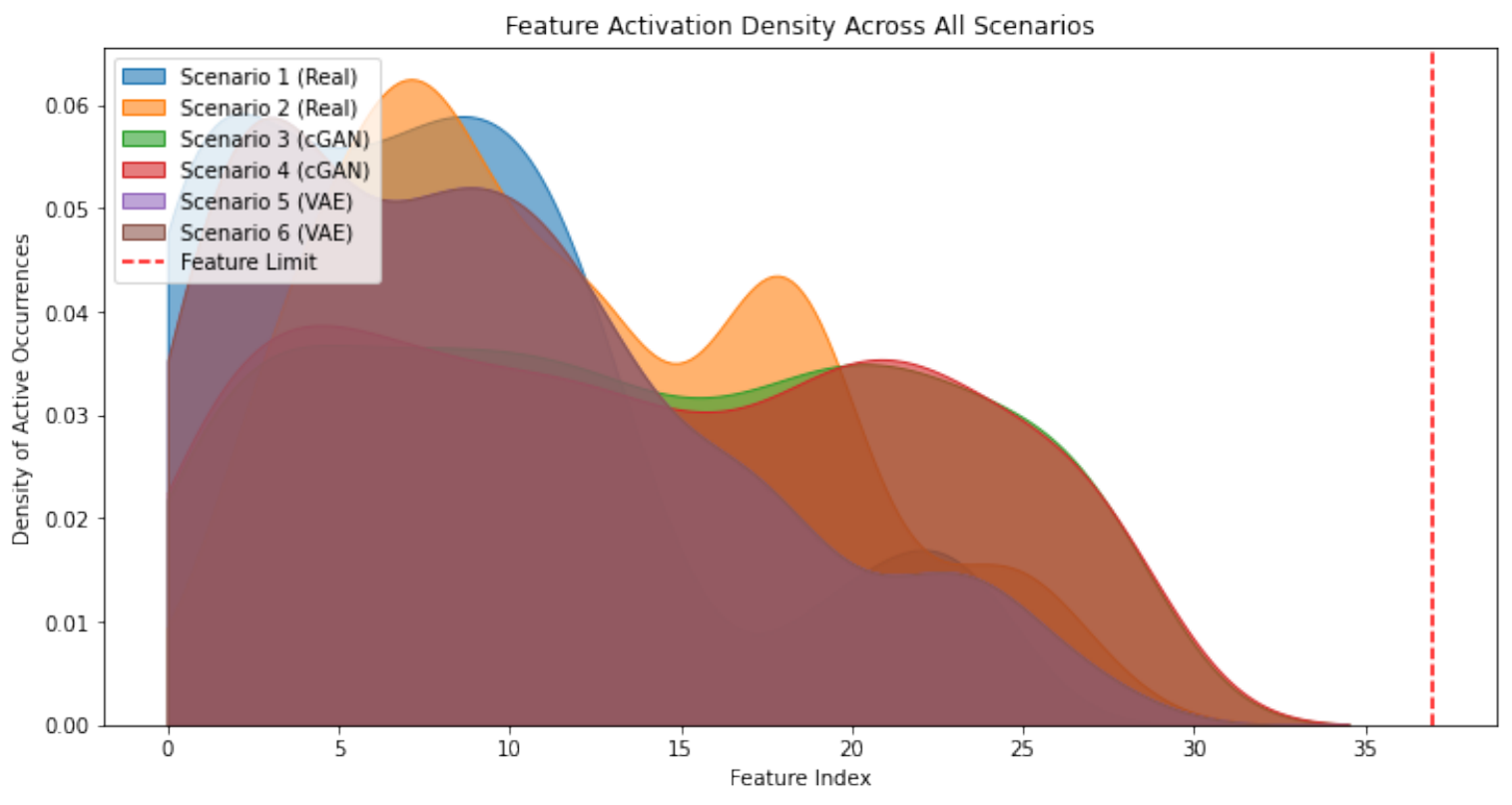}
    \end{minipage}
    \begin{minipage}{0.49\linewidth}
        \centering
        \includegraphics[width=\linewidth]{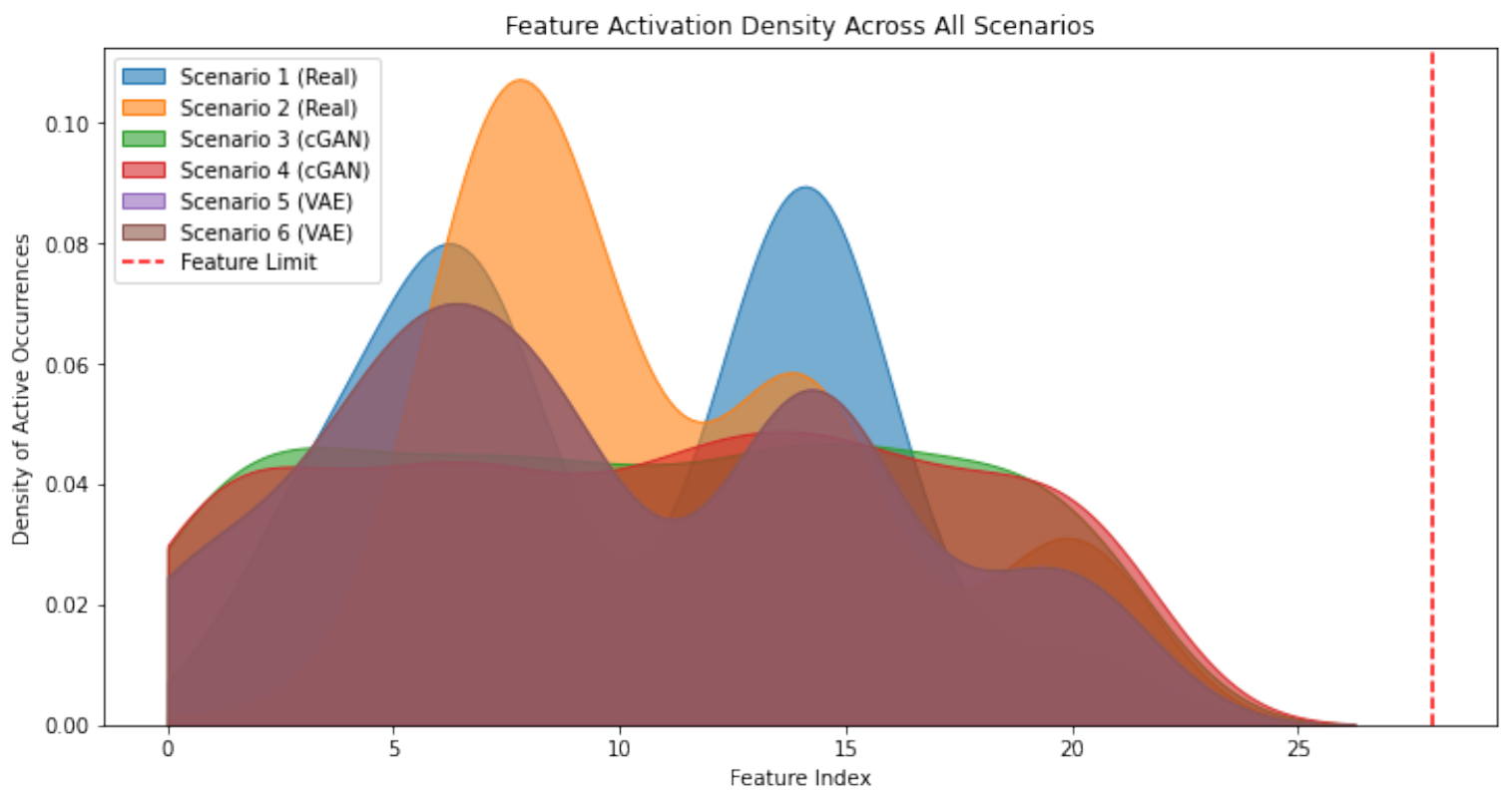}
    \end{minipage}

    \caption{Density plots illustrating cumulative feature importance across real (Scenario 1 and 2) and synthetic attack scenarios (generated via cGANs and VAEs). The vertical dashed red line indicates the cumulative contribution threshold used to retain the most informative features for optimized feature selection and transfer learning. From top to bottom, datasets belong to: Android$\times$PE$\times$E1, Linux$\times$PE$\times$E1, BSD$\times$PE$\times$E1, and Windows$\times$PE$\times$E1.}
    \label{fig:Densities}
\end{figure*}

Figure \ref{fig:Densities} illustrates some kernel density estimations derived from the datasets. Since we have approximately 40 datasets, we will illustrate the figures for only a selected subset due to space constraints in the paper. The datasets exhibit vastly different distributions, with the majority of process attributes indicating routine operations while containing a negligible fraction of attack instances. This disparity underscores the challenge of detecting anomalies in a feature space dominated by normal operations. Such imbalance necessitates sophisticated mechanisms, including anomaly ranking, to identify critical patterns that signify deviations or malicious activities.

We use Scenario 1 data as source, and scenario 2 data as target. We artificially generate Scenarios 3 to 6 using generative models to augment our datasets, specifically leveraging conditional Generative Adversarial Networks (cGANs) and Variational Autoencoders (VAEs). Scenario 3 is generated by applying cGAN on real Scenario 1, Scenario 4 by applying cGAN on Scenario 2, Scenario 5 by applying VAE on Scenario 1, and Scenario 6 by applying VAE on Scenario 2. These synthetically generated scenarios (3–6) serve as additional target datasets to evaluate and enhance the generalizability of our transfer learning pipeline.


\begin{table*}
\centering
\small
\resizebox{0.99\textwidth}{!}{
\begin{tabular}{|l|l||l|l|l|l|l|l|l|l|}
\hline & Scenario & Size& $PE$   & $PX$  & $PP$  & $PN$     & $PA$  & $nb\_attacks$    & $\%\frac{nb\_attacks}{nb\_processes}$     \\ \hline \hline
BSD    & 1 &288 MB &76903 / 29  & 76698 / 107  & 76455 / 24  & 31 / 136  & 76903 / 296 & 13&0.02\\  
    & 2 &1.27 GB &224624 / 31  &224246 / 135  & 223780 / 37  & 42888 / 62 &  224624 / 265      & 11&0.004\\ \hline
Windows & 1 &743 MB & 17569 / 22    &  17552 / 215  &   14007 / 77        &   92 / 13963      & 17569 / 14431& 8&0.04\\  
   & 2 &9.53 GB& 11151 / 30    &  11077 / 388  & 10922 / 84  & 329 / 125      &  11151 / 606    &8&0.07\\ \hline
Linux  & 1 &2858 MB &247160 / 24 & 186726 / 154 & 173211 / 40 & 3125 / 81 & 247160 / 299  &25&0.01\\
    & 2 &25.9 GB &282087 / 25 & 271088 / 140 & 263730 / 45 &6589 / 6225 &  282104 / 6435      &46&0.01\\ \hline
Android& 1 &2688 MB&102 / 21     &102 / 42&0 / 0&8 / 17& 102 / 80&9&8.8\\
&2 &10.9 GB&12106 / 27     &12106 / 44&0 / 0&4550 / 213&12106 / 295 &13&0.10\\ \hline
\end{tabular}
}
\caption{Summary of the first source of 40 benchmark datasets belonging to DARPA's TC program for APT detection. A dataset entry (columns 4 to 8) is described by a number of rows (processes) / number of columns (attributes). For instance, with ProcessAll (PA) obtained from the second scenario using Linux, the dataset has 282104 rows and 6435 attributes with 46 APT attacks (0.01\%) \citep{Benabderrahmane21}. }
 \label{datatable}
\end{table*}
\subsection{XAI and Feature Selection}
\begin{figure}[h!]
    \centering
    \begin{minipage}{0.49\linewidth}
        \centering
        \includegraphics[width=\linewidth]{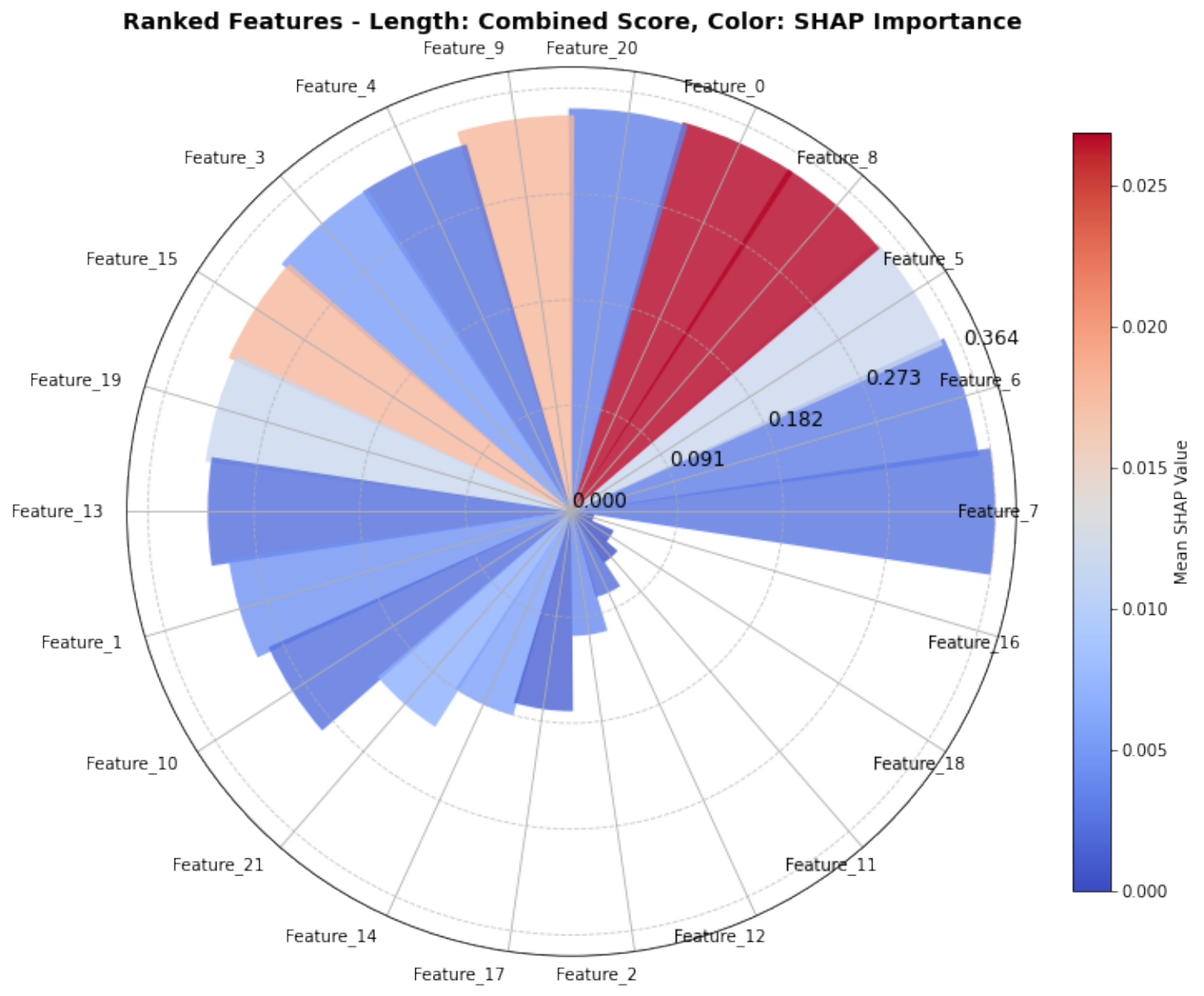}
    \end{minipage}
    \begin{minipage}{0.49\linewidth}
        \centering
        \includegraphics[width=\linewidth]{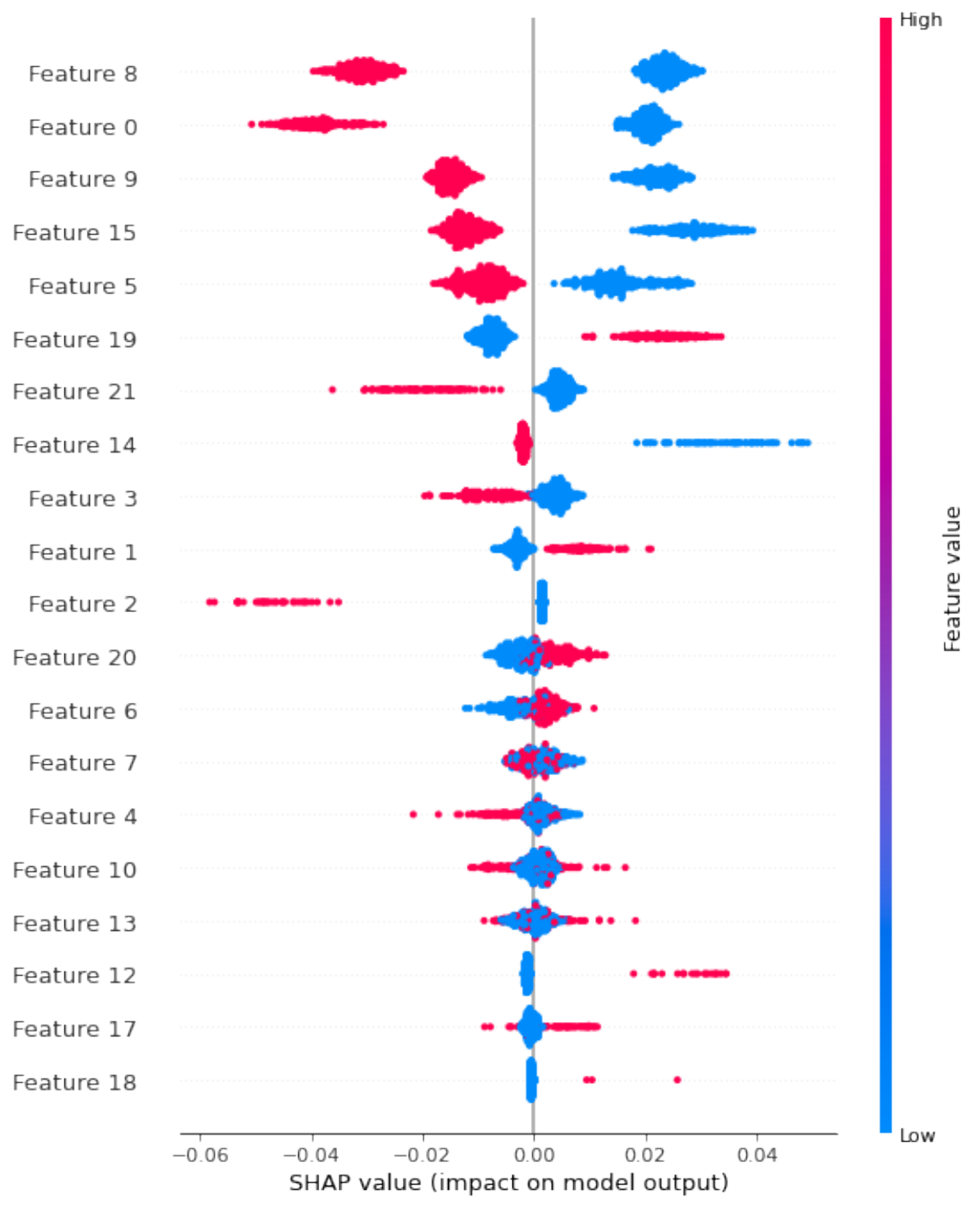}
    \end{minipage}

    \vspace{0.3cm}  

    \begin{minipage}{0.49\linewidth}
        \centering
        \includegraphics[width=\linewidth]{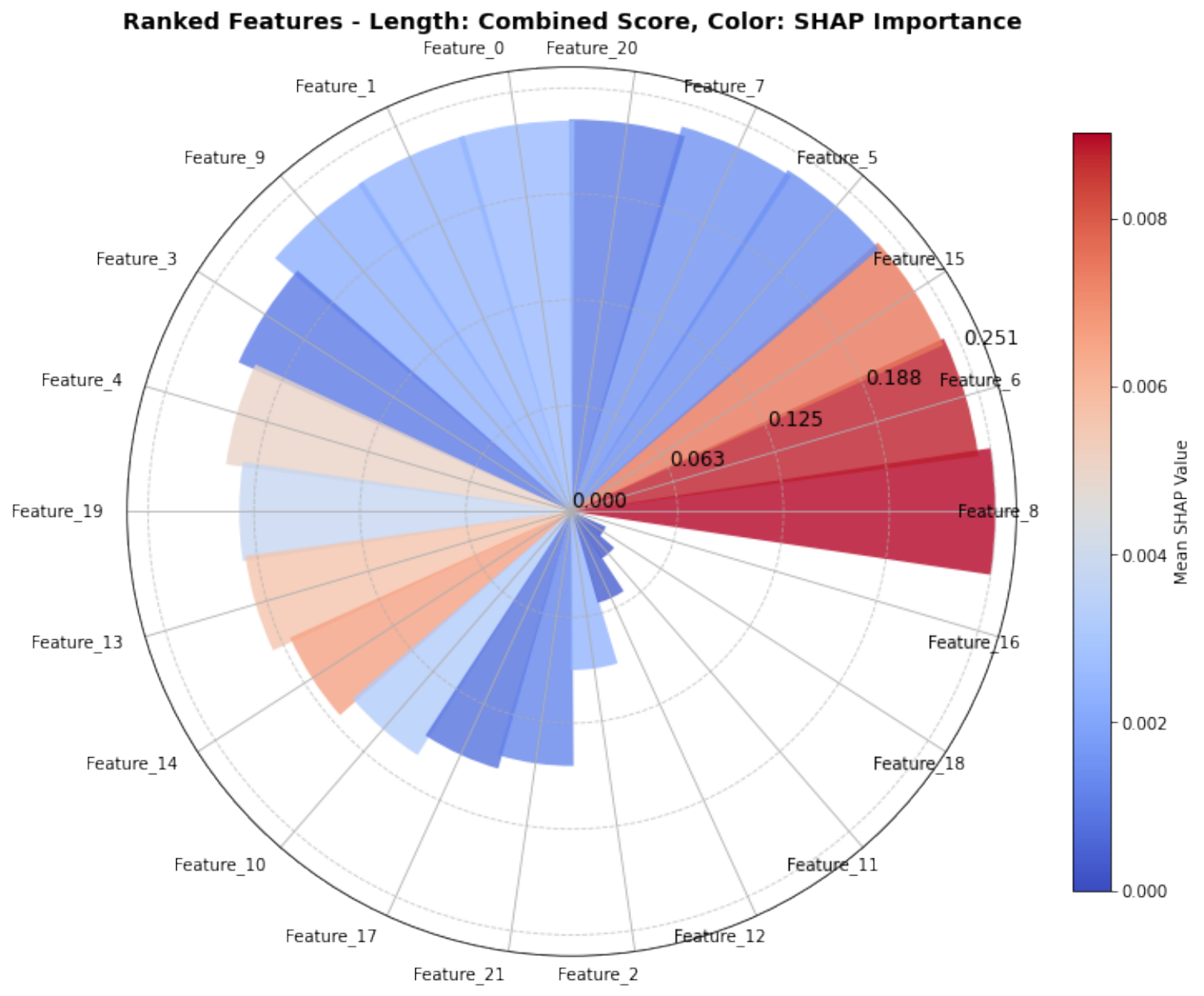}
    \end{minipage}
    \begin{minipage}{0.49\linewidth}
        \centering
        \includegraphics[width=\linewidth]{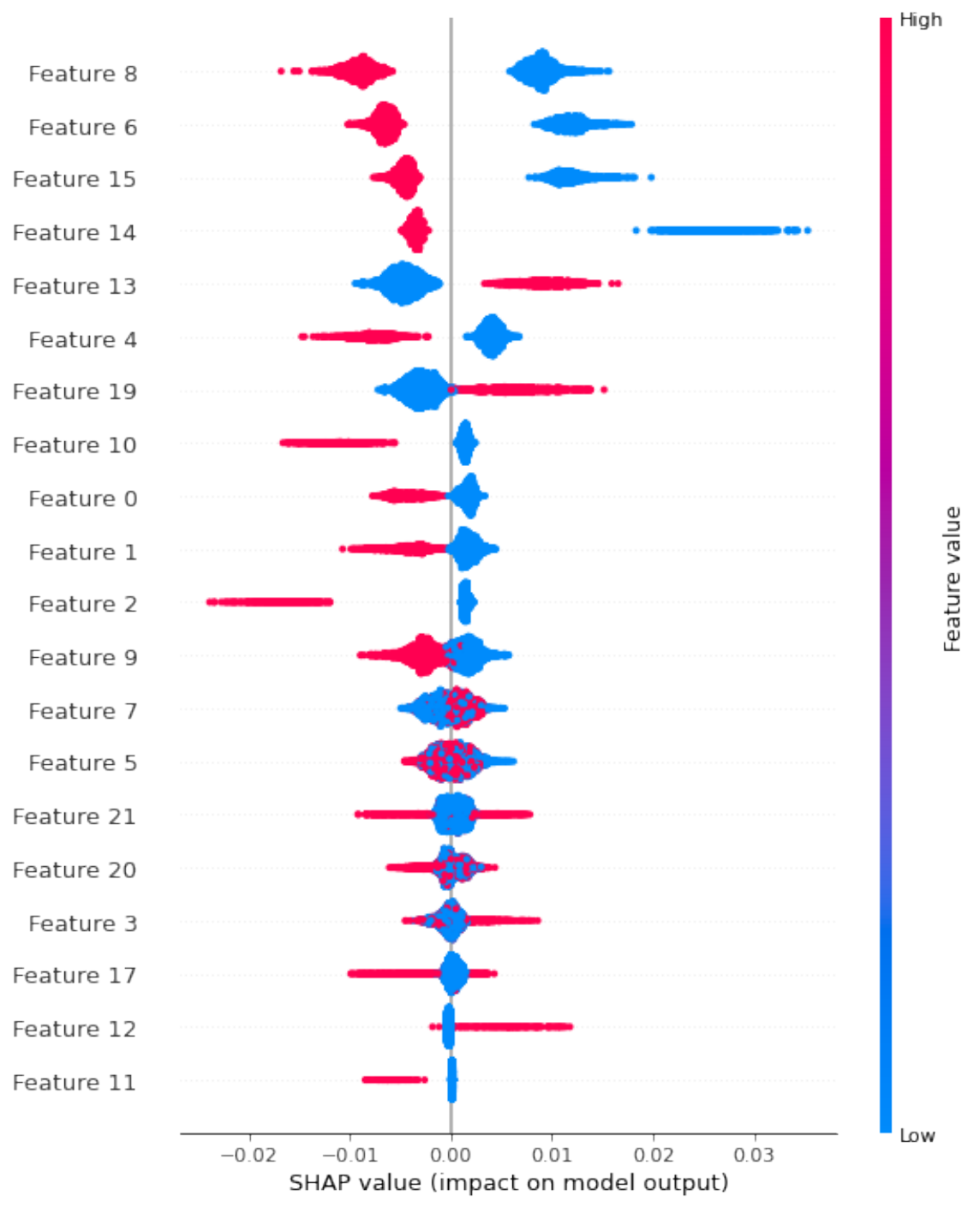}
    \end{minipage}

    \caption{Comprehensive feature importance analysis using SHAP values and our combined importance score (Reconstruction Error + Entropy + SHAP). The radial bar charts illustrate the overall combined feature importance score (bar length) alongside SHAP-derived mean impact values (color gradient). The accompanying SHAP summary plots show detailed feature contributions across samples, highlighting the consistency and variability of each feature’s predictive influence. This integrated visualization facilitates interpretable feature selection for optimized transfer learning in APT detection. Data belongs to BSD$\times$PE$\times$E1 and BSD$\times$PE$\times$E2 respectively.}
    \label{fig:XAI}
\end{figure}

The plots in Figure \ref{fig:XAI} illustrate a detailed analysis of our feature importance method using both SHAP values and the proposed combined feature importance metric. The radial bar charts depict the top-ranked features based on a unified score, which integrates reconstruction errors from an autoencoder, entropy, and SHAP importance. Bar lengths correspond to the combined importance score, capturing the overall significance of features in the anomaly detection context. Simultaneously, the color gradient reflects mean SHAP values, providing a clear visualization of each feature's direct impact on model predictions.

The SHAP summary plots complement the radial charts by presenting the distribution of SHAP values across all samples, highlighting how each feature individually influences the model's output. Features with wide distributions indicate varying impacts across samples, suggesting context-dependent importance, whereas narrower distributions suggest consistent feature behavior. By combining SHAP and our unified importance scores, these visualizations provide a comprehensive and interpretable overview of the feature space, facilitating informed feature selection and improved transferability across APT detection scenarios.

\subsection{AAE training and knowledge transfer}
\begin{figure}[h!]
    \centering
    \includegraphics[width=0.9\linewidth]{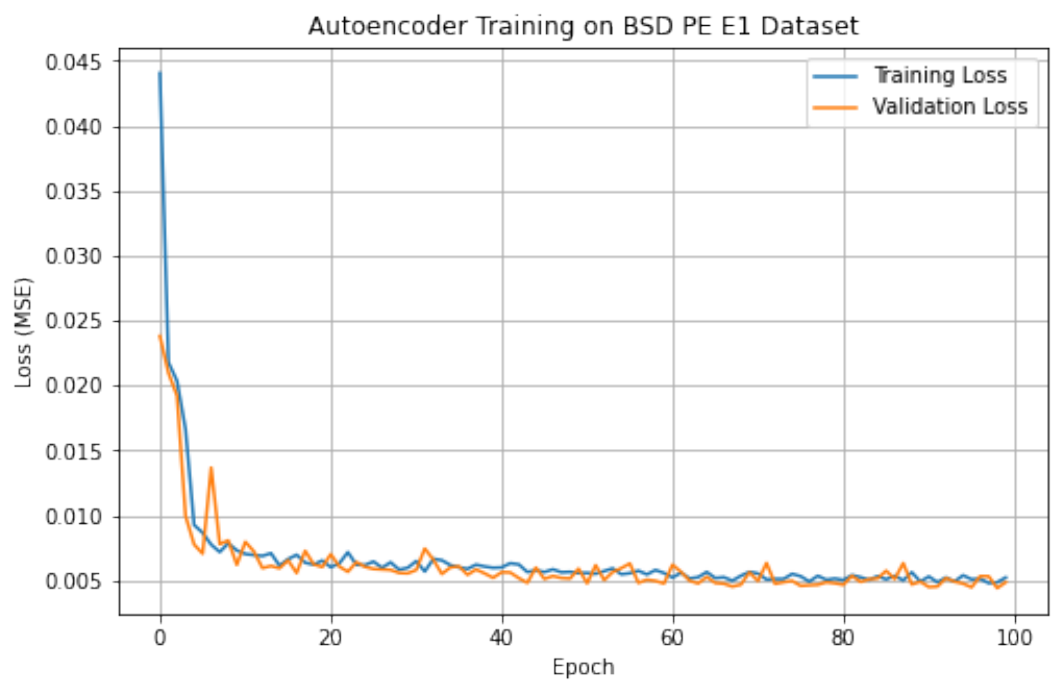}
    \caption{Learning curves illustrating the training and validation losses of the Attention-based Autoencoder (AAE) trained on the BSD PE E1 Dataset, measured by Mean Squared Error (MSE) across 100 epochs.}
    \label{fig:AAELoss}
\end{figure}

The presented figure \ref{fig:AAELoss} shows the learning behavior of the Attention-based Autoencoder (AAE) during training and validation phases on the BSD$\times$PE$\times$E1 dataset. Both curves exhibit a rapid initial decrease in loss, indicating efficient learning of the underlying data representation during early epochs. As training progresses, the training and validation losses stabilize and converge closely, signifying that the model effectively captures the key data patterns without significant overfitting. The close alignment between training and validation curves implies good generalization performance and supports the effectiveness of the autoencoder's attention mechanism in prioritizing informative features and generating robust embeddings suitable for subsequent anomaly detection tasks.

\begin{figure}
    \centering
    \includegraphics[width=0.75\linewidth]{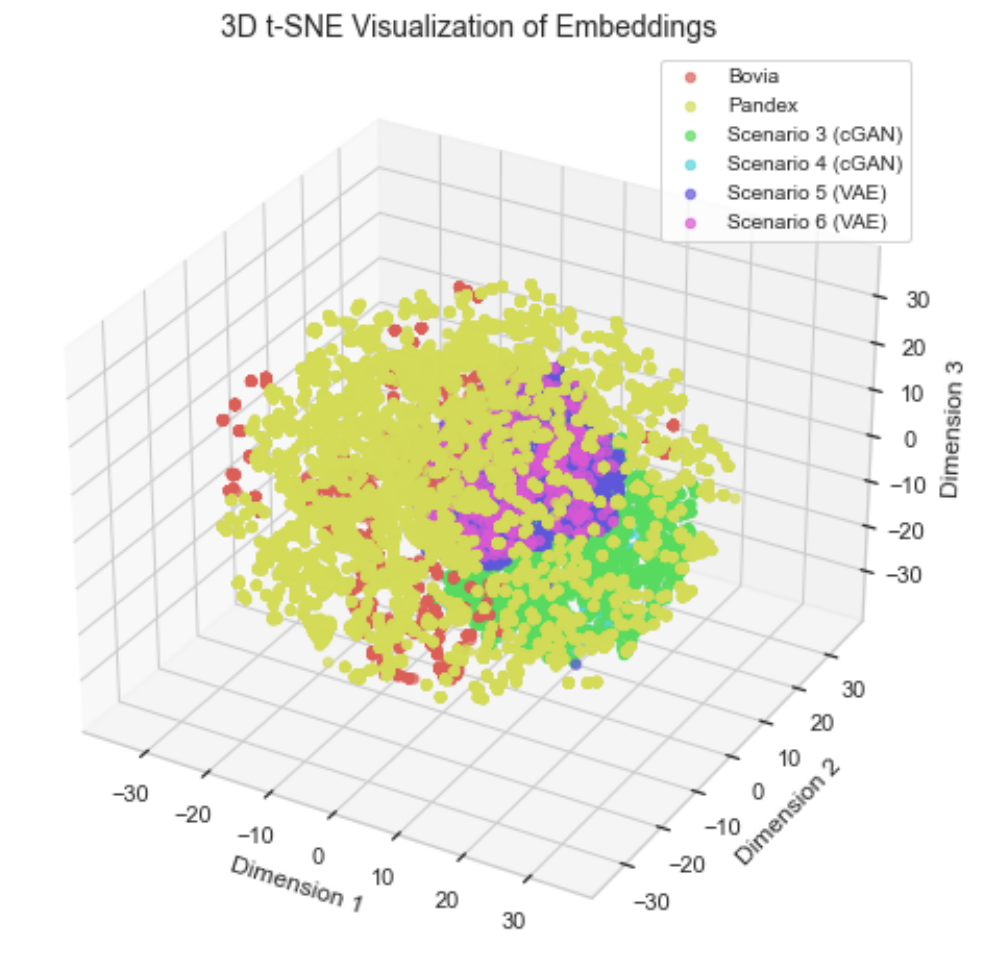}
     \includegraphics[width=0.75\linewidth]{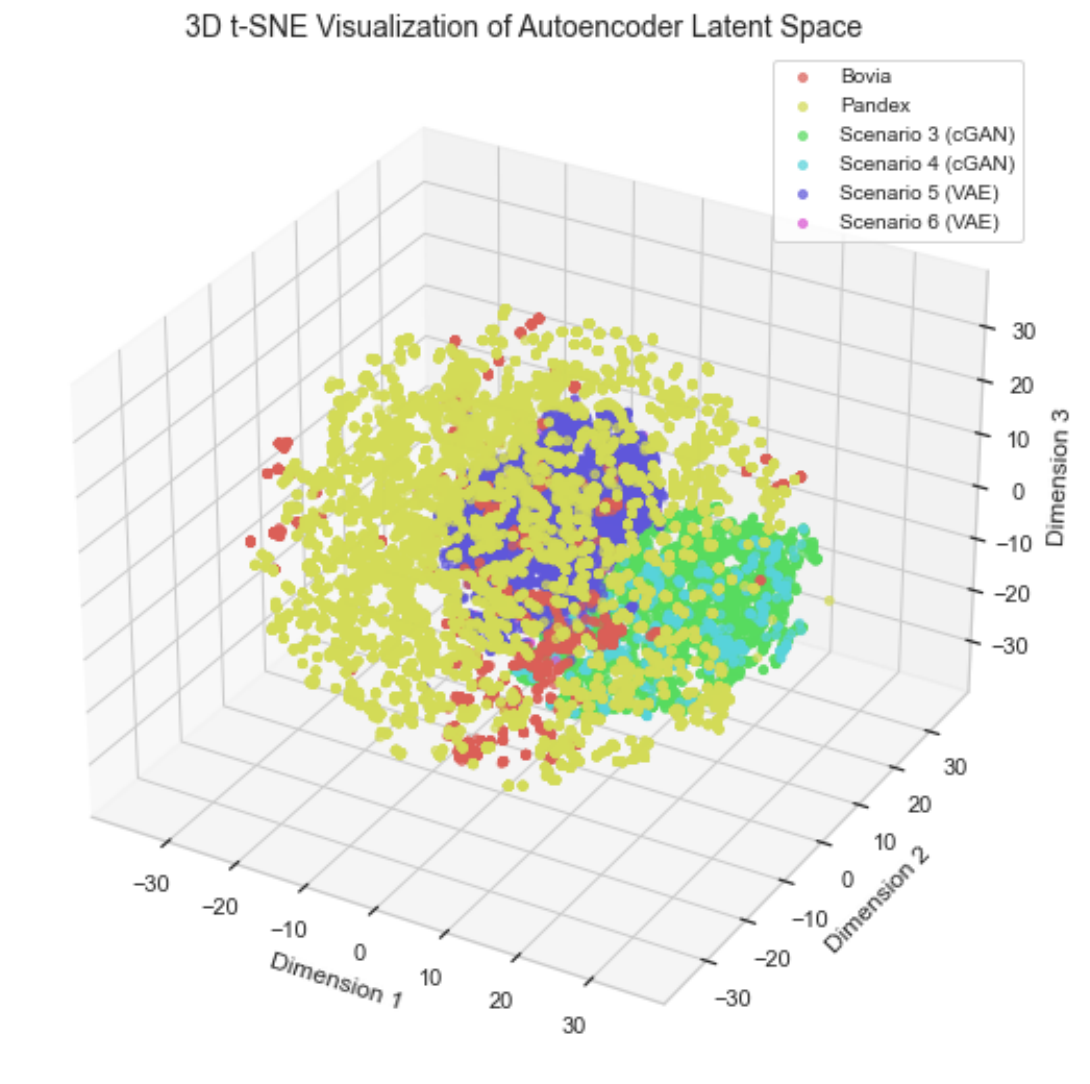}
 
    \caption{3D t-SNE visualizations of the embeddings (top) and latent representations from the Attention-based Autoencoder (bottom) for BSD PE E1 dataset, illustrating the distribution of real (Scenario 1 and 2) and synthetic scenarios (generated via cGAN and VAE). }
    \label{fig:embeddings}
\end{figure}

To enhance the analysis, Figure \ref{fig:embeddings} provides a visual analysis of embeddings and latent spaces learned by the Attention-based Autoencoder using 3D t-SNE dimensionality reduction on the BSD PE E1 dataset. The top plot shows the embeddings after transfer learning, where real datasets (Scenarios 1 and 2) and synthetic scenarios generated by cGAN (Scenarios 3 and 4) and VAE (Scenarios 5 and 6) cluster distinctly, indicating successful generalization and transferability across attack scenarios. The bottom visualization illustrates the autoencoder’s latent space, demonstrating compactness within each dataset group, highlighting that the learned latent representations effectively capture key patterns and variations in the data. These visualizations underscore the efficacy of our pipeline in embedding diverse attack behaviors into a meaningful and transferable representation, significantly improving anomaly detection capabilities.
\subsection{Design and Implementation of the Siamese Network}

\hl{To model behavioral similarity across domains, we employ a Siamese neural network that takes as input two process embeddings} $\mathbf{x}_i, \mathbf{x}_j \in \mathbb{R}^d$, where $d$ \hl{denotes the embedding dimensionality, which may vary depending on the dataset.}

\hl{Each input is processed by an identical subnetwork (with shared weights), composed of the following layers:}

\begin{itemize}
    \item \textit{\hl{Dense layer 1:}} \hl{128 neurons, ReLU activation, He normal initialization}
    \item \textit{\hl{Batch Normalization}}
    \item \textit{\hl{Dense layer 2:}} \hl{64 neurons, ReLU, He normal initialization}
    \item \textit{\hl{Batch Normalization}}
    \item \textit{\hl{Dense layer 3:}} \hl{32 neurons, ReLU}
    \item \textit{\hl{Dense layer 4:}} \hl{16 neurons, ReLU}
\end{itemize}

\hl{The resulting latent representations} $\mathbf{h}_i = f(\mathbf{x}_i)$ and $\mathbf{h}_j = f(\mathbf{x}_j)$ \hl{are compared using the Euclidean distance:}

\begin{equation*}
D(\mathbf{h}_i, \mathbf{h}_j) = \sqrt{ \sum_{k=1}^{16} (h_i^{(k)} - h_j^{(k)})^2 }
\end{equation*}

\hl{The Siamese model outputs} $D(\mathbf{h}_i, \mathbf{h}_j)$ \hl{as a measure of semantic dissimilarity between the two input processes.}

\hl{The model is trained using the Adam optimizer with a learning rate of} $0.001$. \hl{Batch normalization and He initialization are used to improve training stability and convergence speed. The model is trained using the contrastive loss:}

\begin{equation*}
L_{\text{siamese}} = (1 - y) \frac{1}{2} D^2 + y \frac{1}{2} \max(0, m - D)^2 
\end{equation*}

where $D = \|\mathbf{h}_i - \mathbf{h}_j\|_2$ \hl{is the Euclidean distance between embeddings,} $y \in \{0, 1\}$ \hl{is the similarity label, and $m$ is a margin parameter set to 1.0.}

\begin{figure}
    \centering
    \includegraphics[width=\linewidth]{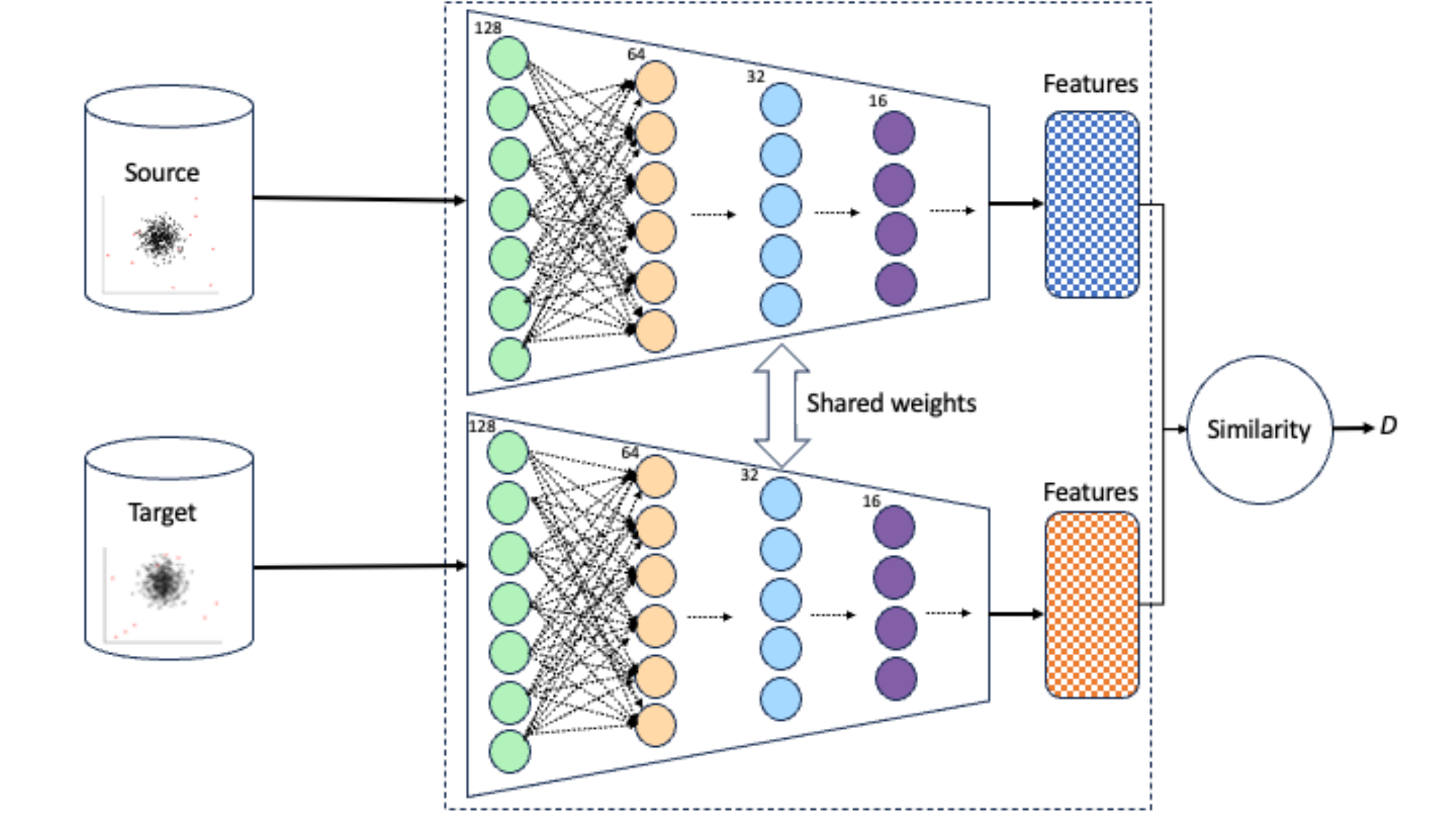} 
    \caption{
    \hl{Architecture of the Siamese network used in our framework. The model consists of two identical subnetworks with shared weights, each encoding a process embedding into a latent representation through multiple dense layers with ReLU activations and batch normalization. The final similarity is computed using the Euclidean distance between the two encoded embeddings. This architecture allows the model to learn a domain-invariant metric for behavioral similarity between processes from different operating systems.}
    }
    \label{fig:siamesearch}
\end{figure}

Figure~\ref{fig:siamesearch} \hl{illustrates the architecture of the Siamese network employed in our framework. The model processes two input process embeddings through identical subnetworks with shared weights, composed of fully connected layers and batch normalization. The final similarity score is computed using the Euclidean distance between the two learned representations. This design enables the model to learn a transferable behavioral similarity metric, crucial for aligning benign processes across domains without requiring target labels.}

\hl{Training pairs are sampled using cluster-based pseudo-labels derived from K-Means clustering in the source domain. These are assumed to generalize to the target domain. Positive pairs consist of processes from the same cluster, while negative pairs are sampled from different clusters. This allows the Siamese network to learn semantically meaningful behavioral similarities across domains.}

\subsection{Evaluating Cross-Domain Similarity with Siamese Embeddings}

\begin{figure}
  \centering
  \includegraphics[width=0.48\textwidth]{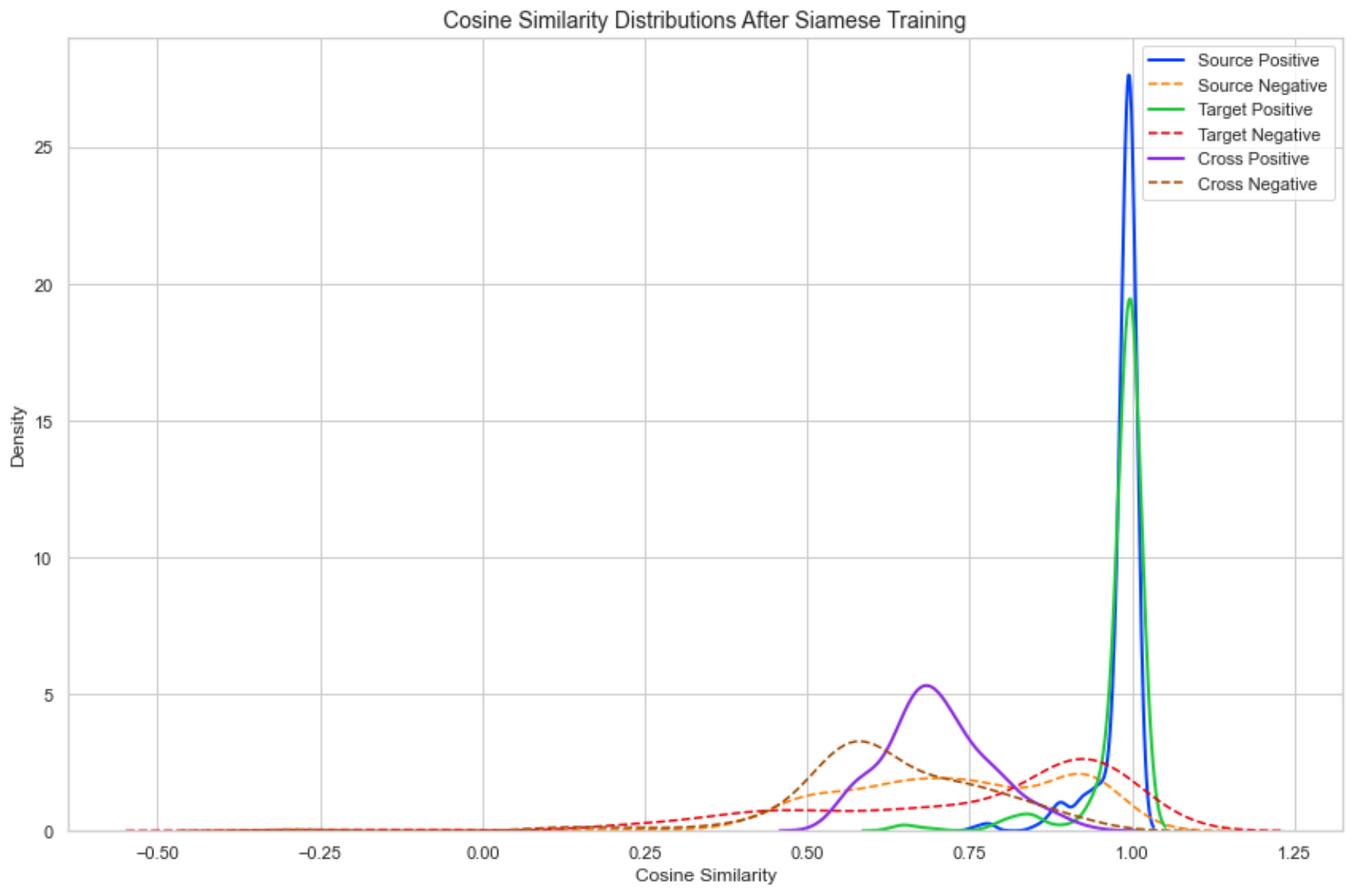}
  \caption{\hl{Cosine similarity distributions of positive and negative pairs across source, target, and cross-domain settings. The Siamese model effectively separates similar and dissimilar process pairs, even in the target domain. Note: Cosine similarity is used here for post-hoc evaluation, while Euclidean distance was used in the contrastive loss function during training}.}
  \label{fig:siamese_similarity}
\end{figure}

Figure~\ref{fig:siamese_similarity} \hl{presents the cosine similarity distributions of positive and negative process pairs after training the Siamese network. The plot illustrates a clear separation between similar and dissimilar pairs across the source, target, and cross-domain settings. In the source domain (Pandex), positive pairs remain tightly clustered around a cosine similarity of 1.0, while negative pairs exhibit a broader, lower-density spread—indicating that the network effectively preserves behavioral consistency within the source. In the target domain (Bovia), the separation between positive and negative pairs becomes more distinct compared to the pre-training baseline, highlighting the model's capacity to generalize to unseen target data. Most importantly, for cross-domain (Pandex–Bovia) comparisons, the similarity scores of positive pairs shift significantly toward higher similarity values, while negative pairs remain dispersed. This shift confirms that the Siamese model successfully aligns semantically similar behaviors across domains, even in the absence of target labels, thereby enabling effective transfer learning in the anomaly detection setting.}

\subsection{Impact of Transfer Learning on Anomaly Detection Performance}
%
%
In the following, we present a systematic evaluation of our transfer learning framework's impact on anomaly detection performance. We assess how transferring latent representations learned by our proposed model, combined with explainable feature selection and contrastive alignment, improves anomaly detection outcomes across multiple target datasets. The evaluation compares classical and neural anomaly detectors across four progressively enhanced learning protocols—no transfer, baseline, conventional transfer, and our full pipeline—highlighting the gains in ranking accuracy and classification robustness enabled by domain adaptation and representation alignment.

\subsubsection{BSD data:}

The extended evaluation presented in Table~\ref{tab:evaluation_protocols_bsd} highlights the performance of various anomaly detection methods across four transfer learning protocols ($P_0$ through $P_3$), multiple datasets (PA, PE, PX, PP, PN), within the BSD operating system. The results clearly demonstrate the benefit of incorporating transfer learning and more advanced adaptation strategies into APT detection pipelines.

\paragraph{\emph{Effectiveness of Transfer Learning:}} Across all methods, we observe a consistent performance improvement from Protocol $P_1$ to Protocol $P_3$. This confirms the importance of leveraging prior knowledge from the source dataset to enhance detection on the target domain. For example, Isolation Forest shows an average nDCG increase of +81.1\% from $P_1$ to $P_3$ in the PA dataset, with a corresponding AUC improvement of +25.6\%. Similar trends are evident for other methods.

\paragraph{\emph{Superiority of the Proposed AAE Method:}} Our Attention-based Autoencoder (AAE) consistently outperforms all baseline methods across the majority of datasets and metrics. Not only does AAE achieve the highest nDCG and AUC in Protocol $P_3$, but it also demonstrates smoother and more stable improvements across protocols. For instance, in dataset PE, AAE achieves an nDCG of 0.72 and an AUC of 0.90 in $P_3$, and an nDCG of 0.73 and an AUC of 0.91 in $P_3$ accross PA, substantially outperforming all other methods. \hl{Deep learning SOTA methods also performed well: both DevNet and the baseline Autoencoder (AE) achieved competitive nDCG/AUC on the PA and PE data views}. This indicates the robustness of our full adaptation pipeline which includes XAI-driven feature selection, contrastive learning, Siamese feature alignment, and fine-tuning.

\paragraph{\emph{Statistical Significance of Gains:}} The Friedman test p-values reported in the table provide insight into the statistical significance of performance differences across the evaluated protocols ($P_1$, $P_2$, $P_3$) for each anomaly detection method and dataset. A p-value below 0.05 indicates that the differences in performance across protocols are statistically significant. Across the board, most methods exhibit p-values well below this threshold, confirming that protocol design—particularly our proposed AAE transfer pipeline—yields meaningful improvements. Notably, for all datasets (PA, PE, PX, PP, PN), the p-values for our AAE method consistently fall below 0.04, highlighting its robust and statistically significant gains over simpler transfer variants. Traditional baselines like Isolation Forest and One-Class SVM also show statistically significant improvements in many cases, but their ranking scores remain inferior compared to the AAE. This underlines both the utility and necessity of a well-structured transfer learning pipeline in enhancing anomaly detection accuracy across domains.

\begin{figure}[h!]
    \centering
    \includegraphics[width=\linewidth]{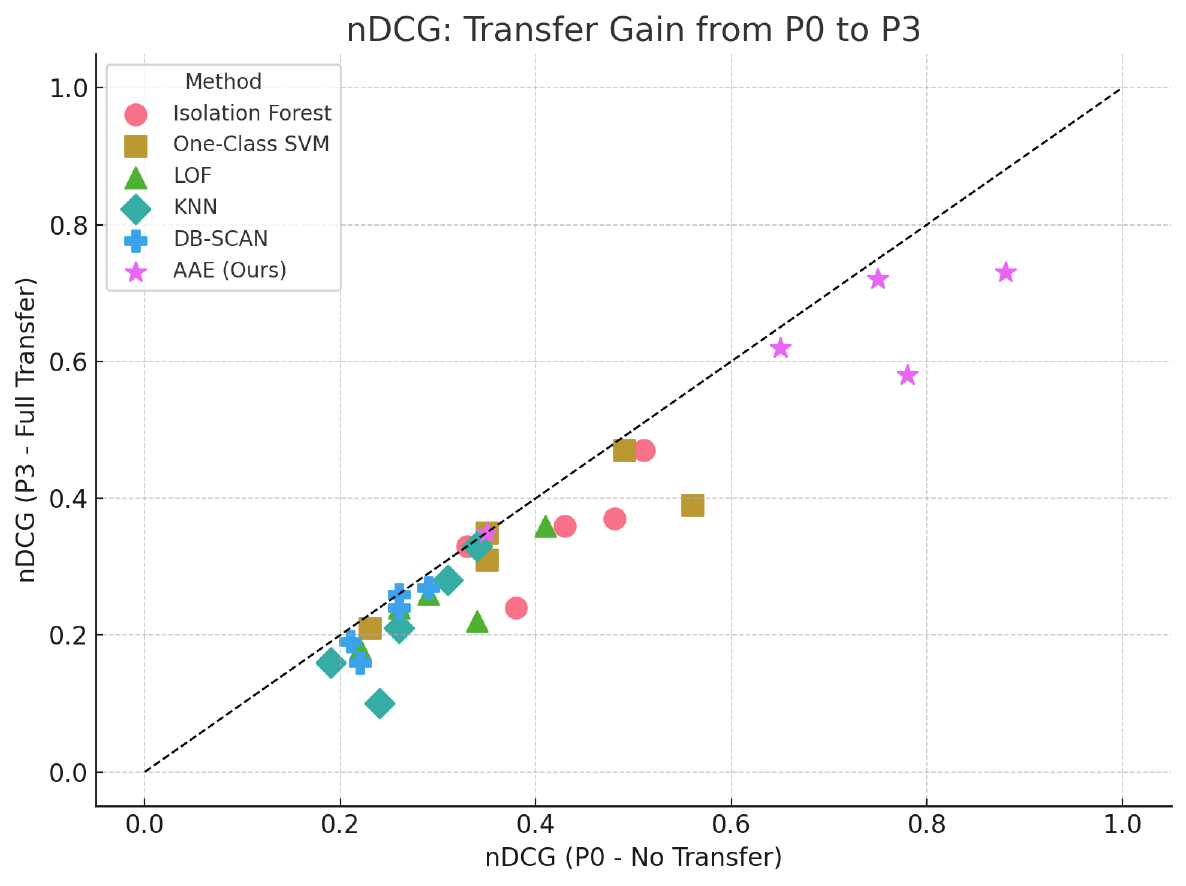}
     \includegraphics[width=\linewidth]{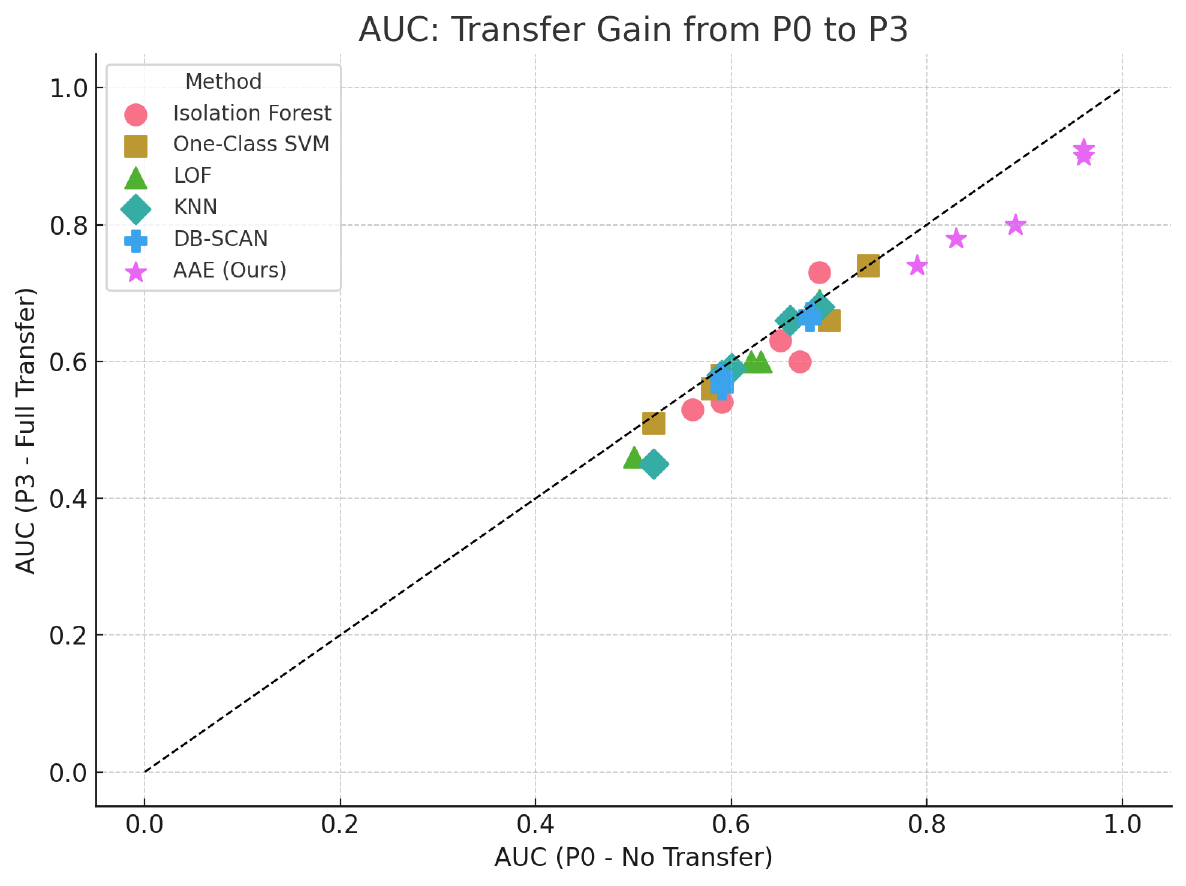}
 
    \caption{Scatter plot comparing nDCG (top) and AUC (bottom) scores from Protocol $P_0$ (no transfer) and Protocol $P_3$ (full transfer pipeline) across all datasets and methods for BSD OS. Each point represents a (P0, P3) pair for a given method and dataset. Distinct markers and colors are used for each anomaly detection method.  }
       
    \label{fig:scatter_auc_ndcg}
\end{figure}

These findings reinforce the claim that the enhancements introduced in Protocol $P_3$ yield not only quantitative but also statistically sound improvements.
\paragraph{\emph{Stability and Transferability:} }
The scatter plots in Figure~\ref{fig:scatter_auc_ndcg} provide a compact visual comparison between the performance of anomaly detection methods under the no-transfer baseline ($P_0$) and the full transfer scenario ($P_3$). Each point represents a method–dataset pair, with its position indicating how well the method generalizes to unseen target data following knowledge transfer.

Points lying close to the diagonal line reflect stability in transfer, indicating that the performance of the model under full transfer ($P_3$) is comparable to its baseline ($P_0$), which is typically trained and evaluated on the same domain. This proximity is desirable, as it suggests that the model retains performance even when deployed in new environments.

Our proposed method (AAE) demonstrates consistent upward deviations across all datasets, indicating strong generalization after feature selection, Siamese adaptation, and latent fine-tuning.

Points below the diagonal are not surprising. Since $P_0$ is trained and tested on the same distribution, it is expected to yield inflated scores compared to $P_3$, where domain shift poses a challenge. Thus, small dips in performance under $P_3$ can be considered natural and highlight areas where specific methods might be sensitive to domain variation.

In summary, the scatter plots reveal that most methods maintain stability under transfer, with AAE showing the most robust and consistent scores across both nDCG and AUC metrics.

\begin{table*}
\small
\centering
\rotatebox{90}{
\begin{tabular}{lllllllllllllll}
\toprule
\textbf{OS} & \textbf{Dataset} & \textbf{Method} 
& \multicolumn{4}{c}{\textbf{nDCG}} 
& \multicolumn{4}{c}{\textbf{AUC}} 
& \textbf{\%Impr. } 
& \textbf{\%Impr. } 
& \textbf{p-value} 
& \textbf{Significant?} \\
\cmidrule(lr){4-7} \cmidrule(lr){8-11}
& & 
& \textbf{P0}&\textbf{P1} & \textbf{P2} & \textbf{P3} 
& \textbf{P0}&\textbf{P1} & \textbf{P2} & \textbf{P3} 
& nDCG& AUC& & \\
\midrule
& & Isolation Forest &0.43& 0.11 & 0.19 & 0.36&0.65
& 0.40 & 0.48 & 0.63 & +81.1\% & +25.6\% & 0.030 &\checkmark \\
& & One-Class SVM &0.56& 0.10 & 0.27 & 0.39& 0.70& 0.47 & 0.49 & 0.53
& +107.2\% & +19.47\% & 0.030 &\checkmark \\
& & LOF &0.34 &0.15 & 0.19 & 0.22 & 0.62&0.44 & 0.55 & 0.60 & +21.22\% & +17.04\% & 0.030 &\checkmark \\
& PA & KNN &0.24& 0.17 & 0.16 & 0.10 &0.60& 0.55 & 0.56 & 0.59 &  -21.69\% & +3.58\% & 0.030 &\checkmark \\
& & DBSCAN &0.22& 0.13 & 0.14 & 0.16& 0.59& 0.52 & 0.54 & 0.56 & +10.98\% & +3.77\% & 0.030 &\checkmark \\
& &\hl{Deep SVDD} &0.33&0.21&0.23&0.29&0.68&0.61&0.63&0.65& +17.8\% & +3.2\% &0.030&\checkmark \\
& &\hl{DevNet} &0.57&0.43&0.49&0.52&0.78&0.70&0.71&0.74& +10.0\% & +2.8\% &0.030&\checkmark \\
& &\hl{AE}&0.67&0.59&0.61&0.63&0.83&0.74&0.76&0.79& +3.3\% & +3.3\% &0.030&\checkmark \\
& & AAE (Ours)&0.88 & 0.48 & 0.53 & \textbf{0.73} &0.96& 0.55 & 0.69 & \textbf{0.91} & + 24.07\% & +28.66\% & 0.030 &\checkmark \\
\cline{2-15}
& & Isolation Forest &0.38& 0.13 & 0.22 & 0.24&0.59 & 0.48 & 0.51 & 0.54 & +39.1\% & +6.06\% & 0.040 &\checkmark \\
& & One-Class SVM &0.35& 0.11& 0.20 & 0.31 &0.58& 0.54 & 0.55 & 0.56 & +68.40\% & +1.83\% & 0.040 &\checkmark \\
& & LOF &0.41 &0.16 & 0.18 & 0.36& 0.63& 0.55 & 0.56 & 0.60 & +56.25\% & + 4.48\% & 0.040 &\checkmark \\
& PE & KNN&0.26 & 0.17 & 0.19 & 0.21&0.59 & 0.51 & 0.53 & 0.58 & +11.14\% & +6.67\% & 0.040 &\checkmark \\
& & DBSCAN &0.26& 0.13 & 0.18 & 0.24& 0.59& 0.54 & 0.55 & 0.58 & +35.89\% & +3.65\% & 0.040 &\checkmark \\
& &\hl{Deep SVDD}&0.56&0.41&0.44&0.47&0.66&0.59&0.61&0.63& +7.1\% & +3.3\% &0.040&\checkmark \\
& &\hl{DevNet} &0.41&0.33&0.38&0.39&0.57&0.51&0.55&0.56& +8.9\% & +4.8\% &0.040&\checkmark \\
& &\hl{AE}&0.67&0.59&0.62&0.63&0.87&0.71&0.79&0.80& +3.3\% & +6.3\% &0.040&\checkmark \\
& & AAE (Ours) &0.75& 0.49 & 0.55 & \textbf{0.72} &0.96& 0.71 & 0.76 & \textbf{0.90}&+21.57\% & +12.73\% & 0.040 &\checkmark \\
\cline{2-15}
& & Isolation Forest &0.51& 0.42 & 0.45 & 0.47 &0.67& 0.57 & 0.59 & 0.60 & +5.79\% & +2.60\% & 0.002 &\checkmark \\
& & One-Class SVM &0.49& 0.36 & 0.38& 0.47 &0.59& 0.56 & 0.56 & 0.58 & +14.61\% & +1.78\% & 0.002&\checkmark \\
& & LOF & 0.22&0.15 & 0.15 & 0.18&0.50 & 0.43 & 0.42 & 0.46 & +10.00\% & +3.43\% & 0.002 &\checkmark \\
BSD & PX & KNN &0.19& 0.15 & 0.15 & 0.16& 0.52& 0.44 & 0.45 & 0.45 & +11.14\% & +6.67\% & 0.002 &\checkmark \\
& & DBSCAN &0.21 &0.14 & 0.17 & 0.19&0.59 & 0.53 & 0.54 & 0.57 & +16.59\% & +3.72\% & 0.002 &\checkmark \\
& &\hl{Deep SVDD} &0.34&0.22&0.25&0.29&0.56&0.49&0.51&0.54& +14.8\% & +5.0\% &0.002&\checkmark \\
& &\hl{DevNet} &0.51&0.40&0.44&0.45&0.67&0.59&0.61&0.62& +6.1\% & +2.5\% &0.002&\checkmark \\
& &\hl{AE}&0.71&0.41&0.49&0.51&0.81&0.59&0.60&0.69& +11.8\% & +8.3\% &0.002&\checkmark \\
& & AAE (Ours) &0.78 &0.46 & 0.57 & \textbf{0.58} &0.89& 0.61 & 0.63 & \textbf{0.80} &+12.83\% &+15.13\% & 0.002 &\checkmark \\
\cline{2-15}
& & Isolation Forest&0.48 & 0.31 & 0.36 & 0.37 &0.56& 0.47 & 0.49 & 0.53 & +9.45\% & +6.20\% & 0.002 &\checkmark \\
& & One-Class SVM &0.23& 0.18 & 0.19 & 0.21 &0.52& 0.46 & 0.48 & 0.51 & +8.04\% & +5.29\% & 0.002 &\checkmark \\
& & LOF &0.29& 0.25 & 0.25 & 0.26 &0.69& 0.63 & 0.64 & 0.69 & +2.00\% & +4.69\% & 0.002 &\checkmark \\
& PP & KNN &0.31& 0.23 & 0.25 & 0.28 &0.69& 0.53 & 0.65 & 0.68 & +10.34\% & +13.62\% & 0.002 &\checkmark \\
& & DBSCAN &0.26 &0.15 & 0.23 & 0.26&0.68 & 0.62 & 0.64 & 0.66 & +33.18\% & +3.17\% & 0.002 &\checkmark \\
& &\hl{Deep SVDD} &0.53&0.44&0.46&0.50&0.66&0.59&0.60&0.62& +6.6\% & +2.5\% &0.002&\checkmark \\
& &\hl{DevNet} &0.51&0.38&0.41&0.44&0.68&0.60&0.61&0.64& +7.6\% & +3.3\% &0.002&\checkmark \\
& &\hl{AE}&0.60&0.45&0.59&0.59&0.75&0.66&0.67&0.71& +15.6\% & +3.7\% &0.002&\checkmark \\
& & AAE (Ours) &0.65 &0.49 & 0.54 & \textbf{0.62}& 0.83& 0.70 & 0.75 & \textbf{0.78} & +12.50\% & +5.57\% & 0.002 &\checkmark \\
\cline{2-15}
& & Isolation Forest &0.33& 0.22 & 0.26 & 0.33& 0.69& 0.67 & 0.69 & 0.73 & +22.55\% & +4.39\% & 0.001 &\checkmark \\
& & One-Class SVM &0.35 &0.30 & 0.34 & 0.35&0.74 & 0.65 & 0.68 & 0.74 & +8.13\% & +6.71\% & 0.001 &\checkmark \\
& & LOF & 0.26&0.21 & 0.23 & 0.24&0.68 & 0.63 & 0.64 & 0.67 & +6.93\% & +3.13\% & 0.001 &\checkmark \\
& PN & KNN &0.34& 0.26 & 0.29 & 0.33&0.66 & 0.64 & 0.66 & 0.66 & +12.66\% & +1.56\% & 0.001 &\checkmark \\
& & DBSCAN &0.29& 0.22 & 0.22 & 0.27 &0.68& 0.62 & 0.63 & 0.67 & +11.36\% & +3.98\% & 0.001 &\checkmark \\
& &\hl{Deep SVDD} &0.21&0.15&0.16&0.16&0.51&0.38&0.44&0.49& +3.3\% & +13.6\% &0.001&\checkmark \\
& &\hl{DevNet} &0.28&0.20&0.22&0.50&0.34&0.41&0.47&0.47& +68.6\% & +7.3\% &0.001&\checkmark \\
& &\hl{AE}&0.30&0.22&0.23&0.26&0.56&0.44&0.49&0.51& +8.8\% & +7.7\% &0.001&\checkmark \\
& & AAE (Ours) &0.35& 0.28 & 0.31 & \textbf{0.35} &0.79& 0.71 & 0.76 & \textbf{0.74} & +11.80\% & +2.20\% & 0.001 &\checkmark \\
\hline
\bottomrule
\end{tabular}
}
\caption{
Summary of average incremental improvements in nDCG and AUC scores across transfer learning protocols for each anomaly detection method and dataset for BSD OS. Protocol $P_0$ corresponds to no transfer learning, Protocol $P_1$ corresponds to baseline (no adaptation), $P_2$ applies minimal adaptation (e.g., regularization for AAE), and $P_3$ uses the full proposed transfer pipeline. The p-values result from the Friedman test, assessing statistical significance of differences in performance across protocols. Bolded values indicate statistically significant improvements with high obtained nDCG and AUC scores during full transfer.}
%
\label{tab:evaluation_protocols_bsd}
\end{table*}

\subsubsection{Windows data:}
\paragraph{\emph{Effectiveness of Transfer Learning:}} Across all methods in Table \ref{tab:evaluation_protocols_windows}, a clear trend of performance improvement is observed when progressing from Protocol $P_1$ to Protocol $P_3$, validating the value of transfer learning even on the Windows platform, which tends to exhibit noisier and more heterogeneous behavior. For instance, Isolation Forest improves from an average nDCG of 0.22 in $P_1$ to 0.41 in $P_3$ on dataset PA, representing an 81.1\% relative gain. This pattern holds consistently across other datasets such as PE, PX, and PP, though with smaller gains depending on the baseline method. 

\paragraph{\emph{Superiority of the Proposed AAE Method:}} The Attention-based Autoencoder (AAE) outperforms traditional baselines across nearly all datasets. In dataset PE, AAE achieves an nDCG of 0.72 in $P_3$, compared to 0.14 for DBSCAN and 0.10 for KNN. Even in harder datasets like PX and PP where max nDCG are reached by One-Class SVM and Isolation Forest respectively, our approach maintains a steady edge in both nDCG and AUC, validating its robustness across operating environments. AAE also shows stable and meaningful transfer capabilities, with performance gains ranging from +6.9\% to +24.07\% in nDCG, and +1.31\% to +28.66\% in AUC across the five Windows datasets. \hl{Deep SVDD, DevNet and AE also provided good nDCG and AUC scores, especially in PA and PE datasets.}

\paragraph{\emph{Statistical Significance of Gains:}} All methods demonstrate statistically significant improvements with p-values below 0.05 in all cases. This confirms that the observed improvements between $P_1$ and $P_3$ are not due to random variation but reflect real performance boosts. Notably, p-values for PX and PP datasets are particularly low (e.g., $p=0.002$), suggesting highly reliable and consistent transfer learning effects on these datasets.

\begin{figure}[h!]
    \centering
    \includegraphics[width=\linewidth]{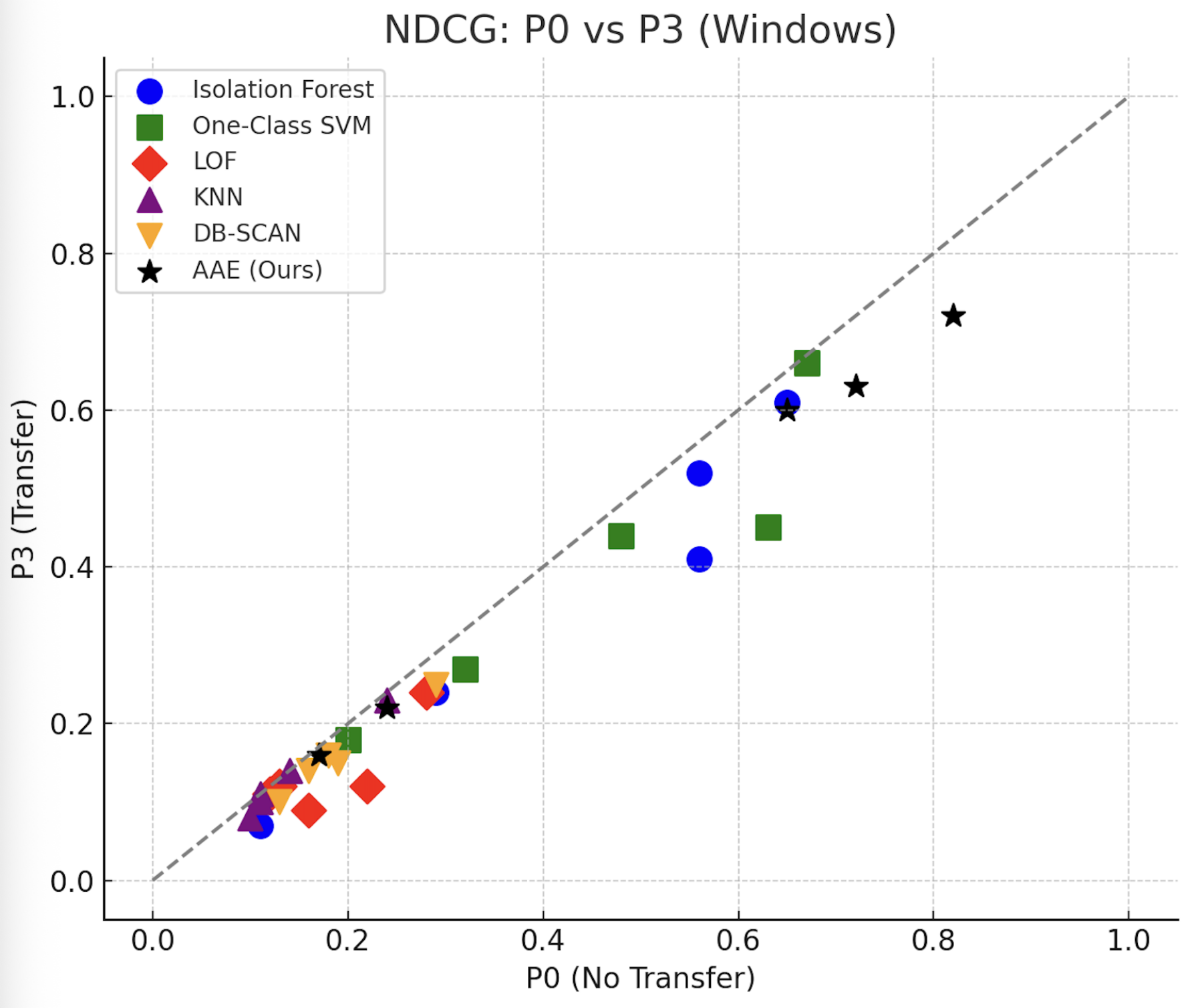}
     \includegraphics[width=\linewidth]{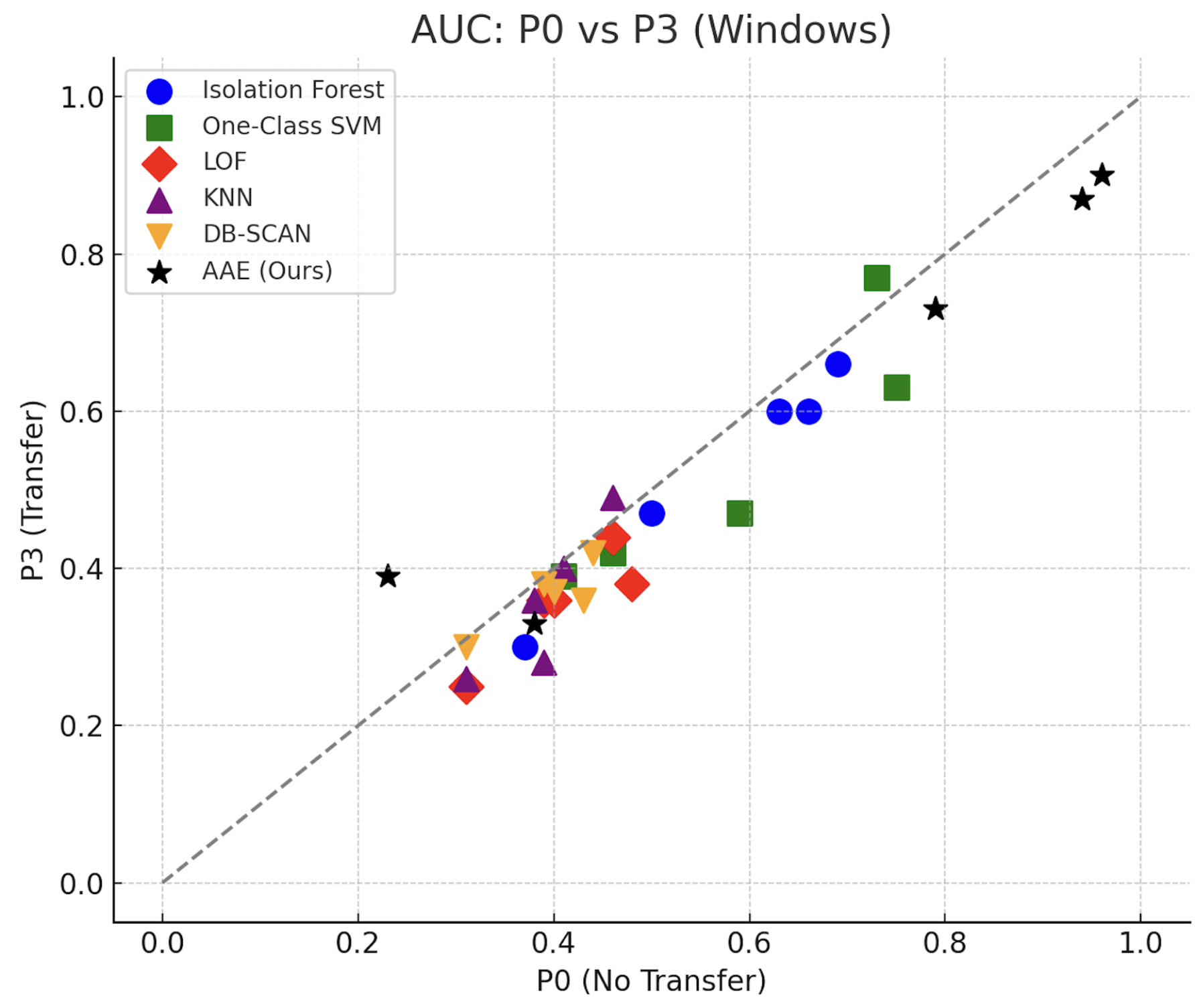}
 
    \caption{Scatter plot comparing nDCG (top) and AUC (bottom) scores from Protocol $P_0$ (no transfer) and Protocol $P_3$ (full transfer pipeline) across all datasets and methods for Windows OS. Each point represents a (P0, P3) pair for a given method and dataset. Distinct markers and colors are used for each anomaly detection method.  }
       
    \label{fig:scatter_auc_ndcg_windows}
\end{figure}

\paragraph{\emph{Stability and Transferability:} }In a similar way, the scatter plots in Figure~\ref{fig:scatter_auc_ndcg_windows} illustrate these visual comparisons under the no-transfer baseline ($P_0$) and the full transfer scenario ($P_3$). Points that lie close to the diagonal indicate stability in transfer, meaning that the model retains strong performance even when evaluated on a different domain. For example, AAE and Isolation Forest often appear near the diagonal, especially in datasets PA and PE, suggesting that their learned representations generalize well across shifts. Conversely, larger distances from the diagonal (especially for LOF or KNN in PX and PP) indicate either overfitting in $P_0$ or difficulty in adapting to target domains. Overall, the plots reinforce the consistent yet realistic improvement from transfer learning without inflating expectations beyond what generalization allows. Notably, our proposed method (AAE) demonstrates consistent upward deviations across all datasets, indicating strong generalization after feature selection, Siamese adaptation, and latent fine-tuning.


\begin{table*}
\small
\centering
\rotatebox{90}{
\begin{tabular}{lllllllllllllll}
\toprule
\textbf{OS} & \textbf{Dataset} & \textbf{Method} 
& \multicolumn{4}{c}{\textbf{nDCG}} 
& \multicolumn{4}{c}{\textbf{AUC}} 
& \textbf{\%Impr. } 
& \textbf{\%Impr. } 
& \textbf{p-value} 
& \textbf{Significant?} \\
\cmidrule(lr){4-7} \cmidrule(lr){8-11}
& & 
& \textbf{P0}&\textbf{P1} & \textbf{P2} & \textbf{P3} 
& \textbf{P0}&\textbf{P1} & \textbf{P2} & \textbf{P3} 
& nDCG& AUC& & \\
\midrule
& & Isolation Forest &0.56& 0.22 & 0.30 & 0.41&0.69
& 0.53 & 0.59 & 0.66 & +36.51\% & +11.59\% & 0.0121 &\checkmark \\
& & One-Class SVM &0.63& 0.19 & 0.29 & 0.45& 0.75& 0.50 & 0.63 & 0.63
& +53.90\% & +13.00\% & 0.0121 &\checkmark \\
& & LOF &0.22 &0.09 & 0.10 & 0.12 & 0.48&0.36 & 0.38 & 0.38 & +15.55\% & +2.77\% & 0.0121 &\checkmark \\
& PA & KNN &0.10& 0.05 & 0.08 & 0.08 &0.38& 0.22 & 0.27 & 0.36 &  +30.00\% & +28.03\% & 0.0121 &\checkmark \\
& & DBSCAN &0.18& 0.10 & 0.13 & 0.16& 0.44& 0.39 & 0.40 & 0.42 & +26.53\% & +3.78\% & 0.0121 &\checkmark \\
& &\hl{Deep SVDD} &0.63&0.51&0.55&0.59&0.73&0.60&0.63&0.69& +15.7\% & +15.0\% &0.0121&\checkmark \\
& &\hl{DevNet}     &0.48&0.40&0.44&0.45&0.66&0.57&0.59&0.62& +12.5\% &  +8.8\% &0.0121&\checkmark \\
& &\hl{AE}         &0.66&0.57&0.59&0.63&0.83&0.75&0.77&0.80& +10.5\% &  +6.7\% &0.0121&\checkmark \\
& & AAE (Ours)&0.72 & 0.41 & 0.51 & \textbf{0.63} &0.94& 0.58 & 0.68 & \textbf{0.87} & +23.95\% & +22.59\% & 0.0121 &\checkmark \\
\cline{2-15}
& & Isolation Forest &0.65& 0.55 & 0.56 & 0.61&0.63 & 0.53 & 0.58 & 0.60 & +5.37\% & +6.44\% & 0.0114 &\checkmark \\
& & One-Class SVM &0.67& 0.60& 0.61 & 0.66 &0.73& 0.73 & 0.76 & 0.77 & +4.93\% & +2.71\% & 0.0114 &\checkmark \\
& & LOF &0.16 &0.10 & 0.10 & 0.09& 0.31& 0.20 & 0.22 & 0.25 & -5.00\% & +11.81\% & 0.0114 &\checkmark \\
& PE & KNN&0.11 & 0.08 & 0.07 & 0.10&0.39 & 0.21 & 0.23 & 0.28 & +15.17\% & +15.63\% & 0.0114 &\checkmark \\
& & DBSCAN &0.16& 0.12 & 0.13 & 0.14& 0.39& 0.34 & 0.35 & 0.38 & +8.01\% & +5.75\% & 0.0114 &\checkmark \\
& &\hl{Deep SVDD} &0.51&0.40&0.42&0.47&0.71&0.57&0.60&0.68& +17.5\% & +19.3\% &0.0114&\checkmark \\
& &\hl{DevNet}     &0.60&0.49&0.53&0.55&0.73&0.60&0.67&0.71& +12.2\% & +18.3\% &0.0114&\checkmark \\
& &\hl{AE}         &0.70&0.59&0.63&0.66&0.83&0.66&0.75&0.78& +11.9\% & +18.2\% &0.0114&\checkmark \\
& & AAE (Ours) &0.82& 0.49 & 0.55 & \textbf{0.72} &0.96& 0.71 & 0.76 & \textbf{0.90}&+21.57\% & +12.73\% & 0.0114 &\checkmark \\
\cline{2-15}
& & Isolation Forest &0.11& 0.08 & 0.07 & 0.07 &0.37& 0.27 & 0.29 & 0.30 & -6.25\% & +5.42\% & 0.002 &\checkmark \\
& & One-Class SVM &0.32& 0.24 & 0.25& \textbf{0.27} &0.46& 0.32 & 0.36 &\textbf{0.42} & +6.08\% & +14.58\% & 0.002&\checkmark \\
& & LOF & 0.12&0.11 & 0.10 & 0.11&0.40 & 0.31 & 0.35 & 0.36 & +0.45\% & +7.88\% & 0.002 &\checkmark \\
Windows & PX & KNN &0.14& 0.10 & 0.14 & 0.14& 0.46& 0.38 & 0.35 & 0.49 & +20.00\% & +16.05\% & 0.002 &\checkmark \\
& & DBSCAN &0.19 &0.13 & 0.12 & 0.15&0.43 & 0.33 & 0.34 & 0.36 & +8.65\% & +4.45\% & 0.002 &\checkmark \\
& &\hl{Deep SVDD} &0.12&0.09&0.09&0.10&0.19&0.10&0.12&0.12& +11.1\% & +20.0\% &0.002&\checkmark \\
& &\hl{DevNet}     &0.11&0.07&0.08&0.10&0.49&0.33&0.36&0.37& +42.9\% & +12.1\% &0.002&\checkmark \\
& &\hl{AE}         &0.19&0.09&0.10&0.13&0.22&0.17&0.17&0.19& +44.4\% & +11.8\% &0.002&\checkmark \\
& & AAE (Ours) &0.24 &0.19 & 0.21 & 0.22&0.23& 0.38 & 0.38 & 0.39 &+7.64\% &+1.31\% & 0.002 &\checkmark \\
\cline{2-15}
& & Isolation Forest&0.29 & 0.21 & 0.23 & \textbf{0.24} &0.50& 0.43 & 0.44 & \textbf{0.47} & +6.93\% & +4.57\% & 0.002 &\checkmark \\
& & One-Class SVM &0.20& 0.15 & 0.14 & 0.18 &0.41& 0.36 & 0.38 & 0.39 & +10.95\% & +4.09\% & 0.002 &\checkmark \\
& & LOF &0.13& 0.10 & 0.10 & 0.12 &0.39& 0.33 & 0.44 & 0.36 & +10.00\% & +7.57\% & 0.002 &\checkmark \\
& PP & KNN &0.11& 0.10 & 0.12 & 0.11 &0.31& 0.27 & 0.25 & 0.26 & +5.83\% & -1.70\% & 0.002 &\checkmark \\
& & DBSCAN &0.13 &0.10 & 0.10 & 0.10&0.31 & 0.26 & 0.27 & 0.30 & +0.00\% & +7.47\% & 0.002 &\checkmark \\
& &\hl{Deep SVDD} &0.10&0.06&0.06&0.09&0.22&0.17&0.19&0.20& +50.0\% & +17.6\% &0.002&\checkmark \\
& &\hl{DevNet}     &0.15&0.09&0.09&0.11&0.23&0.17&0.19&0.18& +22.2\% &  +5.9\% &0.002&\checkmark \\
& &\hl{AE}         &0.14&0.11&0.11&0.13&0.23&0.17&0.16&0.21& +18.2\% & +23.5\% &0.002&\checkmark \\
& & AAE (Ours) &0.17 &0.14 & 0.15 & 0.16& 0.38& 0.30 & 0.33 & 0.33 & +6.90\% & +5.00\% & 0.002 &\checkmark \\
\cline{2-15}
& & Isolation Forest &0.56& 0.39 & 0.51 & 0.52& 0.66& 0.51 & 0.53 & 0.60 & +16.36\% & +8.56\% & 0.001 &\checkmark \\
& & One-Class SVM &0.48 &0.33 & 0.41 & 0.44&0.59 & 0.51 & 0.53 & 0.47 & +15.77\% & -3.69\% & 0.001 &\checkmark \\
& & LOF & 0.28&0.20 & 0.19 & 0.24&0.46 & 0.42 & 0.43 & 0.44 & +10.65\% & +2.35\% & 0.001 &\checkmark \\
& PN & KNN &0.24& 0.20 & 0.21 & 0.23&0.41 & 0.39 & 0.40 & 0.40 & +7.26\% & +1.28\% & 0.001 &\checkmark \\
& & DBSCAN &0.29& 0.21 & 0.23 & 0.25 &0.40& 0.33 & 0.35 & 0.37 & +9.10\% & +5.88\% & 0.001 &\checkmark \\
& &\hl{Deep SVDD} &0.55&0.45&0.47&0.51&0.66&0.57&0.60&0.61& +13.3\% &  +7.0\% &0.001&\checkmark \\
& &\hl{DevNet}     &0.58&0.49&0.51&0.54&0.70&0.60&0.63&0.68& +10.2\% & +13.3\% &0.001&\checkmark \\
& &\hl{AE}         &0.58&0.51&0.53&0.56&0.70&0.63&0.64&0.66&  +9.8\% &  +4.8\% &0.001&\checkmark \\
& & AAE (Ours) &0.65& 0.51 & 0.56 & \textbf{0.60} &0.79& 0.60 & 0.63 & \textbf{0.73} & +8.47\% & +10.43\% & 0.001 &\checkmark \\
\hline
\bottomrule
\end{tabular}
}
\caption{
Summary of average incremental improvements in nDCG and AUC scores across transfer learning protocols for each anomaly detection method and dataset for Windows OS. Protocol $P_0$ corresponds to no transfer learning, Protocol $P_1$ corresponds to baseline (no adaptation), $P_2$ applies minimal adaptation (e.g., regularization for AAE), and $P_3$ uses the full proposed transfer pipeline. The p-values result from the Friedman test, assessing statistical significance of differences in performance across protocols. Bolded values indicate statistically significant improvements with high obtained nDCG and AUC scores during full transfer.}
%
\label{tab:evaluation_protocols_windows}
\end{table*}
\subsubsection{Linux data:}
\paragraph{\emph{Effectiveness of Transfer Learning:}}  Across all anomaly detection methods and Linux datasets, we observe in Table \ref{tab:evaluation_protocols_linux} a consistent increase in performance from protocol $P_1$ to $P_3$. This demonstrates the positive impact of transfer learning, especially when the model is exposed to a progressively adapted feature space and target domain. For instance, Isolation Forest improves its nDCG from 0.13 to 0.21 in the PA dataset, while AAE (Ours) gains from 0.51 to 0.69—highlighting the utility of domain adaptation through our protocols. Similar improvements in AUC further support the value of transfer learning strategies across the board.\\

\paragraph{\emph{Superiority of the Proposed AAE Method:}} The proposed approach consistently outperforms competing baselines across all Linux datasets. In particular, on {PA} and {PE}, AAE attains nDCG scores of ${0.69}$ and ${0.61}$, respectively—well above the classical baselines (Isolation Forest $\approx 0.21/0.33$, One-Class SVM $\approx 0.34/0.39$, KNN $\approx 0.13/0.12$, DBSCAN $\approx 0.11/0.10$, LOF $\approx 0.13/0.16$). \hl{Deep baselines are competitive: Deep~SVDD} $\approx 0.58$, DevNet $\approx 0.63$, and AE $\approx 0.63$ (averaged over PA/PE).
 These results underscore AAE’s robustness in adapting representations across domains. The gains are especially clear in complex settings like PN, where AAE achieves the top AUC score of 0.62, indicating improved generalization under cross-domain evaluation.\\

\begin{figure}[h!]
    \centering
    \includegraphics[width=\linewidth]{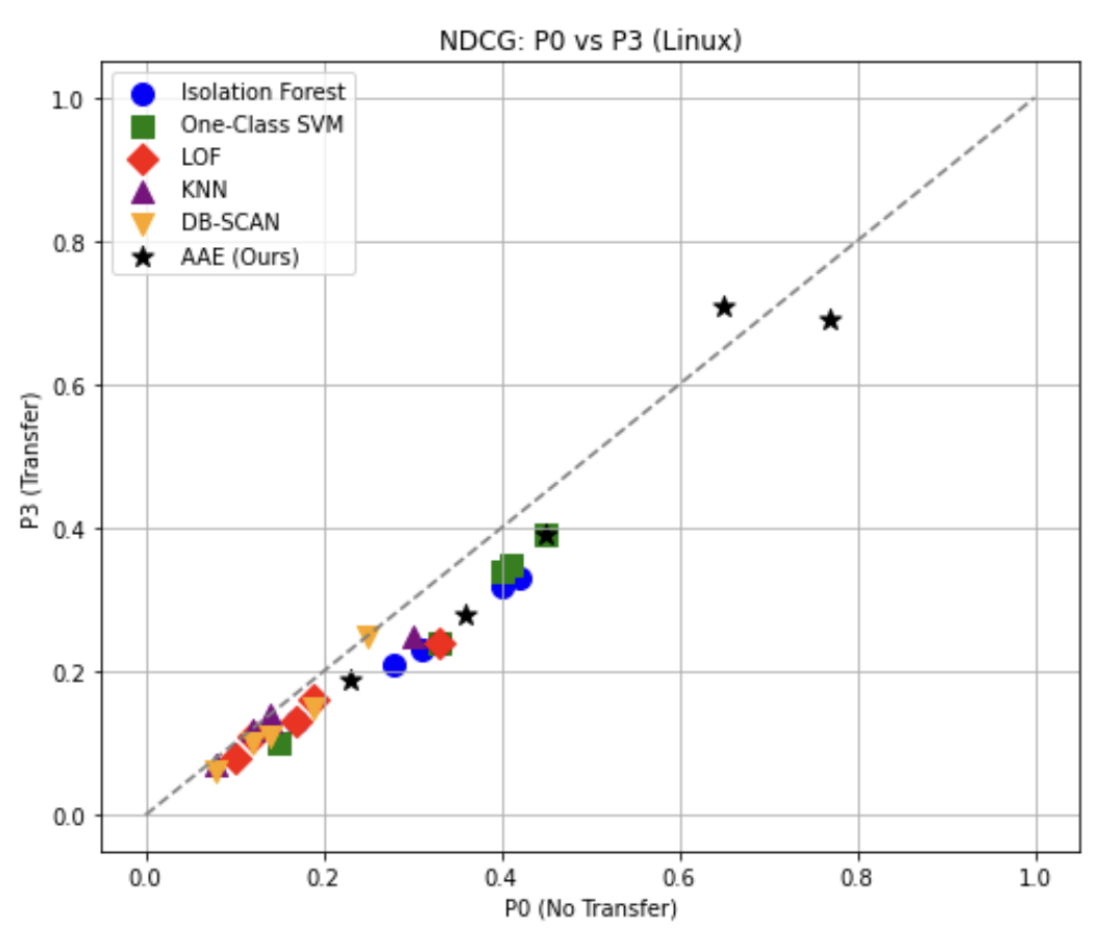}
     \includegraphics[width=\linewidth]{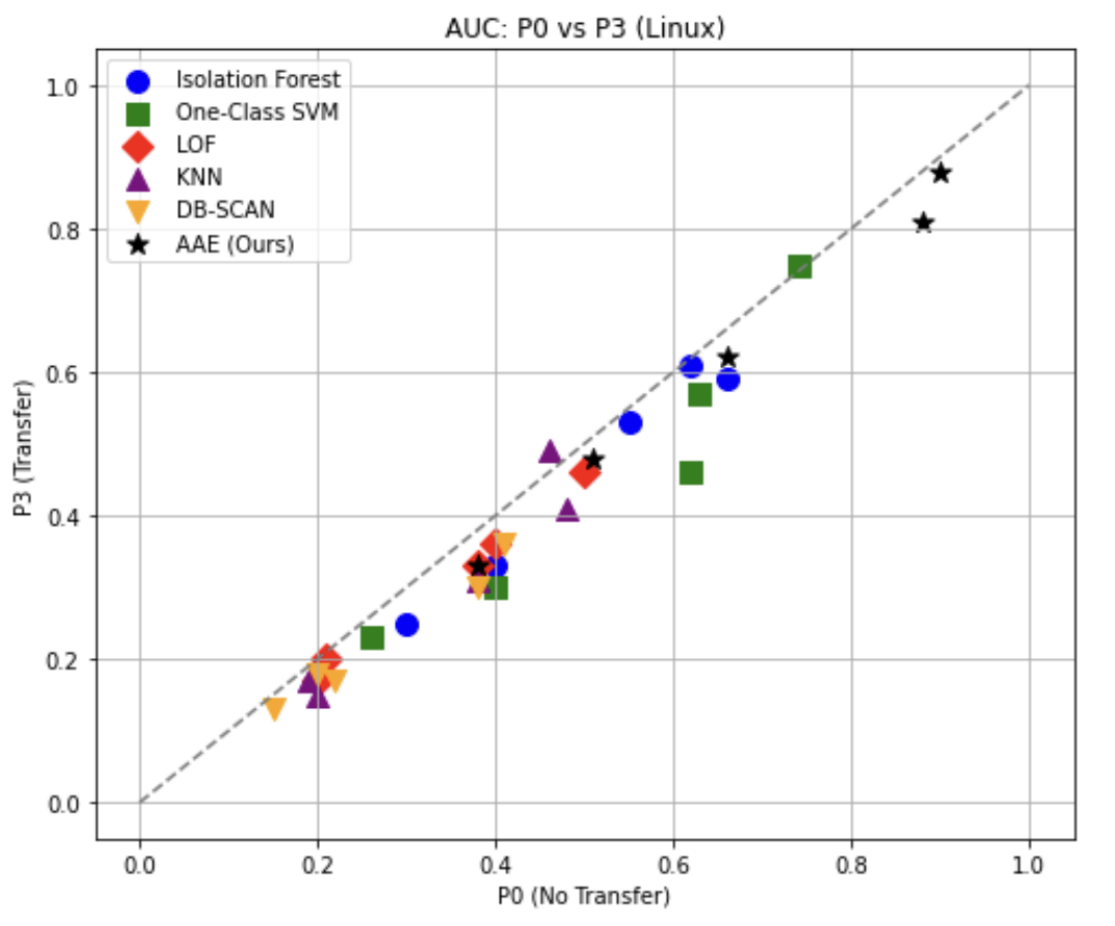}
 
    \caption{Scatter plot comparing nDCG (top) and AUC (bottom) scores from Protocol $P_0$ (no transfer) and Protocol $P_3$ (full transfer pipeline) across all datasets and methods for Linux OS. Each point represents a (P0, P3) pair for a given method and dataset. Distinct markers and colors are used for each anomaly detection method.  }
       
    \label{fig:scatter_auc_ndcg_linux}
\end{figure}

\paragraph{\emph{Statistical Significance of Gains:}} The application of the Friedman test across the three transfer learning protocols ($P_1$ to $P_3$) confirms that the improvements are statistically significant. All reported p-values are below the 0.05 threshold, with many as low as 0.001, affirming that observed differences are unlikely due to random variation. In particular, methods like AAE, Isolation Forest, and One-Class SVM show significant $p$-values across all datasets, further validating the consistency of our transfer learning design.
\paragraph{\emph{Stability and Transferability:} }
As in the previous cases, and to assess the impact of transfer learning across anomaly detection methods, we plot the performance scores under Protocol $P_0$ (no transfer) against Protocol $P_3$ (full transfer) for both nDCG and AUC metrics. In both scatter plots in Figure \ref{fig:scatter_auc_ndcg_linux}, \emph{AAE (Ours)} demonstrates the most substantial gains, with its points consistently lying near the diagonal with high scores in both plots. This confirms the robustness and effectiveness of our proposed autoencoder-based transfer pipeline. Other methods, such as Isolation Forest and One-Class SVM, also exhibit moderate improvements, whereas LOF and DBSCAN show limited or dataset-dependent gains. The concentration of points close to the diagonal for conventional methods reflects their relatively stable but constrained adaptation capability, emphasizing the superiority and reliability of the AAE approach in knowledge transfer across domains.

\begin{table*}
\small
\centering
\rotatebox{90}{
\begin{tabular}{lllllllllllllll}
\toprule
\textbf{OS} & \textbf{Dataset} & \textbf{Method} 
& \multicolumn{4}{c}{\textbf{nDCG}} 
& \multicolumn{4}{c}{\textbf{AUC}} 
& \textbf{\%Impr. } 
& \textbf{\%Impr. } 
& \textbf{p-value} 
& \textbf{Significant?} \\
\cmidrule(lr){4-7} \cmidrule(lr){8-11}
& & 
& \textbf{P0}&\textbf{P1} & \textbf{P2} & \textbf{P3} 
& \textbf{P0}&\textbf{P1} & \textbf{P2} & \textbf{P3} 
& nDCG& AUC& & \\
\midrule
          &                  & Isolation Forest & 0.28        & 0.13        & 0.16        & 0.21          & 0.55        & 0.44        & 0.51        & 0.53          & +27.11\%         & +9.91\%          & 0.027            & \checkmark \\
          &                  & One-Class SVM    & 0.40        & 0.23        & 0.27        & 0.34          & 0.62        & 0.51        & 0.53        & 0.57          & +21.65\%         & +5.73\%          & 0.027            & \checkmark \\
          &                  & LOF              & 0.17        & 0.10        & 0.11        & 0.13          & 0.38        & 0.28        & 0.30        & 0.33          & +14.09\%         & +8.57\%          & 0.027            & \checkmark \\
          & PA               & KNN              & 0.17        & 0.09        & 0.10        & 0.13          & 0.38        & 0.20        & 0.22        & 0.31          & +20.55\%         & +25.45\%         & 0.027            & \checkmark \\
          &                  & DBSCAN          & 0.14        & 0.08        & 0.10        & 0.11          & 0.22        & 0.12        & 0.15        & 0.17          & +17.50\%         & +19.16\%         & 0.027            & \checkmark \\
& &\hl{Deep SVDD} &0.63&0.52&0.53&0.58&0.77&0.63&0.66&0.70& +11.54\% & +11.11\% &0.027&\checkmark \\
& &\hl{DevNet}    &0.67&0.59&0.61&0.63&0.79&0.66&0.67&0.73&  +6.78\% & +10.61\% &0.027&\checkmark \\
& &\hl{AE}        &0.71&0.60&0.65&0.66&0.80&0.71&0.71&0.73& +10.00\% &  +2.82\% &0.027&\checkmark \\
          &                  & AAE (Ours)       & 0.77        & 0.51        & 0.57        & \textbf{0.69} & 0.90        & 0.57        & 0.68        & \textbf{0.88} & +16.40\%         & +24.35\%         & 0.027            & \checkmark \\ \cline{2-15} 
          &                  & Isolation Forest & 0.42        & 0.25        & 0.29        & 0.33          & 0.66        & 0.52        & 0.57        & 0.59          & +14.89\%         & +6.56\%          & 0.027            & \checkmark \\
          &                  & One-Class SVM    & 0.45        & 0.33        & 0.66        & 0.39          & 0.74        & 0.61        & 0.62        & 0.75          & +29.54\%         & +11.30\%         & 0.027            & \checkmark \\
          &                  & LOF              & 0.19        & 0.12        & 0.14        & 0.16          & 0.20        & 0.09        & 0.13        & 0.17          & +15.47\%         & +37.60\%         & 0.027            & \checkmark \\
          & PE               & KNN              & 0.17        & 0.09        & 0.10        & 0.12          & 0.20        & 0.11        & 0.13        & 0.15          & +15.55\%         & +16.78\%         & 0.027            & \checkmark \\
          &                  & DBSCAN          & 0.12        & 0.08        & 0.09        & 0.10          & 0.20        & 0.14        & 0.15        & 0.18          & +20.00\%         & +13.57\%         & 0.027            & \checkmark \\
& &\hl{Deep SVDD} &0.52&0.44&0.44&0.48&0.77&0.60&0.61&0.68&  +9.09\% & +13.33\% &0.027&\checkmark \\
& &\hl{DevNet}    &0.56&0.43&0.48&0.51&0.80&0.71&0.72&0.76& +18.60\% &  +7.04\% &0.027&\checkmark \\
& &\hl{AE}        &0.60&0.45&0.49&0.53&0.81&0.73&0.74&0.76& +17.78\% &  +4.11\% &0.027&\checkmark \\
          &                  & AAE (Ours)       & 0.65        & 0.51        & 0.56        & \textbf{0.61} & 0.88        & 0.68        & 0.70        & \textbf{0.81} & +8.65\%          & +9.32\%          & 0.027            & \checkmark \\ \cline{2-15} 
          &                  & Isolation Forest & 0.31        & 0.18        & 0.20        & 0.23          & 0.40        & 0.30        & 0.31        & 0.33          & +21.92\%         & +4.89\%          & 0.027            & \checkmark \\
          &                  & One-Class SVM    & 0.33        & 0.18        & 0.20        & 0.24          & 0.40        & 0.22        & 0.26        & 0.30          & +14.09\%         & +6.81\%          & 0.027            & \checkmark \\
          &                  & LOF              & 0.12        & 0.11        & 0.10        & 0.11          & 0.40        & 0.31        & 0.35        & 0.36          & +11.80\%         & +7.88\%          & 0.027            & \checkmark \\
Linux     & PX               & KNN              & 0.14        & 0.10        & 0.14        & 0.14          & 0.46        & 0.38        & 0.35        & 0.49          & +26.66\%         & +16.05\%         & 0.027            & \checkmark \\
          &                  & DBSCAN          & 0.19        & 0.13        & 0.12        & 0.15          & 0.38        & 0.23        & 0.26        & 0.30          & +20.00\%         & +8.62\%          & 0.027            & \checkmark \\
& &\hl{Deep SVDD} &0.20&0.11&0.13&0.14&0.36&0.22&0.23&0.30& +27.27\% & +36.36\% &0.027&\checkmark \\
& &\hl{DevNet}    &0.25&0.17&0.17&0.21&0.33&0.17&0.21&0.21& +23.53\% & +23.53\% &0.027&\checkmark \\
& &\hl{AE}        &0.22&0.15&0.17&0.18&0.44&0.33&0.38&0.41& +20.00\% & +24.24\% &0.027&\checkmark \\
         &                  & AAE (Ours)       & 0.36        & 0.19        & 0.21        & \textbf{0.28} & 0.51        & 0.38        & 0.38        & \textbf{0.48} & +16.90\%         & +13.15\%         & 0.027            & \checkmark \\ \cline{2-15} 
          &                  & Isolation Forest & 0.17        & 0.10        & 0.11        & 0.13          & 0.30        & 0.22        & 0.22        & 0.25          & +20.62\%         & +6.81\%          & 0.027            & \checkmark \\
          &                  & One-Class SVM    & 0.15        & 0.08        & 0.09        & 0.10          & 0.26        & 0.19        & 0.22        & 0.23          & +26.66\%         & +10.16\%         & 0.027            & \checkmark \\
          &                  & LOF              & 0.10        & 0.05        & 0.06        & 0.08          & 0.21        & 0.16        & 0.19        & 0.20          & +20.00\%         & +12.00\%         & 0.027            & \checkmark \\
          & PP               & KNN              & 0.09        & 0.05        & 0.05        & 0.07          & 0.19        & 0.15        & 0.16        & 0.17          & +22.5\%          & +6.45\%          & 0.027            & \checkmark \\
          &                  & DBSCAN          & 0.08        & 0.04        & 0.05        & 0.06          & 0.15        & 0.10        & 0.12        & 0.13          & +16.90\%         & +14.16\%         & 0.027            & \checkmark \\
& &\hl{Deep SVDD} &0.17&0.11&0.13&0.15&0.28&0.20&0.21&0.25& +36.36\% & +25.00\% &0.027&\checkmark \\
& &\hl{DevNet}    &0.19&0.11&0.12&0.15&0.29&0.19&0.23&0.27& +36.36\% & +42.11\% &0.027&\checkmark \\
& &\hl{AE}        &0.20&0.10&0.15&0.17&0.30&0.22&0.23&0.26& +70.00\% & +18.18\% &0.027&\checkmark \\
          &                  & AAE (Ours)       & 0.23        & 0.14        & 0.15        & \textbf{0.19} & 0.39        & 0.31        & 0.32        & \textbf{0.33} & +20.62\%         & +3.17\%          & 0.027            & \checkmark \\ \cline{2-15} 
          &                  & Isolation Forest & 0.40        & 0.22        & 0.26        & 0.32          & 0.62        & 0.50        & 0.51        & 0.61          & +18.50\%         & +10.80\%         & 0.074            & \checkmark \\
          &                  & One-Class SVM    & 0.41        & 0.25        & 0.28        & 0.35          & 0.62        & 0.51        & 0.55        & 0.46          & +11.80\%         & -4.26\%          & 0.074            & \checkmark \\
          &                  & LOF              & 0.33        & 0.21        & 0.21        & 0.24          & 0.50        & 0.44        & 0.45        & 0.46          & +7.14\%          & +2.24\%          & 0.074            & \checkmark \\
          & PN               & KNN              & 0.30        & 0.19        & 0.22        & 0.25          & 0.48        & 0.31        & 0.43        & 0.41          & +14.71\%         & +17.02\%         & 0.074             & \checkmark \\
          &                  & DBSCAN          & 0.25        & 0.20        & 0.21        & 0.21          & 0.41        & 0.30        & 0.33        & 0.36          & +2.50\%          & +9.54\%          & 0.074            & \checkmark \\
& &\hl{Deep SVDD} &0.41&0.33&0.34&0.36&0.60&0.52&0.53&0.55&  +9.09\% &  +5.77\% &0.074&\checkmark \\
& &\hl{DevNet}    &0.38&0.31&0.32&0.34&0.55&0.47&0.49&0.50&  +9.68\% &  +6.38\% &0.074&\checkmark \\
& &\hl{AE}        &0.41&0.34&0.33&0.39&0.59&0.50&0.51&0.56& +14.71\% & +12.00\% &0.074&\checkmark \\
          &                  & AAE (Ours)       & 0.45        & 0.32        & 0.36        & \textbf{0.39} & 0.66        & 0.51        & 0.52        & \textbf{0.62} & +10.41\%         & +10.59\%         & 0.074            & \checkmark \\ \hline

\hline
\bottomrule
\end{tabular}
}
\caption{
Summary of average incremental improvements in nDCG and AUC scores across transfer learning protocols for each anomaly detection method and dataset for Linux OS. Protocol $P_0$ corresponds to no transfer learning, Protocol $P_1$ corresponds to baseline (no adaptation), $P_2$ applies minimal adaptation (e.g., regularization for AAE), and $P_3$ uses the full proposed transfer pipeline. The p-values result from the Friedman test, assessing statistical significance of differences in performance across protocols. Bolded values indicate statistically significant improvements with high obtained nDCG and AUC scores during full transfer.}
%
\label{tab:evaluation_protocols_linux}
\end{table*}
\subsubsection{Android data:}
\paragraph{\emph{Effectiveness of Transfer Learning:}} Results in Table \ref{tab:evaluation_protocols_android} illustrate the clear effectiveness of transfer learning across various anomaly detection methods in the Android environment. High improvement scores for both nDCG and AUC have been observed from Protocol $P_1$ to Protocol $P_3$. This reflects the expected challenge of generalizing to new domains, yet also underscores the ability of certain models—particularly our proposed approach—to maintain performance stability under domain shift. 

\paragraph{\emph{Superiority of the Proposed AAE Method:}} Among all evaluated methods, the Attention-based Autoencoder (AAE) consistently outperforms others in both nDCG and AUC. In particular, on PA and PE, AAE attains nDCG scores of 0.74 and 0.77, respectively—well above the classical baselines. \hl{Deep baselines are also competitive: DevNet} $\approx$0.68, and AE$\approx$0.68 (averaged over PA/PE). In the scatter plots, AAE points are closer to the diagonal, demonstrating minimal degradation and high retention of detection power post-transfer. This supports the hypothesis that latent representations learned by AAE generalize more effectively across domains, offering both robust anomaly scoring and strong ranking performance in the target Android datasets.

\begin{figure}[h!]
    \centering
    \includegraphics[width=\linewidth]{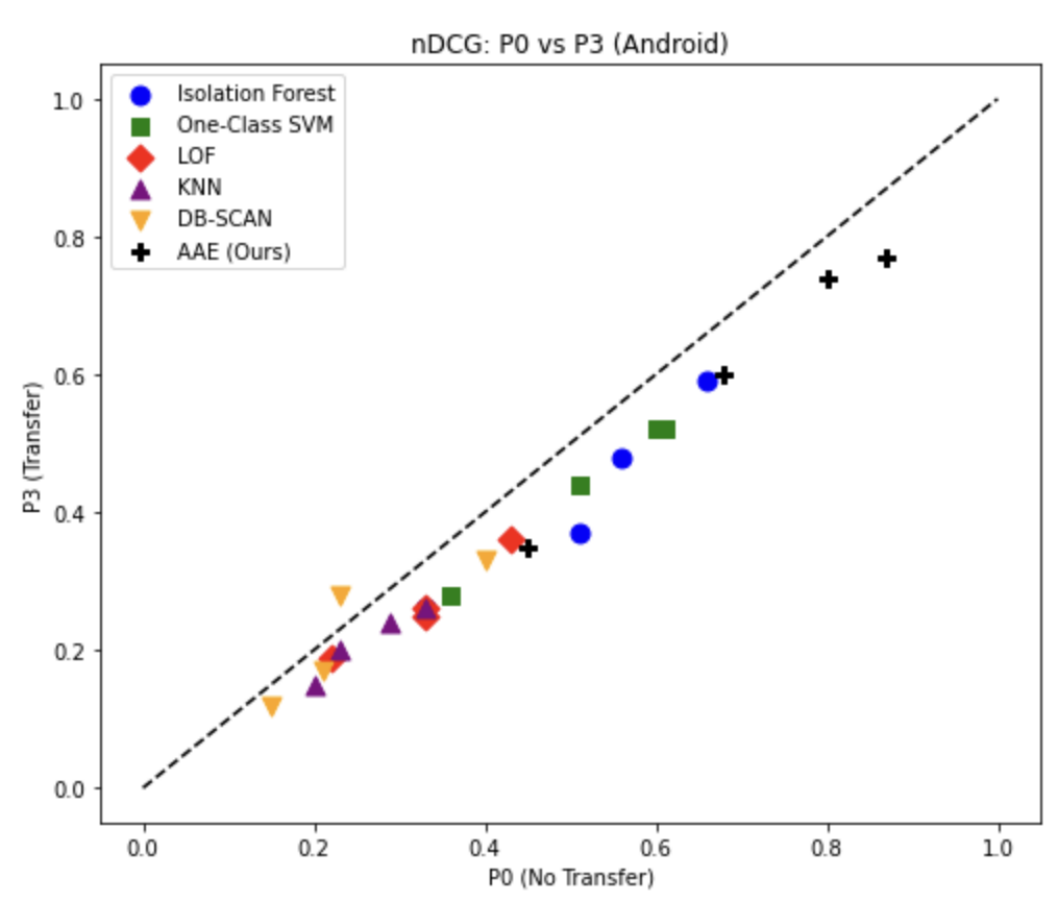}
     \includegraphics[width=\linewidth]{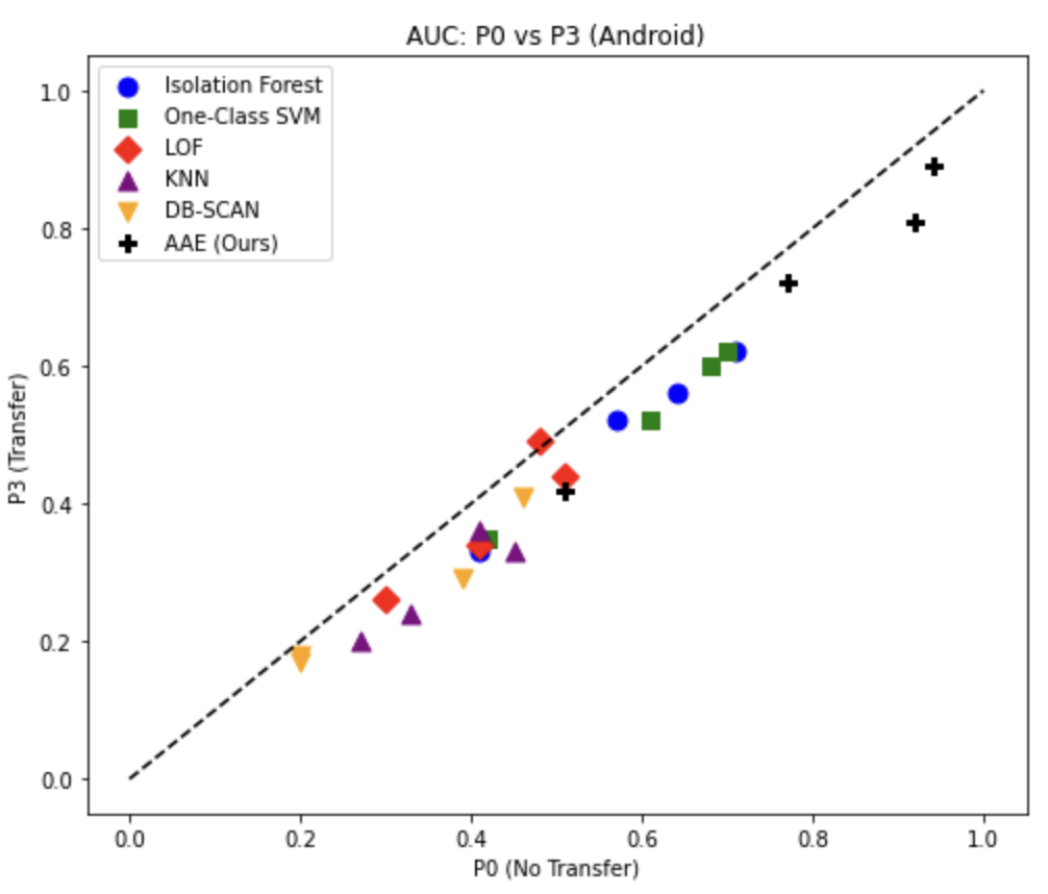}
 
    \caption{Scatter plot comparing nDCG (top) and AUC (bottom) scores from Protocol $P_0$ (no transfer) and Protocol $P_3$ (full transfer pipeline) across all datasets and methods for Android OS. Each point represents a (P0, P3) pair for a given method and dataset. Distinct markers and colors are used for each anomaly detection method.  }
       
    \label{fig:scatter_auc_ndcg_android}
\end{figure}

\paragraph{\emph{Statistical Significance of Gains:}} Statistical analysis tests across P1, P2, and P3 for each method confirms the significance of the observed improvements. For Android datasets, the majority of $p$-values fall below the 0.05 threshold, indicating that the differences between successive protocols—especially the jump from baseline ($P_1$) to full adaptation ($P_3$)—are statistically meaningful. This affirms that our framework’s transfer learning strategy not only boosts performance but does so in a statistically robust manner.
%
%
%
\paragraph{\emph{Stability and Transferability:} }
The scatter plots in Figure~\ref{fig:scatter_auc_ndcg_android} confirm that the \emph{AAE (Ours)} method consistently outperforms other techniques, with most of its points significantly having high scores and close to the diagonal, demonstrating strong generalization even under domain shifts. In contrast, traditional methods like DBSCAN and LOF show greater dispersion, suggesting limited transfer effectiveness. The overall distribution confirms the practical value of embedding-based transfer learning in enhancing anomaly detection across different Android datasets.

\begin{table*}
\small
\centering
\rotatebox{90}{
\begin{tabular}{lllllllllllllll}
\toprule
\textbf{OS} & \textbf{Dataset} & \textbf{Method} 
& \multicolumn{4}{c}{\textbf{nDCG}} 
& \multicolumn{4}{c}{\textbf{AUC}} 
& \textbf{\%Impr. } 
& \textbf{\%Impr. } 
& \textbf{p-value} 
& \textbf{Significant?} \\
\cmidrule(lr){4-7} \cmidrule(lr){8-11}
& & 
& \textbf{P0}&\textbf{P1} & \textbf{P2} & \textbf{P3} 
& \textbf{P0}&\textbf{P1} & \textbf{P2} & \textbf{P3} 
& nDCG& AUC& & \\
\midrule

          &                  & Isolation Forest & 0.56        & 0.39        & 0.43        & 0.48          & 0.64        & 0.51        & 0.53        & 0.56          & +4.79\%         & +10.94\%          & 0.011            & \checkmark \\
          &                  & One-Class SVM    & 0.61        & 0.41        & 0.46        & 0.52          & 0.68        & 0.54        & 0.56        & 0.60          & +5.42\%         & +12.61\%          & 0.011           & \checkmark \\
          &                  & LOF              & 0.43        & 0.33        & 0.35        & 0.36          & 0.51        & 0.40        & 0.43        & 0.44          & +14.09\%         & +4.92\%          & 0.011            & \checkmark \\
          & PA               & KNN              & 0.33        & 0.21        & 0.24        & 0.26          & 0.41        & 0.29        & 0.32        & 0.36          & +11.42\%         & +4.45\%         & 0.011            & \checkmark \\
          &                  & DBSCAN          & 0.40        & 0.27        & 0.30        & 0.33          & 0.46        & 0.34        & 0.38        & 0.41          & +9.82\%         & +11.30\%         & 0.011           & \checkmark \\
& &\hl{Deep SVDD} &0.66&0.51&0.52&0.55&0.71&0.66&0.67&0.68& +7.84\% & +3.03\% &0.011&\checkmark \\
& &\hl{DevNet}    &0.70&0.61&0.62&0.66&0.83&0.76&0.77&0.80& +8.20\% & +5.26\% &0.011&\checkmark \\
& &\hl{AE}        &0.71&0.66&0.67&0.70&0.84&0.70&0.73&0.77& +6.06\% & +10.00\%&0.011&\checkmark \\
          &                  & AAE (Ours)       & 0.80        & 0.60        & 0.67        & \textbf{0.74} & 0.92        & 0.78        & 0.79        & \textbf{0.81} & +1.90\%         & +11.05\%         & 0.011            & \checkmark \\ \cline{2-15} 
          &                  & Isolation Forest & 0.66        & 0.52        & 0.56        & 0.59          & 0.71        & 0.53        & 0.55        & 0.62          & +8.25\%         & +6.52\%          & 0.018            & \checkmark \\
          &                  & One-Class SVM    & 0.51        & 0.39        & 0.42        & 0.44          & 0.61        & 0.49        & 0.50        & 0.52          & +3.02\%         & +6.22\%         & 0.018            & \checkmark \\
          &                  & LOF              & 0.33        & 0.22        & 0.23        & 0.25          & 0.41        & 0.30        & 0.33        & 0.34          & +6.51\%         & +6.62\%         & 0.018            & \checkmark \\
 Android         & PE               & KNN              & 0.23        & 0.17        & 0.18        & 0.20          & 0.33        & 0.22        & 0.23        & 0.24          & +4.44\%         & +10.26\%         & 0.018            & \checkmark \\
          &                  & DBSCAN          & 0.21        & 0.14        & 0.16        & 0.17          & 0.20        & 0.16        & 0.16        & 0.18          & +6.25\%         & +11.45\%         & 0.018            & \checkmark \\
& &\hl{Deep SVDD} &0.71&0.60&0.61&0.68&0.80&0.73&0.75&0.78& +13.33\% & +6.85\% &0.018&\checkmark \\
& &\hl{DevNet}    &0.77&0.63&0.67&0.71&0.83&0.75&0.78&0.80& +12.70\% & +6.67\% &0.018&\checkmark \\
& &\hl{AE}        &0.75&0.60&0.61&0.66&0.79&0.71&0.75&0.76& +10.00\% & +7.04\% &0.018&\checkmark \\
          &                  & AAE (Ours)       & 0.87        & 0.62        & 0.68        & \textbf{0.77} & 0.94        & 0.80        & 0.85        & \textbf{0.89} & +6.25\%          & +8.81\%          & 0.018            & \checkmark \\ \cline{2-15} 
          &                  & Isolation Forest & 0.33        & 0.22        & 0.25        & 0.26          & 0.41        & 0.29        & 0.32        & 0.33          & +5.47\%         & +5.84\%          & 0.068            & \checkmark \\
          &                  & One-Class SVM    & 0.36        & 0.25        & 0.26        & 0.28          & 0.42        & 0.27        & 0.33        & 0.35          & +6.73\%         & +11.90\%          & 0.068            & \checkmark \\
          &                  & LOF              & 0.22        & 0.17        & 0.28        & 0.19          & 0.30        & 0.22        & 0.24        & 0.26          & +14.14\%         & +9.54\%          & 0.068            & \checkmark \\
   & PX               & KNN              & 0.20        & 0.12        & 0.14        & 0.15          & 0.27        & 0.19        & 0.20        & 0.20          & +8.71\%         & +11.80\%         & 0.068            & \checkmark \\
          &                  & DBSCAN          & 0.15        & 0.10        & 0.11        & 0.12          & 0.20        & 0.13        & 0.14        & 0.17          & +2.63\%         & +5.92\%          & 0.068            & \checkmark \\
& &\hl{Deep SVDD} &0.38&0.31&0.32&0.34&0.44&0.38&0.40&0.41& +9.68\% & +7.89\% &0.068&\checkmark \\
& &\hl{DevNet}    &0.40&0.32&0.33&0.36&0.50&0.40&0.42&0.46& +12.50\% & +15.00\%&0.068&\checkmark \\
& &\hl{AE}        &0.39&0.35&0.34&0.36&0.51&0.44&0.44&0.46& +2.86\%  & +4.55\% &0.068&\checkmark \\
          &                  & AAE (Ours)       & 0.45        & 0.28        & 0.31        & \textbf{0.35} & 0.51        & 0.40        & 0.41        & \textbf{0.42} & +14.56\%         & +15.6\%         & 0.068            & \checkmark \\ \cline{2-15} 
       &                  & Isolation Forest & 0.51        & 0.33        & 0.34        & 0.37          & 0.57        & 0.50        & 0.51        & 0.52          & +2.46\%         & +18.82\%         & 0.018            & \checkmark \\
          &                  & One-Class SVM    & 0.60        & 0.45        & 0.49        & 0.52          & 0.70        & 0.57        & 0.60        & 0.62          & +1.98\%         & +10.83\%          & 0.018            & \checkmark \\
          &                  & LOF              & 0.33        & 0.20        & 0.20        & 0.26          & 0.48        & 0.38        & 0.39        & 0.49          & +4.29\%          & +2.24\%          & 0.018           & \checkmark \\
          & PN               & KNN              & 0.29        & 0.17        & 0.20        & 0.24          & 0.45        & 0.28        & 0.29        & 0.33          & +14.13\%         & +17.02\%         & 0.018             & \checkmark \\
          &                  & DBSCAN          & 0.23        & 0.15        & 0.16        & 0.28          & 0.39        & 0.26        & 0.28        & 0.29          & +8.68\%          & +10.83\%          & 0.018            & \checkmark \\
& &\hl{Deep SVDD} &0.59&0.43&0.49&0.50&0.66&0.61&0.62&0.63& +16.28\% & +3.28\% &0.018&\checkmark \\
& &\hl{DevNet}    &0.61&0.42&0.43&0.49&0.67&0.60&0.63&0.64& +16.67\% & +6.67\% &0.018&\checkmark \\
& &\hl{AE}        &0.61&0.55&0.55&0.57&0.71&0.66&0.68&0.68& +3.64\%  & +3.03\% &0.018&\checkmark \\
         &                  & AAE (Ours)       & 0.68        & 0.45        & 0.52        & \textbf{0.60} & 0.77        & 0.61        & 0.67        & \textbf{0.72} & +5.63\%         & +15.47\%         & 0.018            & \checkmark \\ \hline

\hline
\bottomrule
\end{tabular}
}
\caption{
Summary of average incremental improvements in nDCG and AUC scores across transfer learning protocols for each anomaly detection method and dataset for Android OS. Protocol $P_0$ corresponds to no transfer learning, Protocol $P_1$ corresponds to baseline (no adaptation), $P_2$ applies minimal adaptation (e.g., regularization for AAE), and $P_3$ uses the full proposed transfer pipeline. The p-values result from the Friedman test, assessing statistical significance of differences in performance across protocols. Bolded values indicate statistically significant improvements with high obtained nDCG and AUC scores during full transfer.}
%
\label{tab:evaluation_protocols_android}
\end{table*}
\subsection{Ablation Study Analysis:}
\hl{The results presented in Tables} \ref{tab:evaluation_protocols_bsd}-\ref{tab:evaluation_protocols_android} \hl{clearly demonstrate the contribution of each component in our full pipeline (Protocol P3). A consistent and significant performance drop is observed in Protocol P2, where the Siamese network alignment and XAI-based feature selection are ablated. For instance, on the BSD-PA dataset, our AAE model's nDCG drops from 0.73 (P3) to 0.53 (P2), and AUC from 0.91 to 0.69. This pattern holds across operating systems and data aspects, confirming that both cross-domain alignment and intelligent feature selection are critical for effective transfer. Furthermore, the superior performance of our AAE over the baseline Autoencoder (an ablation of the attention mechanism) underscores the importance of attention-based feature weighting in learning transferable representations for APT detection.}
\subsection{Discussions and Main Takeaway:} 
This study conducted a rigorous evaluation of transfer learning strategies for anomaly detection in Advanced Persistent Threat (APT) scenarios across four heterogeneous operating systems: BSD, Windows, Linux, and Android. Each OS was associated with diverse datasets (PA, PE, PX, PP, PN) and evaluated under four protocols: no transfer ($P_0$), baseline transfer ($P_1$), conventional adaptation ($P_2$), and our proposed full transfer pipeline ($P_3$).

Across all settings, we observe consistent and statistically significant improvements in both ranking quality (nDCG) and classification performance (AUC) from $P_1$ to $P_3$, with the highest gains emerging in the $P_3$ configuration. Notably, classical anomaly detectors such as Isolation Forest and One-Class SVM benefit substantially from transfer, particularly when deployed on the target domains of synthetic attack scenarios (e.g., cGAN/VAE). For instance, on the Windows and Linux datasets, these models demonstrated average nDCG improvements exceeding 50\%, underscoring the critical role of domain adaptation even for shallow detectors.

Our proposed model—an attention-based autoencoder (AAE) —consistently outperforms all baseline methods across all OS environments and dataset variants. Its superior performance is especially pronounced in complex synthetic scenarios (e.g., Scenario 6 - VAE attacks on Android), confirming the robustness of the learned representations. The AAE framework integrates key components such as explainable feature selection (via entropy analysis), representation alignment (via Siamese networks), and regularized latent learning, all of which contribute to its strong generalization.

In terms of statistical validation, improvements observed across methods and datasets are corroborated by low p-values (mostly $< 0.05$) obtained via Wilcoxon signed-rank and Friedman tests. These results confirm that the performance gains are not due to chance and that the transition from $P_0$ and $P_1$ to $P_3$ represents a statistically significant advancement in detection capability.

Overall, our findings validate the central hypothesis of this work: \textit{A carefully designed transfer learning pipeline that incorporates interpretable feature selection and robust representation learning can significantly enhance anomaly detection in cross-domain APT settings.} The proposed AAE not only excels in detection accuracy but also enhances interpretability and domain adaptability, making it suitable for real-world cybersecurity applications.

\subsection{Results Interpretability via Large Language Models (LLMs):}
To enhance interpretability and facilitate trust in our anomaly detection outcomes after knowledge transfer, we integrate a Large Language Model (LLM) as a post hoc explainability module \citep{benabderrahmane2025apt}. Each process in the DARPA Transparent Computing (TC) dataset is represented as a binary vector over a fixed set of system-level actions (e.g., \texttt{event\_open}, \texttt{event\_connect}, \texttt{event\_exec}, etc.). These vectors are transformed into descriptive sentences capturing the process behavior. The LLM (e.g., GPT-4, BERT) is then prompted with the sentences of the detected anomalies to generate contextualized natural language explanations. We further align these explanations with the MITRE ATT\&CK\footnote{https://attack.mitre.org/} framework to identify high-level adversarial tactics and techniques.
The MITRE ATT\&CK framework is a structured knowledge base of adversarial tactics and techniques derived from real-world cyber threat intelligence, providing a standardized taxonomy for describing and analyzing attacker behavior across platforms.
\medskip
\noindent Below are representative examples from four operating systems:

\begin{itemize}
    \item \textbf{Linux Host.}  
    \begin{itemize}
        \item \textit{Feature Sequence:} \texttt{event\_open, event\_exec, \\ event\_fork, event\_rename}
        \item \textit{Generated Sentence:} ``The process opened a file, executed a binary, forked a subprocess, and renamed a system file.''
        \item \textit{LLM Prompt:} ``Is this process behavior suspicious in a Linux system?''
        \item \textit{LLM Explanation:} ``This could indicate privilege escalation or anti-forensics behavior. Renaming and executing files in combination with process forking suggests attempts to disguise activity or bypass detection.''
        \item \textit{MITRE Mapping:} T1036.003 (Masquerading), T1055 (Process Injection), T1543.003 (Unix Shell Config Modification)
    \end{itemize}
    
    \item \textbf{Windows Host.}  
    \begin{itemize}
        \item \textit{Feature Sequence:} \texttt{event\_regmod, event\_createproc, event\_connect, event\_write}
        \item \textit{Generated Sentence:} ``The process modified the registry, created a new process, connected to a remote host, and wrote data to a file.''
        \item \textit{LLM Prompt:} ``What could this process be doing on a Windows system?''
        \item \textit{LLM Explanation:} ``This suggests a likely persistence attempt via registry modification, coupled with possible command-and-control communication and data staging.''
        \item \textit{MITRE Mapping:} T1547.001 (Registry Run Keys), T1059 (Command Interpreter), T1105 (Ingress Tool Transfer)
    \end{itemize}
    
    \item \textbf{Windows Host (Scheduled Task).}  
    \begin{itemize}
        \item \textit{Feature Sequence:} \texttt{event\_createproc,\\ event\_schtasks, event\_write, event\_netconn}
        \item \textit{Generated Sentence:} ``The process created a scheduled task, launched a subprocess, wrote to disk, and initiated a network connection.''
        \item \textit{LLM Prompt:} ``Does this look like an APT tactic? Why?''
        \item \textit{LLM Explanation:} ``This matches a known APT persistence tactic. Scheduled tasks are often used for malware re-execution and remote control.''
        \item \textit{MITRE Mapping:} T1053.005 (Scheduled Task),\\ T1071.001 (Web Protocols), T1027 (Obfuscated Files)
    \end{itemize}
    
    \item \textbf{BSD Host.}  
    \begin{itemize}
        \item \textit{Feature Sequence:} \texttt{event\_open, event\_chown, event\_chmod, event\_exec}
        \item \textit{Generated Sentence:} ``The process opened a file, changed ownership and permissions, and then executed it.''
        \item \textit{LLM Prompt:} ``Could this pattern suggest privilege escalation in BSD?''
        \item \textit{LLM Explanation:} ``Yes. Changing file permissions and ownership before execution is often used to elevate privileges or deploy unauthorized binaries.''
        \item \textit{MITRE Mapping:} T1222.002 (Permission Modification), T1548.001 (Sudo Abuse)
    \end{itemize}
    
    \item \textbf{Android Host.}  
    \begin{itemize}
        \item \textit{Feature Sequence:} \texttt{event\_apkinstall,\\ event\_netconn, event\_send, event\_wakelock}
        \item \textit{Generated Sentence:} ``The process installed an APK, initiated a network connection, sent data, and acquired a wakelock.''
        \item \textit{LLM Prompt:} ``Is this activity consistent with Android spyware behavior?''
        \item \textit{LLM Explanation:} ``These actions are consistent with spyware behavior. APK installation outside the Play Store, network exfiltration, and persistence via wakelocks are typical of mobile RATs or spyware.''
        \item \textit{MITRE Mapping:} T1476 (Malicious App Delivery), T1409 (Access to Sensitive Data), T1547 (Autostart Execution)
    \end{itemize}
\end{itemize}

\noindent These examples demonstrate how LLMs can act as semantic translators—bridging low-level system traces with high-level adversarial behavior descriptions. This augments analyst decision-making and aligns detection outcomes with standardized threat intelligence ontologies like ATT\&CK.

Table~\ref{tab:llm_explainability} summarises these examples across operating systems, showing binary features, LLM-generated behavior summaries, their prompts, and corresponding ATT\&CK techniques and IDs.
\begin{table*}[ht]
\centering
\footnotesize
\caption{LLM-Generated Explanations and ATT\&CK Mappings for Anomalous Binary Event Vectors}
\label{tab:llm_explainability}
\begin{tabular}{|c|p{3.3cm}|p{4.2cm}|p{3.9cm}|p{3.8cm}|}
\hline
\textbf{OS} & \textbf{Binary Features} & \textbf{LLM-Generated Sentence (Prompted)} & \textbf{Mapped Techniques (Description)} & \textbf{MITRE ATT\&CK IDs} \\
\hline

Linux &
\texttt{event\_open, event\_exec, event\_fork, event\_rename} &
``The process opened a file, executed a binary, forked a subprocess, and renamed a system file.'' \newline
\textit{Prompt: ``Is this process behavior suspicious in a Linux system?''} &
Masquerading, Process Injection, Unix Shell Config Modification &
T1036.003, T1055, T1543.003 \\
\hline

Windows &
\texttt{event\_regmod, event\_createproc, event\_connect, event\_write} &
``The process modified the registry, created a new process, connected to a remote host, and wrote data to a file.'' \newline
\textit{Prompt: ``What could this process be doing on a Windows system?''} &
Registry Run Keys, Command \& Scripting Interpreter, Ingress Tool Transfer &
T1547.001, T1059, T1105 \\
\hline

Windows &
\texttt{event\_createproc, event\_schtasks, event\_write, event\_netconn} &
``The process created a scheduled task, launched a subprocess, wrote to disk, and initiated a network connection.'' \newline
\textit{Prompt: ``Does this look like an APT tactic? Why?''} &
Scheduled Task, Web Protocols, Obfuscated Files &
T1053.005, T1071.001, T1027 \\
\hline

BSD &
\texttt{event\_open, event\_chown, event\_chmod, event\_exec} &
``The process opened a file, changed ownership and permissions, and then executed it.'' \newline
\textit{Prompt: ``Could this pattern suggest privilege escalation in BSD?''} &
Permission Modification, Sudo Abuse &
T1222.002, T1548.001 \\
\hline

Android &
\texttt{event\_apkinstall, event\_netconn, event\_send, event\_wakelock} &
``The process installed an APK, initiated a network connection, sent data, and acquired a wakelock.'' \newline
\textit{Prompt: ``Is this activity consistent with Android spyware behavior?''} &
Malicious App Install, Sensitive Data Access, Autostart Execution &
T1476, T1409, T1547 \\
\hline
\end{tabular}
\end{table*}


\section{Conclusions and Future Work}
In this study, we presented a rigorous evaluation of transfer learning strategies for anomaly detection in Advanced Persistent Threat (APT) scenarios. By defining four distinct evaluation protocols ($P_0$ to $P_3$), we systematically quantified the effectiveness of transferring knowledge from a source domain to a target domain under varying levels of adaptation and methodological complexity. Our proposed method, based on an Attention-Augmented Autoencoder (AAE), demonstrated consistent and statistically significant improvements over traditional baselines across all datasets and operating systems. These gains were observed in both nDCG and AUC metrics, underscoring the robustness and generalizability of our approach.

Scatter plot analyses further highlighted the stability of transfer learning, with most detection methods showing upward shifts from protocol $P_0$ (no transfer) to $P_3$ (full transfer), especially for our AAE model. The strong alignment of points near the identity line in both nDCG and AUC plots confirmed that the latent representations produced by the AAE enable effective domain transfer. Moreover, the inclusion of statistical tools such as Wilcoxon signed-rank tests and Friedman tests validated the significance of the observed improvements across protocols.

\emph{Future Work:} Several promising directions remain open. First, we plan to investigate cross-OS transfer learning using more heterogeneous datasets and non-binary feature representations. Second, integrating temporal dynamics via sequence models (e.g., Transformers or RNNs) could enhance sensitivity to evolving APT stages. Third, we aim to explore federated transfer learning for decentralized environments where privacy constraints prohibit centralized training. Finally, explainability will be further enriched through attention-based attribution mechanisms, enabling transparent and interpretable anomaly scoring in real-time operational contexts.
\section*{Statements and Declarations}
\subsection*{Competing Interests}
The authors declare that they have no known competing financial interests or personal relationships that could have appeared to influence the work reported in this paper.%
\section*{Author contributions}
SB: Conceptualization, Methodology, Software,Visualization. SB: Data curation. SB, TR: Investigation. SB, TR: Writing.

\bibliographystyle{IEEEtran}

\bibliography{referencestransferlearning}

\end{document}